\documentclass[oneside,english]{book}
\usepackage[T1]{fontenc}
\usepackage[latin9]{inputenc}
\usepackage[a4paper]{geometry}
\usepackage{babel}

\usepackage{array}
\usepackage{float}
\usepackage{wrapfig}
\usepackage{units}
\usepackage{textcomp}
\usepackage{amsthm}
\usepackage{relsize}
\usepackage{amsmath}
\usepackage{graphicx}
\usepackage{amssymb}
\usepackage[unicode=true, 
 bookmarks=true,bookmarksnumbered=true,bookmarksopen=true,bookmarksopenlevel=4,
 breaklinks=false,pdfborder={0 0 1},backref=page,colorlinks=false]
 {hyperref}
\hypersetup{pdftitle={Assessment Of The Wind Farm Impact On The Radar},
 pdfauthor={Evgeny D. Norman},
 pdfsubject={Electronics},
 pdfkeywords={Radar, Propagation, Parabolic Equations, Doppler effect, Advanded Propagation Model, Pattern Propagation Factor, Control Theory, Wind Turbine, Wind Farm}}
 
\makeatletter

\DeclareRobustCommand{\lyxmathsym}[1]{\ifmmode\begingroup\def\b@ld{bold}
  \def\rmorbf##1{\ifx\math@version\b@ld\textbf{##1}\else\textrm{##1}\fi}
  \mathchoice{\hbox{\rmorbf{#1}}}{\hbox{\rmorbf{#1}}}
  {\hbox{\smaller[2]\rmorbf{#1}}}{\hbox{\smaller[3]\rmorbf{#1}}}
  \endgroup\else#1\fi}

\providecommand{\tabularnewline}{\\}
\floatstyle{ruled}
\newfloat{algorithm}{tbp}{loa}[chapter]
\floatname{algorithm}{Algorithm}

\theoremstyle{plain}
\theoremstyle{plain}
\newtheorem{thm}{Theorem}

\@ifundefined{showcaptionsetup}{}{%
 \PassOptionsToPackage{caption=false}{subfig}}
\usepackage{subfig}
\makeatother

\begin{document}
\frontmatter

\title{{\large ÉCOLE NATIONALE SUPÉRIEURE DES INGÉNIEURS DES ETUDES ET TECHNIQUES
D'ARMEMENT, Brest, France}\\
{\large and}\\
{\large THALES AIR SYSTEMS, Limours, France}\\
{\large ~}\\
{\large ~}\\
{\large by}\\
{\large Evgeny D. Norman}\\
{\large ~}\\
{\large ~}\\
{\large Final Year Project Report}\\
{\large Specialised Master in Architecture of Electronics and Computing
Complex Systems}\\
Assessment Of The Wind Farm Impact On The Radar\\
{\large ~}\\
{\large ~}\\
{\large Supervisors}\\
{\large Frédéric Campoy, Thales Air Systems}\\
{\large Arnaud Coatanhay, ENSIETA}\\
{\large ~}\\
{\large ~}\\
{\large ~}\\
{\large ~}\\
{\large ~}\\
{\large ~}\\
{\large ~}\\
}

\date{April - August, 2009}

\maketitle

\chapter*{\newpage{}Declaration }

I hereby declare that I am the sole author of this report. To the
best of my knowledge and belief, this report contains no material
previously published or written by any other person, except where
due reference is given in the text of the report.

This document has been written during an internship working for Thales
Air Systems in collaboration with École Nationale Supérieure des Ingénieurs
des Etudes et Techniques d'Armement (ENSIETA).

I empower École Nationale Supérieure des Ingénieurs des Etudes et
Techniques d'Armement to reproduce this report by photocopying or
by other means either the whole or any portion of the contents without
my further knowledge.

Also I cede copyright of the report in favour of  Thales Air Systems. 

© 2009 Evgeny D. Norman

\chapter*{Preface}
\begin{quotation}
Fortune is arranging matters for us better than we could have shaped
our desires ourselves, for look there, friend Sancho Panza, where
thirty or more monstrous giants present themselves, all of whom I
mean to engage in battle and slay, and with whose spoils we shall
begin to make our fortunes; for this is righteous warfare, and it
is God\textquoteright{}s good service to sweep so evil a breed from
off the face of the earth \cite{DonQuixote}.\end{quotation}
\begin{quote}
\begin{flushright}
- Don Quixote about windmills
\par\end{flushright}
\end{quote}
\begin{wrapfigure}{r}{0.3\columnwidth}%
\raggedleft{}\includegraphics[scale=0.3]{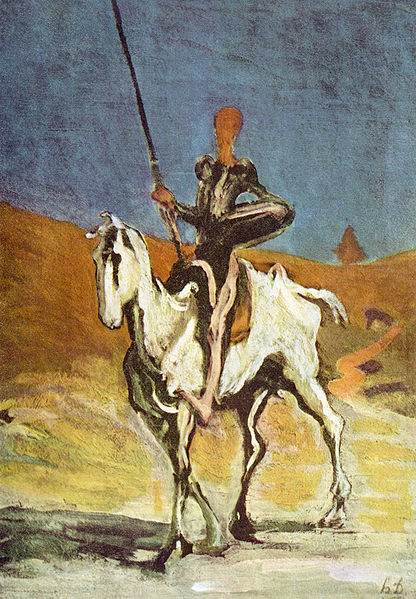}\end{wrapfigure}%

The made-up history written in 1605 repeats itself nowadays. However,
in contrast to delusions of Don Quixote%
\footnote{Painting: Honoré Daumier, Don Quixote, 1868, exhibited in Musée d'Orsay.%
}, modern wind turbines present the real challenge for the radar industry.
In comparison with more than century history of radars, wind turbines
have appeared relatively not long ago. In terms of the radar, wind
turbines differ from known nature and human made objects. They can
be as big as an Airbus A380, the velocity of their blades could be
compared with an aircraft speed. It leads to the new problem for the
radar detection and tracking. It concerns the majority of radars -
civil and military, surface and airborne, 2D and 3D. Wind farms can
have hundreds of wind turbines. One wind farm can cover a few square
kilometres. Each part, each turbine makes the signal, which interfere
with the radar. Although wind farms are, evidently, stationary, they
influence the filters, e.g. Kalman filter, and do not let track an
airplane. It is not easy to recognise the target even for a human
as well. The operator observes the same signals from turbines. In
fact, this problem concerns the security and safety. Difficulties
for military radars threaten the national security. For example, it
is difficult enough to detect the target flying at an extremely low
height. The wind farms make this task even more complex. On the other
hand the problem of the safety consists in difficulties for the air
traffic management. Primary radars cannot recognise the aircraft in
the area of wind farm. It means that for some time the operator has
not the necessary information. Considering that every second there
are thousands of airplanes in the air, one can see the scale of this
challenge. At the same time the wind power is considered as a source
of renewable energy. Nowadays, countries devote attention to the development
of this area. Consequently, one can predict a further spread of wind
farms.

Generally, this work shows the means to evaluate the wind farm impact
on the radar. It proposes the set of tools, which can be used to realise
this objective. The big part of report covers the study of complex
pattern propagation factor as the critical issue of the Advanced Propagation
Model (APM). Finally, the reader can find here the implementation
of this algorithm - the real scenario in Inverness airport (the United
Kingdom), where the ATC radar STAR 2000, developed by Thales Air Systems,
operates in the presence of several wind farms. Basically, the project
is based on terms of the department \textquotedbl{}Strategy Technology
\& Innovation\textquotedbl{}, where it has been done.

Also you can find here how the radar industry can act with the problem
engendered by wind farms. The current strategies in this area are
presented, such as a wind turbine production, improvements of air
traffic handling procedures and the collaboration between developers
of radars and wind turbines. The possible strategy for Thales as a
main pioneer was given as well.

\begin{flushright}
Evgeny D. Norman
\par\end{flushright}

\begin{flushright}
Limours
\par\end{flushright}

\begin{flushright}
August 2009
\par\end{flushright}

\chapter*{Acknowledgements}
\begin{quotation}
{}``We had the best of educations - in fact, we went to school every
day.''

{}``I\textquoteright{}ve been to a day-school, too,'' said Alice;
{}``you needn\textquoteright{}t be so proud as all that.''

{}``With extras?'' asked the Mock Turtle, a little anxiously. 

{}``Yes,'' said Alice, {}``we learned French and music.'' \cite{AlicesAdventuresinWonderland}

\begin{flushright}
- Alice and the Mock Turtle 
\par\end{flushright}
\end{quotation}
First and foremost I wish to thank my adviser at the ThalesAcademia,
Mme Odile Adrian, who is in charge of Advanced Developments. Thanks
to her I had the opportunity to undergo a training at Thales Air Systems.
She proposed me the current topic, carefully taking into account my
preferences, and guided me through this years. Mme Adrian kept inspiring
and motivating me, so I was doing my best.

Thank you to Frédéric Campoy for being my major advisor. At the beginning
of the internship he briefed me on the subject. M Campoy provided
me with all the necessary technical information, contacts and always
guided my work. He has supported me not only by providing a technical
and academic assistance, but also emotionally. Thanks to him I met
with many colleagues at ST\&I playing football on Wednesdays.

I am also grateful to M Gilles Beauquet and M Michel Moruzzis, who
initiated the current project. This report was not possible without
their technical assistance and consultations. M Beauquet gave me a
valuable data presented here.

My thanks also go to my adviser at ENSIETA M Arnaud Coatanhay. He
helped to plan my work and gave important advices. M Coatanhay read
my report making valuable remarks and comments.

Also thank you to everyone at the ST\&I team headed by M Cédric Lignoux.
This is a great professional team with an excellent atmosphere. It
is even a leader of different sport events where I was happy to participate
as well. There I met amazing people and I am extremely lucky to have
worked together. Also I appreciate your difficulties concerning security
questions and apologise for inconvenience if any. Thanks to M Lignoux
I was {}``baptisé'' and now I am able to make a scuba diving.

I would like to thank to Thales Group as a participant of the ThalesAcademia
programme, in particular, to its responsible, Mme Pia Ceccaldi, who
perfectly organised many official events during the year. I owe a
huge debt of gratitude to the other people at Thales Group and at
the French Ministry of Foreign and European Affairs, which participated
in the organisation of the ThalesAcademia programme.

I wish to acknowledge M Michel Rondy, who is responsible for International
Relations \& Masters at ENSIETA. Thanks to his perfect work I was
able to concentrate completely on the study. I have to especially
thank Mme Hélène Thomas, who is responsible for ESSE. She personally
watched over my educational process and helped me to acclimatise at
the beginning. 

I must say a special acknowledgement to students of Bureau des Etudiants
(BDE) and to classmates of ESSE. Thank you to everyone for those informal
evenings, fun and support. Personally, I am grateful to Cyril Bourgoin
and Xavier Quénet for warm welcome and hospitality. Since the first
day I arrived at ENSIETA Xavier helped me to get the hang of the subject.
Thank you to Astrid Michon, ESSE class, for being a good friend. Also
I want to thank my classmate Nikolay Kapyrin, who helped me to learn
French. That was priceless.

I cannot forget to say thanks to my mentors at Bauman Moscow State
Technical University (BMSTU) Prof. Panfilov Yu.V. and Prof. Volchkevich
L.I. They shaped my outlook on engineering as on the Art. They taught
me to do my work professionally and carefully, set high standards
and will remain an example for me.

\chapter*{Summary }

The accent of this master's report is made on the solution of wind
farm problematic, rather than on the presentation of the work that
has been done. However, everywhere the reader can find references
to original solutions which are given in DVD. The full articles, brochures
and presentations, which have references to bibliography in the text,
can be found in DVD.

Following is a brief synopsis of the report:
\begin{description}
\item [{Chapter}] \ref{cha:General-introduction}. This gives an historical
and general introduction to the problematic of wind farms within the
framework of the radar industry. Initial objectives for the internship
are described. The Chapter contains the presentation of Thales Air
Systems, its Surface Radar unit and department \textquotedbl{}Strategy
Technology \& Innovation\textquotedbl{} - the initiator of this topic.
\item [{Chapter}] \ref{cha:Overview-of-the}. State of the art is given
here. A few strategies to mitigate the effect of wind turbines are
shown. They has been compared and evaluated. Finally, we have chosen
the only strategy, which is be examined in the current report. Basically,
this Chapter is based on existed recent publications and presentations
in this area.
\item [{Chapter}] \ref{cha:Assessment-means}. In this chapter we show
the proposed means as a solution of chosen strategy. This means consists
of a few consecutive stages. Each stage is based on the tool. Subsections
\ref{sub:Advanced-propagation-model} - \ref{sub:ASTRAD-Module} present
these tools. The Section \ref{sec:Solution} shows how to use them.
2D, pseudo 3D and vector PE propagation models are given there as
well.
\item [{Chapter}] \ref{cha:Complex-pattern-propagation}. One of the tools
of Chapter \ref{cha:Assessment-means} - Advanced Propagation Model
- was studied in details. The output of APM is the pattern propagation
factor. Technically, Chapter is of great value for this master's report.
A new complex pattern propagation factor was defined and an existing
one was replaced. 
\item [{Chapter}] \ref{cha:Application-of-the}. Here we give an application
of the assessment means given in Chapter \ref{cha:Assessment-means}.
This is a real scenario with a radar of Thales, wind farm of Vestas
and environment in United Kingdom. As an outcome the influence of
wind farm was given in terms of RCS of each wind turbine. Recommendations
to reduce undesirable action of the wind farm were presented as well.
\item [{Chapter}] \ref{cha:Recommendations-for-Further}. This Chapter
has recommendations for the further work.
\item [{Chapter}] \ref{cha:Conclusion}. Here we conclude the result of
this project.
\end{description}

\chapter*{Abbreviations}

~
\begin{quote}
ALBEDO - software that calculates RCS (not an abbreviation)

APM - Advanced Propagation Model

APPF - Amplitude of Pattern Propagation Factor

AREPS - Advanced Refractive Effects Prediction System

ASTRAD - Architecture \& Simulation Tool for Radar Analysis \& Design

ATC - Air Traffic Control

ATM - Air Traffic Management

AWEA - American Wind Energy Association

BWEA - British Wind Energy Association

CAD - Computer-Aided Design

DTED - Digital Terrain Elevation Data

ENSIETA - École Nationale Supérieure des Ingénieurs des Etudes et
Techniques d'Armement

IEEE - Institute of Electrical and Electronics Engineers

FE - Flat Earth

FFT - Fast Fourier Transform

GUI - Graphical User Interface

NURBS - Nonuniform Rational B-Spline

PCT - Patent Cooperation Treaty

PE - Parabolic Equations

PPF - Pattern Propagation Factor

PPPF - Phase of Pattern Propagation Factor

RAM - Radar Absorbing Materials

RCS - Radar Cross Section

RO - Ray Optics

SSR - Secondary Surveillance Radar

ST\&I - Strategy Technology \& Innovation

TEMPER - Tropospheric Electromagnetic Parabolic Equation Routine

VPE - Vector Parabolic Equation

WT - Wind Turbine

XO - Extended Optics
\end{quote}
\tableofcontents{}

\listoffigures

\listoftables

\mainmatter

\chapter{General Introduction\label{cha:General-introduction}}

\section{Wind Farms as a New Challenge for the Radar Industry\label{sec:Wind-Farms-as}}

Wind farms interfere with radar. This leads to unfortunate results
for a few interested parties. The radar industry develops new features
for its radars trying to mitigate the wind farm influence. Users of
existed radar systems receive undesirable signals. Produces of wind
turbines and their customers face with limitations and rejections
of construction. The government is at the crossroads and has to look
for the compromise between the national security and green energy.

The wind turbine has two major features within the framework of the
interference with a radar - its size and revolving blades.

Blades when turn reflect electromagnetic waves creating Doppler effect,
and the radar considers them for moving objects. The wind turbine
never stands alone but operates together with the others in the wind
farm. It appears as a cluster of secondary targets on radar screens
or swamps the screen with multiple returns. However, this is not the
main problem. The main difficult is to detect and follow an aircraft
in this area. The blades make a very similar signature as radars do
because they have close aerodynamic design. The velocity of the blades
is also comparable. For example, we take the offshore wind turbine
GE Energy 3.6 \emph{MW}. To estimate we need certain parameters \cite{GE3.6MW}
(Table \ref{Flo:Info GE 3.6MW}).

\begin{table}[H]
\begin{centering}
\begin{tabular}{|r||r|}
\hline 
Diameter & 111 \emph{m }\tabularnewline
\hline 
Area swept & 9677 \emph{m$^{2}$}\tabularnewline
\hline 
Revolution speed & 15.3 \emph{rpm}\tabularnewline
\hline 
Operational interval & 8.5-15.3 \emph{rpm}\tabularnewline
\hline 
Number of blades & 3\tabularnewline
\hline 
Hub height & up to 100 \emph{m}\tabularnewline
\hline
\end{tabular}\caption{Technical information on GE Energy 3.6 \emph{MW}}

\par\end{centering}

\label{Flo:Info GE 3.6MW}
\end{table}

The general formula of an angular velocity is:

\begin{equation}
\overrightarrow{V}=\left[\overrightarrow{\omega},\,\overrightarrow{r}\right]\label{eq:Angular velocity}\end{equation}

The angular velocity of the blade's tip with respect to the centre
is determined by perpendicular component of the velocity vector \emph{V}:

$V=\omega r=2\pi fr=2\cdot\pi\cdot(\frac{15.3}{60})\cdot55.1\thickapprox88.3\unitfrac{m}{s}=317.8\unitfrac{km}{h}$

Now we can compare this value with the velocity of the aircraft. For
example, the cruising speed of Airbus A330 is 871$\unitfrac{km}{h}$.
This is a radial speed. When A330 has a tangent component, the radial
speed could be close to velocity of the blade. Moreover, each radial
point of the blade has its own velocity. It means that the radar receives
the frequency range.

Another {}``source'' of velocity is the velocity between the blade
and hub. When the blade passes the lower position and is parallel
to the hub, due to the airflow it deviates for a short period but
this is visible for the radar.

Dimensions of the wind turbine are also comparable with aircraft's
ones. For instance, one of the biggest airplanes - Airbus A380 - has
the length 73 \emph{m}, span 79.8 \emph{m}, while the length of the
normal turbine is 100 \emph{m}, the diameter is 111 \emph{m} (Table
\ref{Flo:Info GE 3.6MW}). Actually, a turbine is bigger than a plane,
and vertically it is a few times bigger than A380 in front. This leads
to the relatively big RCS. 

Wind farms will always make undesirable signatures on the radar display.
But one can significantly mitigate their influence and this problem
will become not very burning issue. It being known that some of the
solutions have already being applied. There is a set of measures and
solutions. Together they could solve this problem from different sides.

Mitigation measures may include modifi{}cations to wind farms (such
as methods to reduce a radar cross section; and telemetry from wind
farms to radar), as well as modifi{}cations to radar (such as improvements
in processing, radar design modifi{}cations, radar replacement, and
the use of gap fillers in radar coverage). 

There is great potential for the mitigation procedures. In general,
the government and industry should cooperate to find methods for funding
studies of technical mitigations. 

Once the potential for different mitigations is understood, there
are no scientific hurdles for constructing regulations that are technically
based and simple to understand and implement, with a single government
entity taking responsibility for overseeing the process. In individual
cases, the best solution might be to replace the aging radar station
with modern and fl{}exible equipment that is more able to separate
wind farm clutter from aircraft. This is a win-win situation for national
security, both improving the radar infrastructure and promoting the
growth of sustainable energy \cite{WindFarmsandRadar}. 

Ideas presented in this report correspond to latest directions in
a radar development. Solutions need more performance and flexibility
from the radar. Digital technologies open new functional capabilities.
Among different advantages of \emph{digital radar} \cite{adrian2008aesa}
the increased performance, throughput and multi-functionality let
implement real time estimation of the wind farm influence.

The problem concerns the regulations of air traffic as well.

Current circumstances provide an interesting opportunity for improving
the aging radar infrastructure.

\section{Thales\label{sec:Thales}}

\subsection{History}

In 1968, la Compagnie générale de la télégraphie Sans Fil (CSF) and
the department of professional electronics of Thomson-Brandt merge
and create Thomson-CSF \cite{Intgrationdunmodledepropagationdondeauseinduneplate-formedesimulationRadar}.
Between 1970 and 1980, the group signs the first big export contract
with Middle Eastern countries and diversifies its activities towards
telephonic communications, semiconductor silicon and medical imagery.

In 1982, Thomson SA is nationalized and the group goes in the public
sector. Then it meets financial difficulties and debts, which do not
cease to grow. This situation takes a turn for the better between
1983 and 1987 thanks to new strategies in professional electronics
and defence. The civil telecommunications are closed down and medical
sector is sold to General Electric.

In 1987, the fusion of semiconductor sector with Italian IRI-Finmeccanica
leads to SGS-Thomson.

Between 1987 and 1996, Thomson-CSF, expecting the recession of the
defence budget, begins the reorganization of its activities and active
politics to external growth of defence sector, generally in Europe,
with acquisition of electronic defence activity of Philips in 1989.
At the same time the group takes the control over Sextant Avionique.
Numerous acquisitions, more modest, extend the industrial presence
of the group beyond the national borders, basically in Europe. The
contribution of overseas subsidiaries is 5 - 25\% of revenue.

In 1998, Aérospatiale, Alcatel and Dassault Industries merge with
Thomson-CSF and Thomson SA. This creates on the one hand renewed Thomson-CSF
and Alcatel Space on the other hand. Alcatel Space includes the space
activities of Alcatel, Aérospatiale and Thomson-CSF. This step lets
Thomson-CSF consolidate its main activity, its competitive positions
in defence and industrial electronics as well as its implantation
in certain European countries (German, Italy, Norway, etc.) As a result
of these operations, in 1998, the majority of capital is transformed
in the private sector. The part of French state decreases from 58\%
to 40\%, in that way Alcatel and Dassault Industries become a shareholder.

After privatization, the dynamic of {}``multi-domestic'' consolidation
of defence activities that was in Europe in 90s goes out from European
continent: South Africa, Australia, Korea, Singapore, etc.

The group, thanks to internal growth and acquisitions, deeply modifies
its activities. The main direction is aimed to the civil market -
information technologies and the most dynamic sector mobile telecommunications.
In July 2000, new organisation, based on the defence, woks in three
directions: aeronautics and information technologies and service (IT\&S). 

In December 2000, Thomson-CSF becomes Thales and announces the creation
with U.S. Raytheon the first transatlantic joint venture between the
defence industry and the world leader of the air defence.

\subsection{Thales Today}

\begin{figure}[H]
\begin{centering}
\includegraphics[scale=0.3]{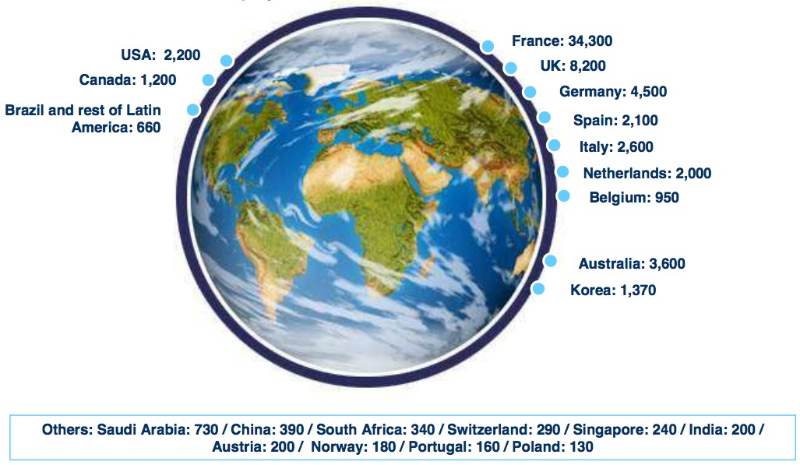}\caption{Global presence}
\label{Flo:Thales presence}
\par\end{centering}

\end{figure}

Since 2000, Thales followed the multi-domestic development, by taking
the control of some joint ventures in defence and in aeronautics and
Alcatel Space. Thales is a defence contractor and a major player in
civil and commercial markets around the world. Its businesses are
organised by market segment \textendash{} Aerospace \& Space, Defence
and Security \textendash{} and operate as a single organisation, sharing
advanced technologies and drawing on complementary capabilities to
meet the specific requirements of each customer. Mr Denis Ranque had
run the group since 1998 until 2009. Today the Chairman and Chief
Executive Officer is Luc Vigneron. Thales is presented in 50 countries.
The main countries are pointed out Figure \ref{Flo:Thales presence},
it represents about 68000 employees in the world. About 50 \% of employees
work out of France. 60 \% of them are engineers.

Thales achieved revenue growth of 8\% in 2008, sustaining organic
growth despite the impact of unfavourable exchange rates, due to the
fall in the pound sterling and the dollar against the euro. The annual
revenue is 12.7 billion euros. Order intake reached a high level in
2008, growing 14\% on a like-for-like basis, with new orders worth
2.7 billion euros for France and 2.8 billion euros for the United
Kingdom. At 31 December 2008, the consolidated order book stood at
almost 23 billion euros, equivalent to nearly two years of revenues.

\begin{figure}[H]
\begin{centering}
\includegraphics[scale=0.3]{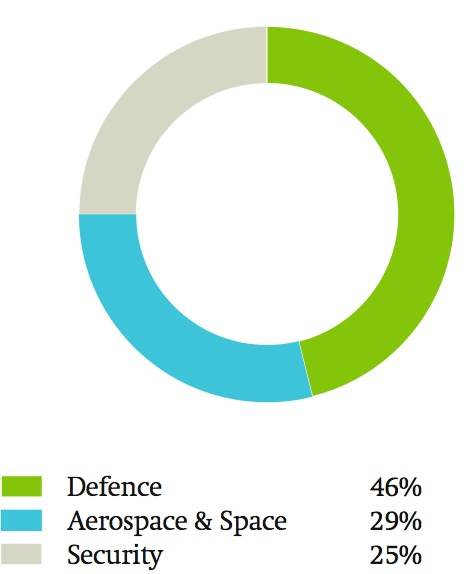}
\par\end{centering}

\caption{Activities of Thales}
\label{Flo:Thales structure}

\end{figure}

Thales operates in three main markets: Defence, Aerospace \& Space
and Security that represented in Figure \ref{Flo:Thales structure}
\cite{Forasaferworld}. 

Thales is committed to continuous improvement in the quality of its
systems, products and services throughout the life cycle. Underpinning
this commitment there are three basic principles: 
\begin{itemize}
\item Anticipating and answering customer requirements: Thales has a network
of commercial experts, key account managers and company-wide platforms
all over of the world.
\item Supporting customers: Thales guarantees the operational availability
and reliability of its products, equipment and systems, throughout
their life cycle. Thanks to its international footprint, the company
has the capability to provide customers with timely, tailored through-life
support and innovative services. A dedicated Internet portal, Customer
Online, is also available for customers.
\item Measuring the quality of all programme deliverables: Thales conducts
regular customer satisfaction surveys and carefully monitors each
customer's supplier ratings.
\end{itemize}
The group is based on three strategical pillars: its presence in all
chains, from equipment and systems to the integration of systems;
from technologies to dual applications, with balanced activities between
civil and military areas; and, finally, an international presence,
keeping up long-term partnerships with the clients.

Thales has a strong base in Europe with 55000 employees in 11 countries.
In France, the company employs 34000 people. Thales has an involvement
in all the country\textquoteright{}s major programmes for air, land,
naval, and joint forces (Rafale combat aircraft, FREMM multi-mission
frigates, Syracuse III telecommunication system, etc.).

\subsection{Thales Air Systems}

Thales Air Systems division (or simply Air Systems or TR6) develops
and gives total solutions of air safety, which bring an answer adapted
to needs of civil and military users. It is currently run by Richard
Deakin and includes 7 sectors (Weapon Systems, Thales Raytheon Systems,
Air Traffic Management Systems, Customer Service, Missile Electronics,
Navigation And Airport Solutions and Surface Radar). The division
has 6400 employees and an industrial presence in 14 countries supplemented
by an international commercial network. 

In civil aviation, the division develops air traffic management systems
as solutions for the security of airport zones, helping to land and
to navigate aircrafts. 

Within the framework of activities of air defence, this division offers
a complete range of surface radars, control systems of air operations,
solutions of protection of the battlefield and the sensitive sites
as well as anti-aircraft defence with both earth and naval weapon
system. 

The division offers a complete service range of service, of renovation
and extension of life as well as services of logistical engineering,
etc.

Aerospace markets offer a vivid illustration of the benefits of dual
civil/military technologies. Thales is the company with leadership
positions in both onboard equipment and ground- based navigational
aids and control systems. The company equips all types of aircraft
\textendash{} commercial airliners, military aircraft and helicopters
\textendash{} and is a first-tier partner of the world\textquoteright{}s
leading manufacturers, including Airbus, Dassault Aviation, Boeing,
Sukhoi and ATR, on all of their major programmes. In air traffic management,
Thales\textquoteright{} capabilities span the entire flight plan surveillance
and security chain, from departure gate through en-route control to
arrival gate, in complex and saturated air transport environments.
The Air Systems invested a noticeable part in Research and Development.

\subsection{Limours - Surface Radar }

\begin{figure}[H]
\begin{centering}
\includegraphics[scale=0.4]{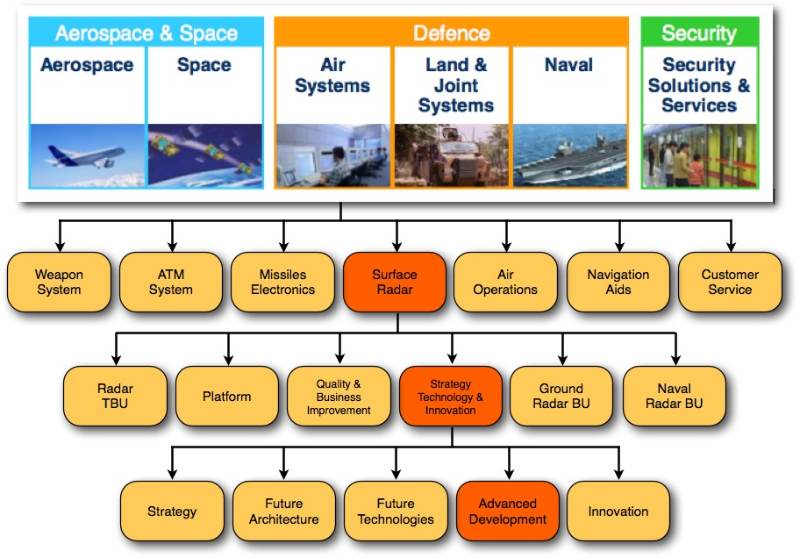}
\par\end{centering}

\caption{Organisation of Air Systems and the place of ST\&I }
\label{Flo:Thales STI}
\end{figure}

Business Line Surface Radar is run by Jean-Loïc Galle and represents
1550 employees in France and in Netherlands. The main target is to
develop detection systems and service: 
\begin{itemize}
\item Civil and military radars 
\item Surveillance radars
\item Short and long range radars 
\item Low frequencies (HF, UHF, etc.) and high bands (L, S, C, X, etc.) 
\item Passive and active radars
\end{itemize}
This report was written at Thales Group, Surface Radar unit (SR),
department Strategy Technology \& Innovation (Figure \ref{Flo:Thales STI}),
headed by Jean-Philippe Hardange. I worked in collaboration with Gilles
Beauquet, Michel Moruzzis and Frédéric Campoy. The responsible for
my internship is Frédéric Campoy. He briefed me on the subject and
helped with technical issues. Gilles Beauquet provided a number of
calculations shown in this work. Also Michel Moruzzis has initiated
the topic, guided me and participated in my work. Odile Adrian is
in charge of Advanced Developments. She is responsible for my educational
programme Thales Academia as well. Mme Adrian proposed this internship
according to my preferences and kept motivating me during the work
on this report.

In October 2006 Thales Air Systems in Limours has merged with the
site in Bagneux. Today there are about 640 employees. Thales Limours
combines production activities of Air System and its R\&D. Another
site of Air Systems is situated in Rungis (head office and R\&D).

\begin{figure}[H]
\begin{centering}
\includegraphics[scale=0.3]{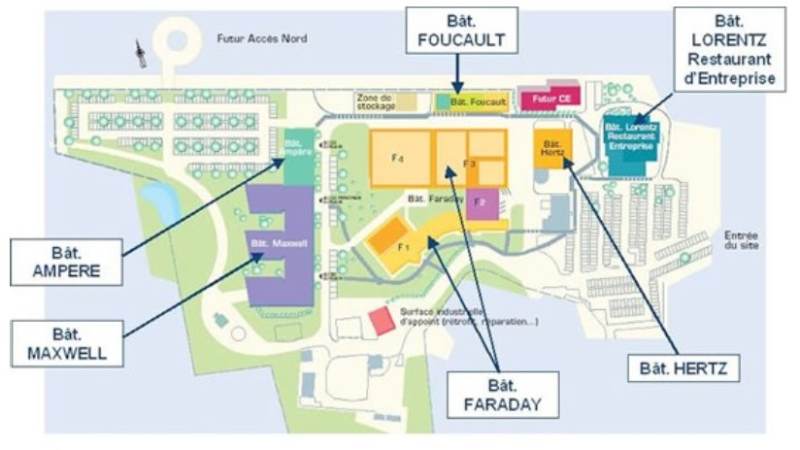}
\par\end{centering}

\caption{Site of Thales Air Systems in Limours}
\label{Flo:Thales Limours}

\end{figure}

My office is situated in the building Maxwell (Figure \ref{Flo:Thales Limours}).

\section{Objectives of Work\label{sec:Objectives-of-Work}}

Initially, according to the internship proposition there were three
main objectives: 
\begin{enumerate}
\item Drawing up a state of the art on the 2D and 3D pattern propagation
factors (PPF).
\item Defining a complex PPF function replacing an existing one.
\item Analysing the impact on a radar signal using an existing simulation
of Air Traffic Management radar and different wind farms scenarios.
\end{enumerate}
List of these objectives has been expanded in order to conduct the
complete research that can be used as a complex solution of the problematic
regarding wind farms. Therefore, I raised additional goals. 

Firstly, it is necessary to make a state-of-the-art that shows the
reasons why the wind power industry causes problems for radars. Also
in this part we need to show present solutions that allow mitigation
of the wind turbine impact on the radar. One of the solutions concerns
the initial objectives stated above. So we need to understand when
and how this solution can be used, its advantages and disadvantages.

As an an example, someone wants to construct the wind farm near the
radar. Each turbine interferes with this radar. However, before the
construction one can choose the place where the wind farm impact will
be minimal. The proposed solution must show the right place and give
quantitative and qualitative substantiation.

When we know the problematic in general, one must develop the chosen
solution. In order to realise that we need to use different software,
source code and develop our own one if needed. There is an important
element of our solution \textendash{} a program that propagates an
electromagnetic field in atmosphere. It is called Advanced Propagation
Model (APM). Complex pattern propagation factor (PPF) function concerns
exactly this program. Other elements of our solution are a 3D CAD
model of the wind turbine and a software that computes RCS. These
different programs can be combined into one solution by connecting
elements written, for instance, in Matlab because we need an output
data visualisation. 

The proposed solution should be tested using the real data of the
radar, wind farm and environment. The solution must give the practical
result as a recommendation of the best disposition of the wind farm.

Table \ref{Flo:Schedule of the project} presents the schedule of
this project.

\begin{table}[H]
\begin{centering}
\begin{tabular}{|l|l|c|}
\hline 
Main objectives & Sub-objectives  & Week\tabularnewline
\hline
\hline 
State-of-the-art & Make internal (Thales Air Systems) state-of-the-art & 15-16\tabularnewline
\cline{2-3} 
 & Make external state-of-the-art & 16-18\tabularnewline
\cline{2-3} 
 & Present the result & 18\tabularnewline
\hline 
Receive the complex PPF  & Develop Matlab script to run APM & 18\tabularnewline
\cline{2-3} 
 & Analyse APM code & 19-21\tabularnewline
\cline{2-3} 
 & Retrieve the phase from FE model of APM & 22\tabularnewline
\cline{2-3} 
 & Retrieve the phase from FE model of APM & 23\tabularnewline
\cline{2-3} 
 & Retrieve the phase from PE and XO model of APM & 24\tabularnewline
\hline 
Test the solution on the real scenario & Collect the necessary data about the radar,  & 25\tabularnewline
(simulation) & wind farm and environment & \tabularnewline
\cline{2-3} 
 & Prepare different wind farms scenarios & 26\tabularnewline
\cline{2-3} 
 & Develop Matlab script to run APM for the wind farm & 26\tabularnewline
\cline{2-3} 
 & Develop Matlab script to convert  & 27\tabularnewline
 & APM output for ALBEDO input & \tabularnewline
\cline{2-3} 
 & Receive RCS of wind turbines by ALBEDO & 28-29\tabularnewline
\cline{2-3} 
 & Analyse the result and corrections & 29\tabularnewline
\hline 
Make a fi{}nal report  & Plan the report & 30\tabularnewline
\cline{2-3} 
 & Write all the chapters  & 31-33\tabularnewline
\cline{2-3} 
 & Correct the text  & 33\tabularnewline
\cline{2-3} 
 & Make the presentation and DVD & 34\tabularnewline
\hline
\end{tabular}\caption{Schedule of the project}

\par\end{centering}

\label{Flo:Schedule of the project}
\end{table}

Although I have been working according to the schedule that I drew
up in 18$^{th}$ week at first I corrected my plan a few times during
the work. One can declare that I have done all that I planned excepting
the simulation of my project in ASTRAD. I suppose, this is an independent
project that can be done by a student, for example. However, all the
initial objectives given by Thales Air Systems has been completed.

\chapter{Overview of the Subject Matter\label{cha:Overview-of-the}}

\section{Introduction}

After the explication and comprehension of the problem we now show
the possible solutions to mitigate the wind turbine influence. There
are three main directions to do that. We begin with solution that
concerns only the radar (Section \ref{sub:Prediction-of-the}). That
does not need any relations with third party; it means that the problem
can be solved by us as a radar developer. The next solution is connected
with producers of wind turbines. The Section \ref{sub:Wind-Turbine-Development}
deals with materials and construction of the turbine; how to develop
the wind turbine with the goal that it will be less visible for a
radar. Other solutions concern administrative aspects: air traffic
management and relations with third party (government, wind energy
associations, customers, etc.) As a result we show the analysis and
comparison of these solutions (Section \ref{sec:Comparison-and-Choice}).
There, we specify one solution that will be considered in next Chapters
as a main part of the project.

Also we give different models, which simulate the propagation of electromagnetic
field in the space. This section is needed to understand the study
given in Chapter \ref{cha:Complex-pattern-propagation}.

\section{Strategies to Mitigate the Effect of Wind Turbines\label{sec:Strategies-to-Mitigate}}

\subsection{Prediction of the Influence and Signal Processing\label{sub:Prediction-of-the}}

It may be possible to use sophisticated radar signal and data processing
to overshadow turbine radar returns while preserving returns from
objects of interest, such as aircraft. It would seem much easier to
do so if the actual configuration of the turbines were known at every
instant. The data about the instantaneous state of every turbine (angular
velocity, phase, azimuthal orientation of the turbine axis, and pitch
angle) could be telemetered to the radar processors and electronics.
The turbine-mounted sensors needed for the four quantities listed
above are straightforward and not expensive (although not necessarily
available without retrofits). Armed with this information, the processor,
with the aid of a relatively simple model of the turbine radar cross
section, could make a near real time calculation of the time-varying
amplitude expected from each turbine in the farm and subtract it coherently
from the radar input signal. Significant networking, data processing,
and implementation challenges might exist, to be investigated in a
research project. We can predict not only the state of wind turbines
but also the state of environment using the real time data about wind,
temperature, refractivity, reflectivity, terrain, humidity, etc. This
will let us predict the influence of the atmosphere and surface on
the propagation of electromagnetic waves in order to know the propagation
loss in the area of each wind turbine. In that way knowing the electromagnetic
field and the state of the wind turbine, we are able to compute RCS
and evaluate the influence of entire wind farm. Thus, when the radar
receives the signal from a wind farm area, it may estimate a possible
presence of the aircraft. Of course it is necessary to provide the
{}``training'' of the radar using the measured and computed data,
design classifiers and even artificial neural networks. This is the
application for the radar signal processing. However, there is the
second application of this method. If we want to maintain the radar,
with this tool we can take into account the influence of existed wind
farms and, finally, choose the balanced place, where the performance
of our radar will be higher. On the other side, if the wind farm is
maintained, one can come to agreement with neighbour radar operators
about the placement of the farm \cite{WindFarmsandRadar}.

Exactly this means is examined in details in current report. Further
chapters cover the tools needed to realise that, the algorithm of
their application and an example. 

Also it is possible to deal with the gap fillers. When a wind farm
has caused an unacceptable loss of coverage, supplementary gap fi{}ller
radar could be installed, with appropriate data fusion. The gap fi{}ller,
by allowing a second view of the wind farm radar interference, makes
it considerably easier to process this interference out through data
fusion. 

The radar could be modified to have shorter pulses, a higher pulse
repetition frequency (PRF), and local oscillators coherent over a
turbine blade period, or multiple elevation beams to avoid ground
scraping. The higher PRF allows for painting a given turbine blade
with more pulses before the blade rotates significantly. The design
of the entire radar signature (including side lobes) needs to take
into account the presence of wind farms. Nevertheless, these modifications
must not worsen the primary function of a radar - to detect airplanes.

An interesting study has been done by QinetiQ for Swaffham wind farm
near the Prestwick Airport \cite{AssessmentoftheimpactdCreedwindturbineontheSandwickSSRradar}.

\subsection{Wind Turbine Development\label{sub:Wind-Turbine-Development}}

Evidently, the wind turbine developers, manufacturers and customers
do not care about the effect of wind turbines on the radar. At least
they did from recently. The customer makes high demands of the product
such as efficiency, performance, reliability, after-sales service,
some times low noise pollution, etc. He is far from believing that
wind farm can interfere with the radar. From the other hand, to satisfy
the client, the developer is looking for competitive solutions working
on light and durable materials, strong design, improvements of the
aerodynamic model. 

Here we examine features of the wind turbine development, which directly
affects the radar.

The radar emits electromagnetic beams and receives reflected waves.
The reflectivity depends mainly on the material. There is the study
about materials used in wind turbines \cite{ancona2001wind}.

Most rotor blades in use today are built from glass fibre reinforced
plastic. As the rotor size increases on larger machines, the trend
will be toward high strength, fatigue resistant materials. As the
turbine designs continually evolve, composites involving steel, glass
fibre reinforced plastic, carbon filament reinforced plastic and possibly
other materials will likely come into use.

The nacelle contains an array of complex machinery including yaw drives,
blade pitch change mechanisms, drive brakes, shafts, bearings, oil
pumps and coolers, controllers and more. Basically, the nacelle is
built from steal.

Low cost materials are especially important in towers, since towers
can represent upto 65\% of the weight of the turbine. Prestressed
concrete is a material that is starting to be used in greater amounts
in European turbines, especially in offshore or near-shore applications.
Concrete in towers has the potential to lower cost, but may involve
nearly as much steel in the reinforcing bars as a conventional steel
tower.

The following observations are based on the results of the material
usage analysis. Turbine material usage is and will continue to be
dominated by steel, but opportunities exist for introducing aluminium
or other light weight composites, provided strength and fatigue requirements
can be met. Blades are primarily made of glass fibre reinforced plastic,
which is expected to continue. While use of carbon filament reinforced
plastic may help to reduce weight and cost some, low cost and reliability
are the primary drivers. Increasing the use of offshore applications
may partially offset this trend in favour of the use of composites.
Prestressed concrete towers are likely to be used more, but will need
a substantial amount of steel for reinforcement.

As we can see, basically, wind turbines are built from conductive
materials. Its surface reflects electromagnetic waves well enough
to be visible for the radar. The visibility of a tower is not critical,
since the radar signal processing can separate stationary objects.
The main problem, as we have discussed above in Section \ref{sec:Wind-Farms-as},
is movable blades. The first idea is to use materials for blades,
which will be invisible for the radar. Such a technology is called
stealth. Generally, it is used in military area. There are recent
researches aiming at reducing reduce the reflectivity of rotor blades
\cite{Stealthbladesaprogressreport,Stealthtechnologies:removingradarasabarriertowindpower}.

\begin{figure}[H]
\begin{centering}
\includegraphics[scale=0.3]{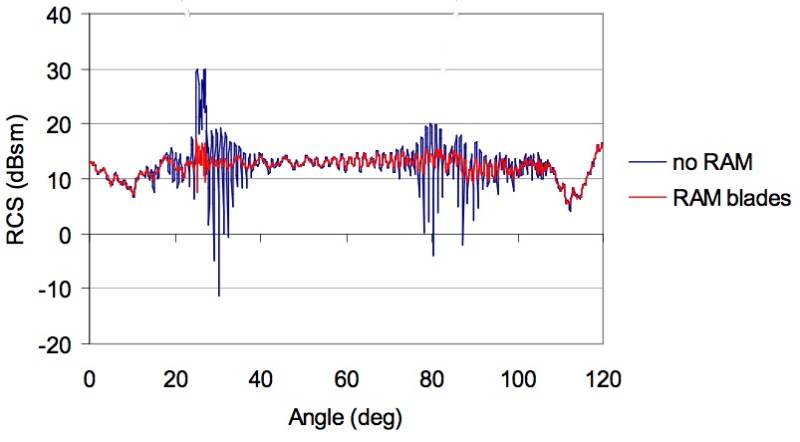}\caption{RCS predictions}
\label{Flo:Stealth RCS=0000A0predictions}
\par\end{centering}

\end{figure}

Using the study presented in Subsection \ref{sub:Prediction-of-the},
QinetiQ simulated two turbines - with standard blades and with blades
covered by radar absorbing materials (RAM). After RCS has been received.
The Figure \ref{Flo:Stealth RCS=0000A0predictions} presents the simulation
at 90\textdegree{} yaw and 0\textdegree{} pitch, blades are rotating towards and away from radar,
leading edge sees directly at radar. According to calculations, RCS peak is reduced from 30dBsm to 15dBsm.

\begin{figure}[H]
\centering{}\includegraphics[scale=0.3]{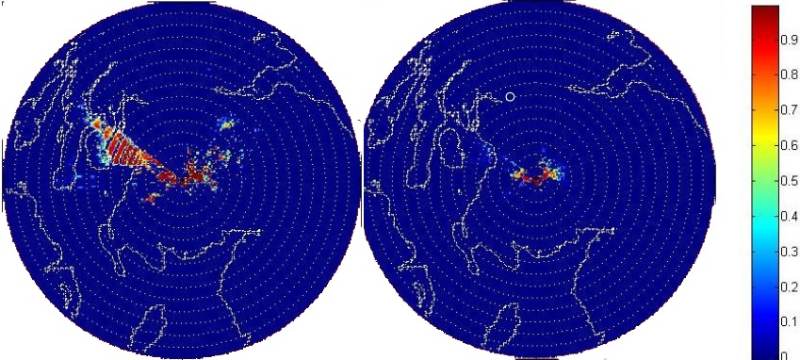}\caption{Impact of  the ''stealth'' treatment on the probability of detection of a
turbine with standard and radar absorbing materials (simulation)}
\label{Flo:Stealth simul}
\end{figure}

The result of the simulation, showing the possible difference between
standard blade and the blade with RAM, is shown in Figure \ref{Flo:Stealth simul}.
This is an example of the wind farm in Hare Hill site near Prestwick Airport.

\begin{figure}[H]
\centering{}\includegraphics[scale=0.2]{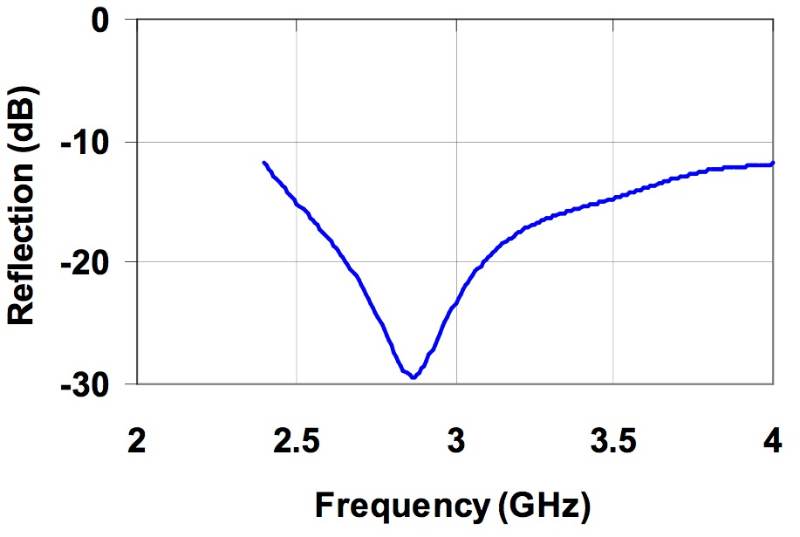}\caption{Measured performance for a blade: ­10\emph{dB} = 10\% reflection,
­20\emph{dB} = 1\% reflection, ­30\emph{dB} = 0.1\% reflection}
\label{Flo:Stealth measured}
\end{figure}

The main shortcoming of this direction is that RAM cover only one
frequency band (Figure \ref{Flo:Stealth measured}). For example,
it works for ATC radars but for the military application, where other
bands are used, does not.

However, one must mention that there are studies to design optimised
composite RAM for operation over wide frequency ranges (typically
within the range of 1-100 \emph{GHz}).

Another barrier of adoption of this technology is its cost \cite{WindFarmsandRadar}.
QinetiQ, for instance, claims that the cost penalty for such treatment
is of the order of 10\% of the total blade cost. We can compare this
with the radar installation. Radars, which don\textquoteright{}t have
the capabilities to mitigate wind farm interference, could simply
be replaced, in a phased upgrade of the aging radar infrastructure.
The new radar would incorporate multidimensional detection, with greatly
enhanced processing, with pulse shapes designed to optimally distinguish
between the aircraft and wind farms. The cost of a single radar installation
was said to be in the range of €3-8M, to be compared with the €2-4M
cost of a single wind turbine. A wind farm can have hundreds of turbines. 

For example, we want to install a wind farm with 25 turbines. The
price of each is €3M. Using RAM technology, there will be additional
expenses 10\% - it means €7.5M (€3M$\cdot$25$\cdot$10\%). The price
of this new material is equal to the installation of one or two radars
with modern signal processing.

This strategy to mitigate the effect of wind turbines is relatively
new. Presently, in 2009, there is no serial production of RAM for
wind farms, and there is still no wind farm with this technology.
Only R\&D has been done.

Another proposal is to put an active layer on the outside of the turbine
blades to modulate dynamically the blade Doppler signature. These
modulations, it is claimed, could shift the Doppler frequency spectrum
from the blades to lie outside the range of frequencies processed
by the radar. It is not known, to us at least, whether such modifi{}cations
to the outside of the blades would produce unacceptable changes to
their aerodynamic properties or whether they would last the lifetime
of the blades. 

Besides these solutions, there are a few recommendations based on
study \cite{poupart2003wind} regarding the design of certain turbine's
parts. The following points summarise some of the results:
\begin{itemize}
\item The design of the tower and nacelle should have the smallest RCS signature
possible. The RCS of the tower and nacelle can be effectively reduced
though careful shaping.
\item Large turbines do not necessarily lead to large RCS (i.e. tower height
does not greatly affect RCS).
\item For a low probability of detection, but a large clutter return, set
wind turbines such that they are mainly yawed close to \textpm{}90\textdegree{}
from the radar direction.
\item For a high probability of detection, but a smaller area of clutter,
set wind turbines such that they are mainly yawed close to 0\textdegree{}
and 180\textdegree{} from radar direction.
\end{itemize}

\subsection{Air Traffic Handling Procedure\label{sub:Air-Traffic-Handling}}

Regulatory changes for air traffic could make considerable impact
on the problem. For example, the government could consider mandating
that the air space up to some reasonable altitude above an air-security
radar with potential turbine interference be a controlled space, with
transponders required for all aircraft flying in that space. This
would both solve the problem of radar interference over critical wind
farms and would provide a direct way to identify bad actors, flying
without transponders \cite{WindFarmsandRadar}.

\subsection{Others\label{sub:Others}}

It is evident that the problem involves different actors: radar and
wind turbine developers, government and clients both of radars and
turbines. The only way to find the solution is to speak and collaborate
with each other. In that way they could understand their opportunities
and limitations, find the compromise, put forward their arguments.
It is necessity at least to outline the possible solution. In order
to provide the dialogue one must organise the common symposiums, conferences
and meetings. As a result we could receive the standards for the turbine's
construction, methods to estimate the influence of wind farms, start
to conduct common research. The actors have to find the balance between
the safety of air traffic, national security and the green energy
that are one of the national priorities as well. There are many wind
energy associations, which are quite influential and usually presents
producers, government and clients partially. The radar industry can
work through them. Moreover, the wind energy is successfully developing,
and associations sponsor many projects every year. There is a demonstrative
example in the United Kingdom. The sponsorship has existed there since
the beginning of the XXI century. A lot of money has been invested
\cite{WindRDintheDTIsEmergingEnergyTechnologiesProgramme} to the
development of appropriate solutions for the mitigation of radar interference
from wind turbines. Among the grantees one can hear famous names QinetiQ,
Vestas Blades Ltd., BAE Systems and many others. Generally, it is
the government that makes grants.

It is necessary to create the complete database about every wind farm.
These bases could contain coordinates of turbines, their model, characteristics,
the information about environment where turbines operate, and so on.
The access should be given to producers of radars. For instance, in
the United Kingdom there is existed database with the information
about all wind farms there. This project has been done by the British
Wind Energy Association (BWEA) and is available through www.bwea.com.
The access to detailed wind farm information is given only for members.

\begin{figure}[H]
\begin{centering}
\subfloat[BWEA]{

\begin{centering}
\includegraphics[scale=0.3]{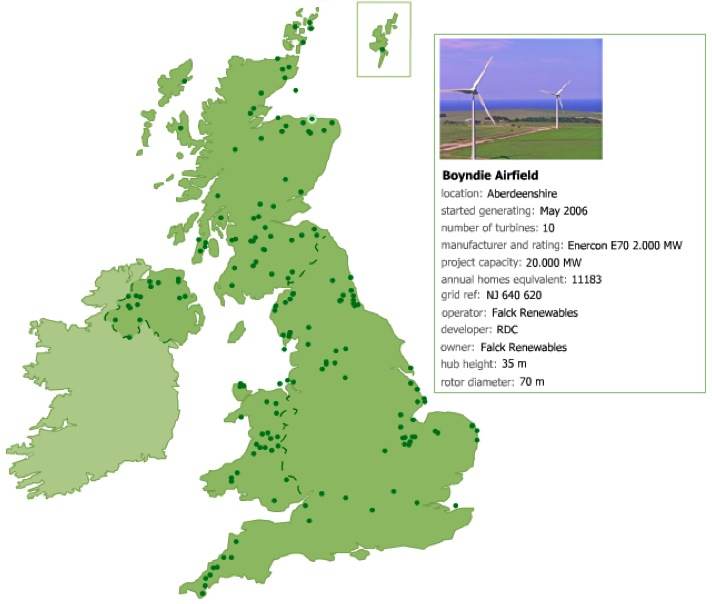}
\par\end{centering}

}\subfloat[AWEA]{

\begin{centering}
\includegraphics[scale=0.3]{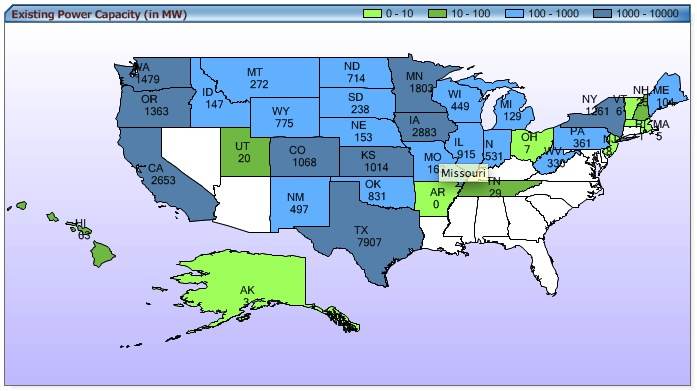}
\par\end{centering}

}\caption{Databases}
\label{Flo:Wind farm data base}
\par\end{centering}

\end{figure}

There is an influential national society in the US - American Wind
Energy Association (AWEA) as well. AWEA conglomerates all activities
of wind power industry in US. In comparison with the UK market, an
american one is bigger and presents more opportunities for cooperation
with the radar industry in this field. Interfaces of both BWEA and
AWEA are shown in Figure \ref{Flo:Wind farm data base}.

BWEA has a list of wind farms under construction, operational, consented
and submitted. In Chapter \ref{cha:Application-of-the} we will examine
one example of the wind farm in the UK. The information about this
farm can be found in BWEA database as well (Table \ref{Flo:BWEA operational}%
\footnote{www.bwea.com/ukwed/operational.asp%
}).

\begin{table}[H]
\begin{centering}
\begin{tabular}{|r||r|}
\hline 
Online & September 2006\tabularnewline
\hline 
Wind farm & Beinn Tharsuinn\tabularnewline
\hline 
Location & Highland\tabularnewline
\hline 
Turbine model & Vestas\tabularnewline
\hline 
Power, \emph{MW} & 1.65\tabularnewline
\hline 
Turbines & 17\tabularnewline
\hline
\end{tabular} ~\begin{tabular}{|r||r|}
\hline 
Capacity, \emph{MW} & 30\tabularnewline
\hline 
Homes equiv. & 16774\tabularnewline
\hline 
Developer & Scottish Power\tabularnewline
\hline 
Owner & Scottish Power\tabularnewline
\hline 
Latitude & 57 48 06 N\tabularnewline
\hline 
Longitude & 04 19 56 W\tabularnewline
\hline
\end{tabular}
\par\end{centering}

\centering{}\caption{Operational wind farm in the UK, the data from BWEA}
\label{Flo:BWEA operational}
\end{table}

One can see that besides existed turbines there are two turbines under
construction in the area of our wind farm constructed by new developer
(Table \ref{Flo:BWEA construction}%
\footnote{www.bwea.com/ukwed/construction.asp%
}).

\begin{table}[H]
\begin{centering}
\begin{tabular}{|r||r|}
\hline 
Date & February 2009 \tabularnewline
\hline 
Wind farm & Beinn Tharsuinn extension \tabularnewline
\hline 
Location & Highland\tabularnewline
\hline 
Power, \emph{MW} & 2.3\tabularnewline
\hline 
Turbines & 2\tabularnewline
\hline
\end{tabular} ~\begin{tabular}{|r||r|}
\hline 
Capacity, \emph{MW} & 4.6\tabularnewline
\hline 
Homes equiv. & 2572\tabularnewline
\hline 
Developer & RockBySea \& Midfern Renewbles \tabularnewline
\hline 
Latitude & 57 48 06 N\tabularnewline
\hline 
Longitude & 04 19 56 W\tabularnewline
\hline
\end{tabular}
\par\end{centering}

\centering{}\caption{The UK wind farm under construction, the data from BWEA}
\label{Flo:BWEA construction}
\end{table}

This project of BWEA lets us suppose with good reason that the unified
complex European database with all wind farms could exist and be needed.

\section{Comparison and Choice of the Strategy\label{sec:Comparison-and-Choice}}

Finally, we can conclude about directions of solutions of the wind
farm problematic. Let us take a look at the all strategies briefly:
\begin{itemize}
\item The method of evaluation of the wind farm influence on the radar.
\item The improvement of the radar signal processing.
\item The wind turbine development using radar friendly solutions (basically
materials and design).
\item The Air Traffic Handling Procedure.
\item Other collaboration with wind energy industry and implementation of
common projects.
\end{itemize}
It is wrong to suppose that wind power industry is the negative factor
for the radar producer. Quite the contrary, it opens the new market
and opportunities. Taking into account the activities of Thales, one
can declare that Thales Group has already had all the necessary means
to make a complex solution. Thus, there is Surface Radar (SR) unit
to develop the modern signal processing with updated filters and thresholds.
This is a stimulus to replace the ageing radar station with modern
and flexible equipment of Thales that can separate the wind farm clutter
from the aircraft. Additionally, in particular, ST\&I currently carries
out the study of impact evaluation. There is ATM unit to provide solutions
in ATC. Thales has more international presence than its competitors
in this field. Playing the main role in French defence, Thales has
an experience of work with French government. This is the time to
start the close collaboration with wind energy industry, state and
with customers of wind energy. In other words, today the wind power
industry is the new unique sector, where the civil and military interests
are crossed. Thales has all opportunities to be the pioneer in new
technologies, regulations and solutions. In any case all the steps
should be made in cooperation with wind energy associations.

However, now Thales is not a leader in {}``stealth'' materials for
wind farms (e.g. QinetiQ succeeded there and patented its solutions).
But there are doubts that this technology will be competitive in comparison
with others.

Below we summarise different aforementioned methods and compare them
in Table \ref{Flo:Summary of methods}.

\begin{table}[H]
\centering{}\begin{tabular}{|l|c|c|c|c|}
\hline 
 & Simplicity of R\&D & Simplicity of application & Diversity of & Benefit\tabularnewline
 & and its cheapness  & and its cheapness  & applications & \tabularnewline
\hline 
Evaluation  & $\bigstar$$\bigstar$ & $\bigstar$$\bigstar$ & $\bigstar$$\bigstar$$\bigstar$ & $\bigstar$$\bigstar$\tabularnewline
of Influence &  &  &  & \tabularnewline
\hline 
Signal  & $\bigstar$  & $\bigstar$$\bigstar$$\bigstar$ & $\bigstar$$\bigstar$ & $\bigstar$$\bigstar$$\bigstar$\tabularnewline
Processing &  &  &  & \tabularnewline
\hline 
Air Traffic & $\bigstar$$\bigstar$ & $\bigstar$ & $\bigstar$$\bigstar$ & $\bigstar$$\bigstar$\tabularnewline
Management &  &  &  & \tabularnewline
\hline 
WT Development &  &  &  & \tabularnewline
\quad{}$\bullet$ materials & $\bigstar$ & $\bigstar$ & $\bigstar$ & $\bigstar$ - $\bigstar$$\bigstar$$\bigstar$\tabularnewline
\quad{}$\bullet$ construction & $\bigstar$$\bigstar$$\bigstar$ & $\bigstar$$\bigstar$$\bigstar$ & $\bigstar$$\bigstar$$\bigstar$ & $\bigstar$\tabularnewline
\hline
\end{tabular}\caption{Summary of the methods}
\label{Flo:Summary of methods}
\end{table}

At R\&D level the evaluation of the wind farm impact needs a consolidation
of different areas of studies and corresponding software. Also in
order to validate the solution it is necessary to conduct research
with a real radar station and wind farms. Therefore, the project needs
time and economic resources. Talking about the simplicity of its application,
we can declare that every case should be considered separately because
every time we deal with a new radar, environment and wind farm. In
other words every project is unique. However, regarding the diversity
of this solution, the uniqueness of the project is an advantage. We
can evaluate the wind farm impact for every radar. But the solution
usually brings the benefit for the only radar. So we can hide, for
example, the wind farm from one radar but it will be visible for others.

The signal processing is complicated in terms of R\&D. It includes
the complexity of the first solution and also needs new mathematical
models. But once the signal processing is done, its application would
not be very sophisticated. This solution works for existed types of
wind turbines. That is why its diversity is limited. The evolution
of the turbine's design is advancing. Thus, the new design can need
the new signal processing. 

Rules for the Air Traffic Management do not seem very complicated.
But their application is not easy because it concerns many players:
airports, air companies and radar developers. It would be very hard
to find the common solution and to implement it. The diversity of
applications and the benefit are limited by air traffic control. The
solution does not concern military radars, for example, which provide
the air security and they do not care about air traffic control. 

The R\&D of {}``stealth'' materials for wind turbines are quite
complicated and its new solutions appreciably raise the turbine's
price. Moreover, these materials usually work for the certain frequency
range. That is why the diversity of applications is very limited.

Regarding the objectives of this report, we examine the study about
the evaluation of the wind turbine influence. The study is considered
as a complete and independent method. At the same time this is the
necessary part of the signal processing that is described in Subsection
\ref{sub:Prediction-of-the}. Also the study contains steps, which
could be used in development of wind turbines. Therefore, further
chapters contain the relevant information for the problem in general.

\section{2D and 3D Propagation}

\subsection{2D\label{sub:2D}}

Basically, the main 2D propagation model is a parabolic equation.
The model is implemented in a few software presented on the market:
NEMESIS by QinetiQ, AREPS by SSC San Diego, TERPEM by Signal Science
Ltd., and TEMPER by the Johns Hopkins University Applied Physics Laboratory.
Actually, AREPS and TERPEM are an advanced GUI of the APM code.

The parabolic equation models propagation of energy predominantly
in one direction only. The field values at each step are calculated
from the field values at the previous step. This is done subject to
a radiation boundary condition at the top of the field plane and a
terrain boundary condition at the bottom of this plane. Using this
method, the shape and the electrical constants of the terrain can
be accurately incorporated into the model. The majority of long-range
propagation prediction methods in use are 2D and consider propagation
along the great circle path between the transmitter and receiver.
In details PE is considered in Subsection \ref{sub:Parabolic-Equations}.

There is another 2D propagation model - ray optics (RO). The model
is based on the propagation of beams. The value of the field in the
certain point of space is a superposition of all waves - direct and
reflected \cite{barrios2002advanced}.

\subsection{Pseudo 3D\label{sub:Pseudo-3D}}

As stated above a 2D model neglects the propagation in a horizontal
plane and the ray reflects only in a vertical plane. However, it is
possible to run the 2D model along the third axe. In this way we can
receive a 3D view. Vertical layers that begin from the same point
compose the 3D view with an angular step. Also these layers can be
parallel if we examine the place at a long distance where the calculating
error is negligible. As a result we have the field in the volume.
The shortcomings of a 2D propagation apparently remain. The pseudo
3D model does not need any new physical or mathematical theory. It
always uses the principals of a 2D model, which can be based on parabolic
equations, ray optics, etc. That is why it is called {}``pseudo''.
In other words the pseudo 3D model consists of layers where the electromagnetic
field is computed using the 2D model.

\begin{figure}[H]
\begin{centering}
\includegraphics[scale=0.5]{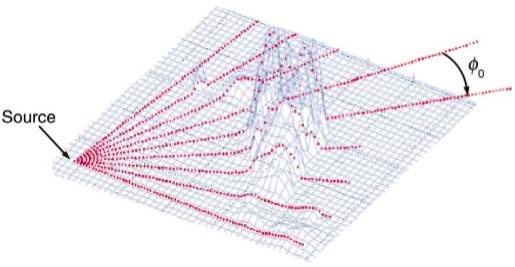}
\par\end{centering}

\caption{Pseudo 3D propagation model}
\label{Flo:pseudo 3D propagation model}

\end{figure}

Propagation over 2D terrain can be simulated by running APM over multiple
1D azimuths (Figure \ref{Flo:pseudo 3D propagation model}) with the
terrain assumed to be 1D along each azimuth. The 2D propagation data
from these azimuths are then combined to approximate propagation in
a 3D environment. It is evident that this approach does not capture
out-of-plane scattering and diffraction effects \cite{ra2004modeling}.

There are a lot of studies that use the plane model in order to examine
the volume. Pseudo 3D model can give satisfactory results. There is
a study \cite{ra2004modeling} where the result of pseudo 3D model
is compared with vector PE (see the next Subsection). The calculating
error of the pseudo 3D model depends on a terrain profile - if it
is relatively homogeneous, results are close to each other. Both model
were developed by the Johns Hopkins University Applied Physics Laboratory,
the U.S. The comparison of models developed in University of Birmingham,
the UK is presented in \cite{zelley1999three}.

\subsection{Vector PE\label{sub:Vector-PE}}

A vector version of the parabolic equation method is required to treat
general three-dimensional electromagnetic problems. The vector PE
is obtained by coupling component scalar parabolic equations via suitable
boundary conditions on the scatterers {[}\cite{levy2000parabolic}{]}.
This allows accurate treatment of polarisation effects within the
paraxial constraints.

The electric and magnetic fields are defined by $E=(E_{x},E_{y},E{}_{z})$
and $H=(H_{x},H_{y},H{}_{z})$, respectively \cite{zaporozhets1999bistatic}.
For two-dimensional (2D) problems when the fields are independent
of the transverse coordinate \emph{y}, the simplest option is to take
is:

\begin{equation}
\begin{cases}
\psi=E_{y} & ,\, for\, horizontal\, polarization\\
\psi=H_{y} & ,\, for\, vertical\, polarization\end{cases}\end{equation}

The determination of $\psi$ suffices to solve the whole electromagnetic
problem: by using the curl equations, all field components are determined
and they automatically satisfy the divergence-free conditions. 

The situation is, of course, different in three dimensions. Some additional
effort is then required to get a solution satisfying Maxwell\textquoteright{}s
equations. First, we obtain a scalar wave equation for each electromagnetic
field component from the curl equations. Second, these component scalar
wave equations are coupled through boundary conditions on the scattering
object and through the divergence-free condition in order to obtain
a well-determined system. In this work, we only examine the case of
perfectly conducting scatterers. Then boundary conditions on the object
can be written in terms of the electric field only, so that we have
a self-contained system of equations to solve for the electric field.
The magnetic field can be obtained through the curl equation if required.
For a perfect conductor, the tangential electric field must be zero
on the object or, equivalently, the electric field must be parallel
to the normal. This gives the following system of equations: 

\begin{equation}
\begin{cases}
n_{x}E_{y}(P)-n_{y}E_{z}(P)=0\\
n_{x}E_{z}(P)-n_{z}E_{x}(P)=0\\
n_{y}E_{x}(P)-n_{x}E_{y}(P)=0\end{cases}\label{eq:system of equations}\end{equation}

where \emph{P} is a point on the surface of the scatterer and $\overrightarrow{n}=(n_{x},n_{y},n_{z})$
is the outer normal to the surface at \emph{P}. In terms of the PE
reduced scattered field, these conditions become non homogeneous.
For example the fi{}rst equation of Eqn. \ref{eq:system of equations}
is written as 

\begin{equation}
n_{x}u_{y}^{s}(P)-n_{y}u_{z}^{s}(P)=-e^{ikx}(n_{x}E_{y}^{i}(P)-n_{y}E_{z}^{i}(P))\label{eq:first eqn}\end{equation}

where $E=(E_{x}^{i},E_{y}^{i},E_{z}^{i})$ is the incident electric
field. 

The three equations in Eqn. \ref{eq:system of equations} are not
independent, but form a system of rank 2. Hence, we need another equation
to ensure unity of the solution. This is provided by the divergence-free
condition of Maxwell\textquoteright{}s equations:

\begin{equation}
\frac{\partial E_{x}^{s}}{\partial x}+\frac{\partial E_{y}^{s}}{\partial y}+\frac{\partial E_{z}^{s}}{\partial z}=0\label{eq: divergence-free condition}\end{equation}

where $\left(E_{x}^{s},E_{y}^{s},E_{z}^{s}\right)$ is the scattered
electric field. It should be noted here that in the 2D case the fi{}elds
are automatically divergence-free since we solve for $ $$E_{y}^{s}$
or $H_{y}^{s}$, which do not depend on \emph{y}. This is no longer
true in the 3D case, where the divergence-free condition must be enforced
explicitly. Enforcing the divergence-free condition on the object
boundary ensures a well-determined system of equations, and we show
in the Appendix that the PE solution is then divergence-free everywhere.
The parabolic equation formulation avoids the need for direct estimation
of the range derivatives in the divergence-free condition, yielding
an expression involving points in the transverse plane only. We can
rewrite \ref{eq: divergence-free condition} as:

\begin{equation}
\frac{i}{2k}\left(\frac{\partial^{2}u_{x}^{s}}{\partial y^{2}}+\frac{\partial^{2}u_{x}^{s}}{\partial z^{2}}\right)+iku_{x}^{s}+\frac{\partial u_{y}^{s}}{\partial y}+\frac{\partial u_{z}^{s}}{\partial z}=0\end{equation}

Vector PE is implemented by a few institutions: Johns Hopkins University
Applied Physics Laboratory, the U.S., University of Birmingham, the
UK, the Radio-communications Agency of the Department of Trade and
Industry. However, there is no successful application for the atmospheric
propagation like 2D models (APM, TERPEM, TEMPER, NEMESIS). The model
was generally driven by Andrew A. Zaporozhets \cite{Applicationofvectorparabolicequationmethodtourbanradiowavepropagationproblems}
and Mireille F. Levy \cite{levy2000parabolic}.

\chapter{Assessment Means\label{cha:Assessment-means}}

\section{Introduction}

The assessment means is a set of tools (here Section \ref{sec:Proposed-Tools})
combined together in order to evaluate the wind farm impact. Actually,
each tool is an independent program: APM, Rhino NURBS, ALBEDO and
ASTRAD. We show how to apply them for our needs in Section \ref{sec:Solution}.
For this reason we develop additional Matlab scripts so that we can
estimate the wind farm impact.

\section{Proposed Tools\label{sec:Proposed-Tools}}

\subsection{Advanced Propagation Model \label{sub:Advanced-propagation-model}}

Advanced propagation model (APM) calculates different characteristics
of electromagnetic propagation. It uses the set of input parameters
about the source, environment and terrain. The source code is written
in Fortran 90. The big advantage of APM is that we have the source
code and can correct it for our purpose. It is possible to integrate
APM as a module of our assessment means. Its output can be automatically
used for the next steps.

APM is integrated in a few software. Initially, APM gives only the
result of calculations of different propagation models. It does not
visualise its result. Also APM has not any GUI. That makes the understanding
of APM relatively hard, especially at the beginning. At the same time
the lack of GUI is the reason why we can adapt APM to our needs either
we want to use GUI or not. There are some successful examples of the
visualisation of APM - AREPS and TEMPER. These programs make calculations
using APM. It is easy and clear for user to enter input parameters.
AREPS, for example, offers the statistical data for almost every place
of the Earth about air temperature, refractivity profile depending
on the time of the year, the possibility to read the standardised
data about the terrain and weather, etc. The program has the other
tools, which makes the work with APM much easier: the possibility
to operate with different projects, to work with the table information,
etc. For the user it is more comfortable than the APM source code.
Thales, for example, uses AREPS during some projects.

ST\&I works with APM version 1.3.5. There are two subversions, which
differ only in structure of their functions. The first one consists
of 53 files: 51 functions, 1 main program and 1 module. The second
version consists of 3 files (\emph{apmmain.f90}, \emph{apmsubs.f90},
\emph{apm\_mod.f90}), where these 51 functions (subroutines) are combined
in one file - \emph{apmsubs.f90. Apmmain.f90} starts the program using
the variables defined in \emph{apm\_mod.f90. Apmsubs.f90} has all
the functions to compute different propagation models (FE, RO, PE
and XO). The APM metrics is shown in Table \ref{Flo:APM metrics}.

\begin{table}[H]
\begin{centering}
\begin{tabular}{|l||c|}
\hline 
Classes & 0\tabularnewline
\hline 
Files & 3\tabularnewline
\hline 
Library Units & 53\tabularnewline
\hline 
Lines & 10121\tabularnewline
\hline 
Lines Blank & 2506\tabularnewline
\hline 
Lines Code & 4184\tabularnewline
\hline 
Lines Comment & 3587\tabularnewline
\hline 
Lines Inactive & 0\tabularnewline
\hline 
Executable Statements & 4188\tabularnewline
\hline 
Declarative Statements & 390\tabularnewline
\hline 
Ratio Comment/Code & 0.86\tabularnewline
\hline
\end{tabular}\caption{APM metrics}
\label{Flo:APM metrics}
\par\end{centering}

\end{table}

The program reads external file with input parameters (in our case
\emph{file.in}, see Algorithm \ref{Flo:APM input file}) and creates
an output file as a result (\emph{file.out}, see Algorithm \ref{Flo:APM output file}). 

\begin{algorithm}[H]
...

2800. :Frequency in MHz 

20. :Antenna height in m 

1 :Antenna type (1=OMNI,2=GAUSS,3=SINC(X),4=COSEC2,5=HTFIND,6=USRHTFIND,7=USRDEF) 

0 :Polarization (0=HOR,1=VER) 

0. :Beam width in deg (this value is ignored for OMNI and USRDEF antenna) 

0. :Antenna elevation angle in deg(this value is ignored for OMNI
and USRDEF antenna) 

0 :Number of cut-back angles and factors (used for specific height-finder
antenna) 

0. :Minimum output height in m 

100. :Maximum output height in m 

1. :Maximum output range in km 

5 :Number of output height points 

100 :Number of output range points 

0 :Extrapolation flag

0. : Surface absolute humidity in g/m3 

0. : Surface air temperature in degrees 

0. : Gaseous absorption attenuation rate in dB/km 

0 : Number of wind speeds/ranges specified 

1 : Number of refractivity profiles 

1 : Number of levels in refractivity profiles 

0. : Range of first refractivity profile in km 

0. 209.2 : Height \& M-unit value of ref. profile 1, level 1 

1 : Number of ground composition types 

0., 0, 0., 0. : Range (km), ground type (integer), permittivity, conductivity 

0 : Number of terrain range/height points 

\caption{Fragment of the APM input file}
\label{Flo:APM input file}
\end{algorithm}

\begin{algorithm}[H]
...

{*}{*}{*}{*}{*}{*}{*}{*}Output Loss and Prop. Factor Values{*}{*}{*}{*}{*}{*}{*}

$\,$

range in km = 0.05

Height(m) Loss(dB) PFac(dB) 

20.00 67.80 -2.20

40.00 65.70 0.40

60.00 85.50 -17.80

80.00 69.10 0.20

100.00 78.70 -7.70

$\,$

range in km = 0.10

Height(m) Loss(dB) PFac(dB) 

20.00 73.70 -2.10

40.00 82.30 -10.60

60.00 74.60 -2.40

80.00 67.30 5.40

100.00 68.10 5.50 

...\caption{Fragment of the APM output file}

\label{Flo:APM output file}
\end{algorithm}

As a tool APM can be the part of the assessment means. It computes
the pattern propagation factor in the space between the source (radar)
and the wind farm.

When APM is launched, it reads the input file, makes the calculations
and writes the output file with initial parameters and the final result.
The result consists of the propagation factor and loss at the range,
height and resolution we want. For our needs we use only the information
about propagation factor.

In order to operate with the output of APM we need to create the tool
that will read and visualise propagation factor. Also it should prepare
the right format of data for the next steps, for example, to compute
RCS of wind turbine. Regarding the idea to automate the algorithm
of assessment means, we have chosen Matlab. 

\begin{figure}[H]
\centering{}\includegraphics[scale=0.3]{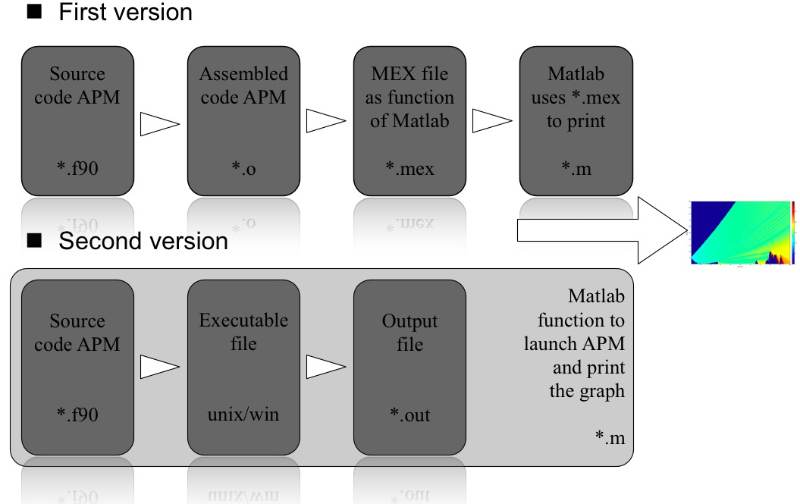}\caption{APM under Matlab}
\label{Flo:APM under Matlab}
\end{figure}

The main script makes a few steps:
\begin{enumerate}
\item Read an external data about the terrain of the wind farm area (profiles
of each wind turbine), an antenna pattern, refractivity profiles, 
\item Run the APM code written in Fortran.
\item Read the APM output.
\item Visualisation.
\item Create the readable file for the RCS program.
\end{enumerate}
We decide to run APM from Matlab environment (the second step). In
fact, there are two ways to do that. Matlab lets create special \emph{mex}
file that runs as its function. This file could be generated running
the compiled code of Fortran 90 by a special function of Matlab. This
solution was applied in ST\&I by another intern \cite{Intgrationdunmodledepropagationdondeauseinduneplate-formedesimulationRadar}.
However, the generated file operates only in the OS where it was created.
Another way is the possibility to use the OS command line from Matlab.
It means that we can compile and run APM directly from the script
or function. At the same time the performance of Fortran 90 code is
high. The difference between two variants to run APM under Matlab
is shown in Figure \ref{Flo:APM under Matlab}.

\begin{figure}[H]
\begin{centering}
\includegraphics[scale=0.3]{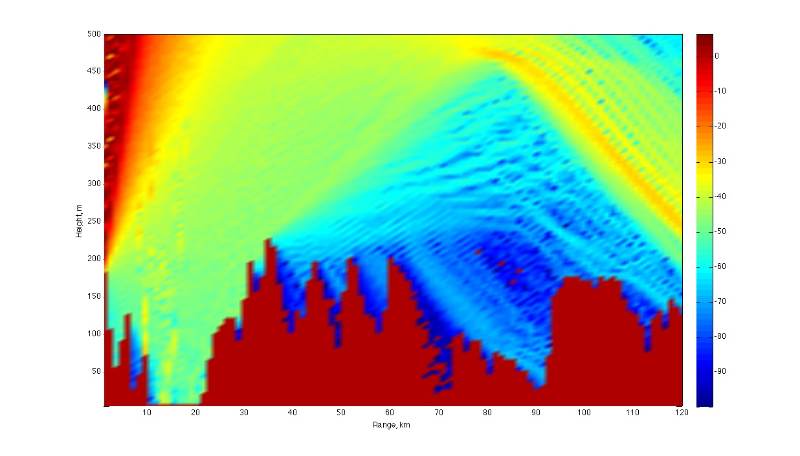}\caption{Example of the visualisation of the APM output}
\label{Flo:example of APM}
\par\end{centering}

\end{figure}

Steps 3 works with file.out (Algorithm \ref{Flo:APM output file}).
Matlab script loads this file and retrieves the PPF data. Afterwards
it stacks the information for the further visualisation. As a result
of the 4$^{th}$ step we receive the image like Figure \ref{Flo:example of APM}.
Here the abscissa and ordinate are the range and height respectively.
The colour bar refers to PPF.

\subsection{CAD Model of the Wind Turbine\label{sub:CAD-Model-of}}

In spite of the fact that the most precise model of a wind turbine
can be given only by its producer, we also can build such a model
based on specifications of the machine. The point is that some critical
parts of wind turbines are more or less standardised. Thus, the overall
dimensions and the dimensions of the nacelle are given in specifications.
The blade design can be found in the wind turbine airfoil catalogue\cite{bertagnolio-wind}
according to its identification number. 

Therefore, we are able to collect from Internet the basis of information.
Afterwards the reflectivity of the turbine can be analysed basing
on a few elements (Table \ref{Flo:Estimation of wind turbine materials}).

\begin{table}[H]
\begin{centering}
\begin{tabular}{|c|c|c|}
\hline 
External surface & Spar & Lightning protection\tabularnewline
\hline
\hline 
2 shells & More stronger material & Conductor wire at \tabularnewline
(internal/external)  & than shell material  & least 40 \emph{mm\texttwosuperior{}}\tabularnewline
\hline 
Variable thickness & Rectangular section  & With or without conductor \tabularnewline
 & and filleted edges  & shell at blade tip\tabularnewline
\hline 
Fibre reinforced  & Association with shell  & Association with shell \tabularnewline
composite material & transmission loss (D/R) & transmission loss (D/R) \tabularnewline
\hline
\end{tabular}
\par\end{centering}

\caption{Estimation of wind turbine materials }
\label{Flo:Estimation of wind turbine materials}
\end{table}

\begin{figure}[H]
\centering{}\subfloat[3D WT, nacelle]{\centering{}\includegraphics[scale=0.25]{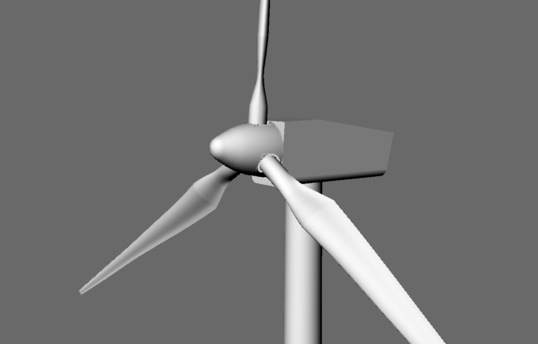}}\subfloat[3D WT from the bottom]{\centering{}\includegraphics[scale=0.25]{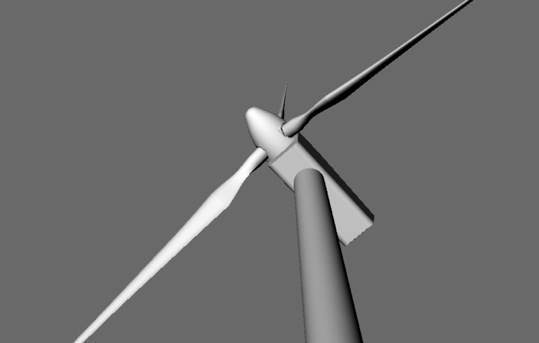}}\subfloat[3D WT without blades]{\centering{}\includegraphics[scale=0.22]{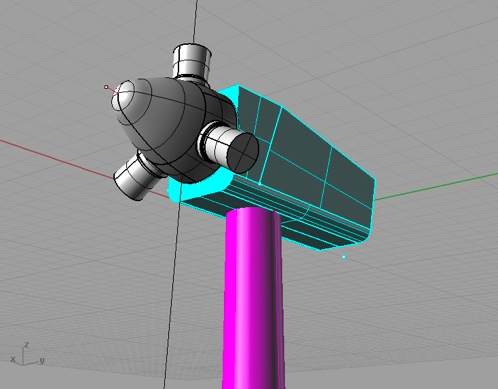}}\caption{Rhinoceros NURBS CAD models of wind turbine}
\label{Flo:Rhinoceros NURBS CAD models of wind turbine}
\end{figure}

When a data about the turbine is known, we can design its CAD model
(Figure \ref{Flo:Rhinoceros NURBS CAD models of wind turbine}). For
this step we use Rhinoceros NURBS CAD software.

In order to use CAD model for RCS computations it is necessary to
transform the data to another format. The code has the capability
to move the rotor (hub+blades) step by step, to change the pitches
of the blades and if needed also to orientate the nacelle in azimuth.

\subsection{Radar Cross Section \label{sub:Radar-cross-section}}

According to the IEEE dictionary of electrical and electronics terms
\cite{jay1984ieee}, RCS is a measure of reflective strength of a
target defined as 4$\pi$ times the ratio of the power per unit solid
scattered in a specified direction. More precisely, it is the limit
of that ratio as the distance from the scatterer to the point, where
the scattered power is measured approaches infinity \cite{knott-radar}:

\begin{equation}
\sigma=\underset{r\rightarrow\infty}{\lim}4\pi r^{2}\frac{|E^{scat}|^{2}}{|E^{inc}|^{2}}\end{equation}

where
\begin{quotation}
\emph{E$^{scat}$} and\emph{ E$^{inc}$} - the scattered electric
field and the field incident at the target respectively
\end{quotation}
ALBEDO is the code used at Thales Air Systems in ST\&I in high frequency
domain. It computes RCS and can run other software that rotate objects.
In this case RCS can be computed as a function of: 
\begin{itemize}
\item angles $\theta$ and $\varphi$ in wind turbine axis system;
\item radar frequency (S band);
\item position of the rotor: 8831 angles (1/3 of the round);
\item pitch law: typical blade pitch fixed angles. 
\end{itemize}
The code retrieves a polynomial surface from the file of ALBEDO format.
After ALBEDO analyses the surface in order to automatically find corner
reflectors and isolated areas. The code carries out the reflection
of the surface C1 (approximation using a physical optics). Also it
takes into account moving parts (the rotor of a helicopter, the screw
propeller) by means of mixed method driven by Generalised Ray Expansion
(GRE).

As a result ALBEDO gives tables with complex coefficients of the matrix
D associated with input position parameters. The output can be presented
in Matlab format.

\subsection{ASTRAD Module\label{sub:ASTRAD-Module}}

Architecture and Simulation Tool for Radar Analysis and Design (ASTRAD)
- or Atelier de Simulations Techniques Radars \& Auto-Directeurs in
French - \cite{guguen2008astrad} is a software developed at Thales
Air Systems.

The ASTRAD project was launched in 2002 for a four-year period of
development, with a release of its first version by mid-2006. Initially
the decision to design ASTRAD was taken by French MoD and Thales group. 

The ASTRAD software stands as a fully integrated development environment
(IDE) aiming at: 
\begin{itemize}
\item Proposing built-in solution, to create and run complex systems simulations
for the full benefit of system designers and operational end users. 
\item Providing a framework with libraries and complete applications to
internal and external partners, giving them the opportunity to focus
on their own added value, 
\item Promoting innovation and cooperation within a community (Agencies,
Laboratories, Industries) in accordance with Intellectual Properties
Rules. 
\end{itemize}
The ASTRAD platform can accommodate a wide range of computer and scientific
languages in a single simulation and that is exactly what we need
for our project. For instance, we use APM written in Fortran 90, RCS
simulation coded in C++ and middleware runs under Matlab.

ASTRAD may be used for the different steps of the development path.
Up to now, ASTRAD supports many computer languages like C / C++, Java
and Fortran 90 and scientific suites like Matlab and PV-Wave.

\begin{figure}[H]
\begin{centering}
\includegraphics[scale=0.3]{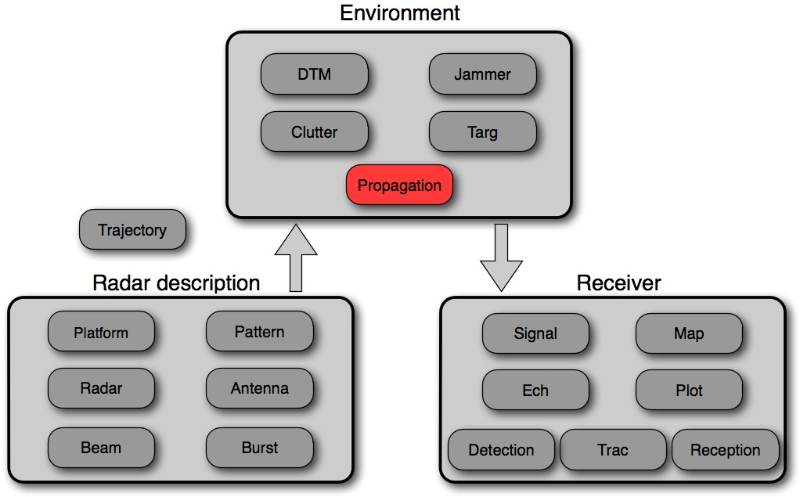}
\par\end{centering}

\caption{Block interface control}
\label{Flo:Block interface control}
\end{figure}

The default library (Figure \ref{Flo:Block interface control}) results
from the effort of a large panel of experts. It gathers a quite extended,
yet manageable, set of data types sufficient for a large majority
of radar developments. If needed, each user is entrusted the right
to modify those types or create a new base for his own interest. Regarding
our project the problem concerns the propagation module marked in
red. In the next Section \ref{sec:Solution} we use the data from
Radar description that goes to Environment block.

Thales Air Systems operates ASTRAD in several ways: 
\begin{itemize}
\item Upstream studies: ASTRAD simulations are exploited to define and validate
future radar time management concepts or Non-Cooperative Target Recognition
processes. 
\item Proposal: ASTRAD is used to evaluate radar performance (coverage,
accuracy, discrimination, etc.) in complex environment (clutter, jammer,
etc.) against several target types (aircraft, cruise missile, helicopter,
etc.).
\item Design: ASTRAD is used to define innovative Signal Processing Building
Blocks of the Thales surface radar family. 
\item Validation: a specific Analysis tool of Integration Verification Validation
and Qualification has been developed and deployed. A current application
is radar parameter tuning before the acceptance test. 
\end{itemize}
This report concerns the proposal and design. On one hand the assessment
of the wind farm impact on the radar deals with new environment and
with radar performance in this environment. On the other hand the
solution concerns the Signal Processing Building Block for the radar
STAR 2000.

\section{Solution\label{sec:Solution}}

The solution consists in method that uses described tools. As a result,
we need to know the wind farm impact in terms of RCS. This data should
be based on the information about the radar, environment and wind
farm.

If the objective is to choose the place to maintain the wind farm
or the radar, it is necessary to find the place, where turbines will
make the least impact on the radar. The further explanation is given
for the case when the place of the radar is known and we evaluate
the disposition of the wind farm. If we would examine the disposition
of the radar, the idea of an algorithm remains the same. The solution
presents the following algorithm:
\begin{enumerate}
\item Collect all the necessary data for further steps (Table \ref{Flo:data for solution}).%
\begin{table}[H]
\begin{centering}
\begin{tabular}{|c|c|c|}
\hline 
Radar & Environment & Wind farm\tabularnewline
\hline
\hline 
Frequency & Terrain profile & Number of turbines\tabularnewline
\hline 
Antenna height & Reflectivity of the surface & Height of turbines\tabularnewline
\hline 
Polarization & Refractivity profiles & Length of blades\tabularnewline
\hline 
Antenna elevation angle & Surface air temperature & Type of the airfoil\tabularnewline
\hline 
 & Windspeed & Revolution speed\tabularnewline
\hline 
 & Gaseous absorption & Number of blades\tabularnewline
\hline 
 & Surface absolute humidity & Angles relative to the radar\tabularnewline
\hline
\end{tabular}\caption{Input data}
\label{Flo:data for solution}
\par\end{centering}

\end{table}

\item Define the possible disposition of the wind farm or radar.
\item Compute an electromagnetic field in the area of every turbine for
the height from the surface up to its top. APM is used for this step,
though it calculates only the amplitude. In Chapter \ref{cha:Complex-pattern-propagation}
we give the solution to calculate the complex field that includes
the phase as well. 
\item Simulate RCS.
\item Choose the disposition of the wind farm, where its influence on the
radar is minimal.
\end{enumerate}
The example of this algorithm is given in Chapter \ref{cha:Application-of-the}.
Usually each wind farm is visible by radars using different wavelengths
and operation in different places. Despite possible solution for one
radar, it can be unfit for others or even makes worse. Nevertheless,
the common solution is to place the farm between the hills, where
it is more or less invisible for radars.

This algorithms could be implemented by the tool ASTRAD shown in Subsection
\ref{sub:ASTRAD-Module}.

\begin{figure}[H]
\begin{centering}
\includegraphics[scale=0.5]{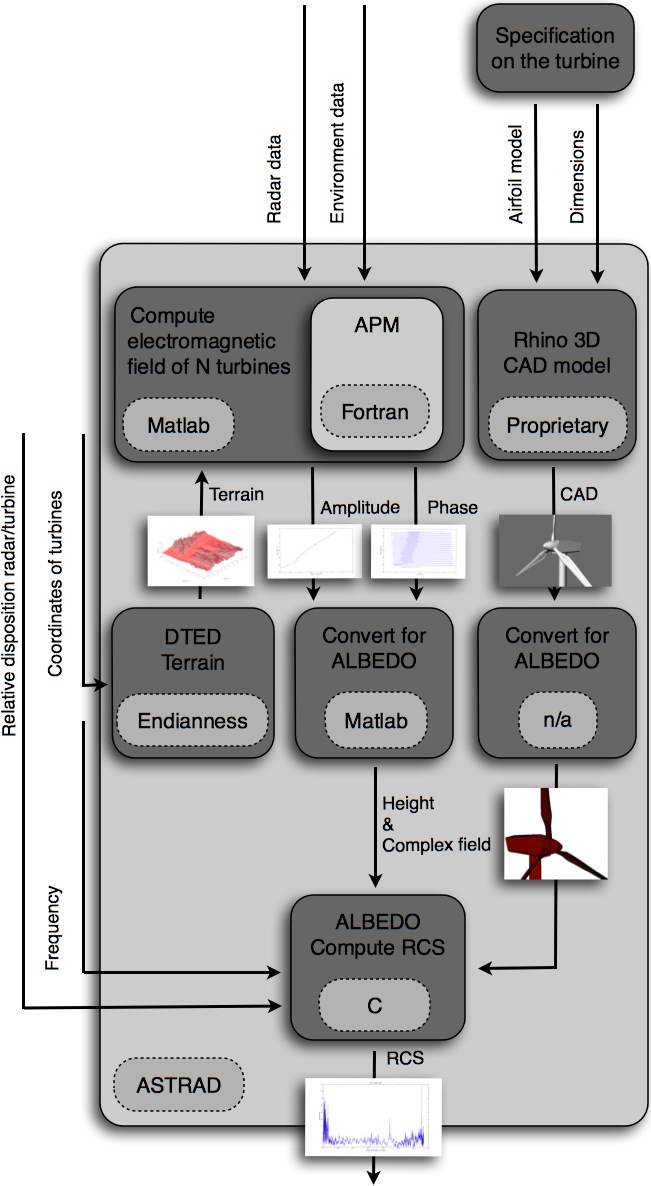}\caption{Diagram of the algorithm}
\label{Flo:Diagram of the algorithm}
\par\end{centering}

\end{figure}

In order to make things clearer Figure \ref{Flo:Diagram of the algorithm}
shows the scheme of the process.

\chapter{Complex Pattern Propagation Factor\label{cha:Complex-pattern-propagation}}

\section{Introduction}

In Subsection \ref{sub:Advanced-propagation-model} we examined APM
as a tool. In this Chapter we always deal with APM.

First of all one must say why we pay attention to the complex PPF.
By default APM calculates PPF based on the amplitude and it does not
take into account the phase of the wave. This is critical for the
next tool of assessment means - RCS (see Subsection \ref{sub:Radar-cross-section})
because in order to calculate RCS we need to have both an amplitude
and a phase of the electromagnetic wave. Having the source code of
APM we can understand its calculations and change the code in order
to have the necessary phase.

\begin{figure}[H]
\begin{centering}
\includegraphics[scale=0.35]{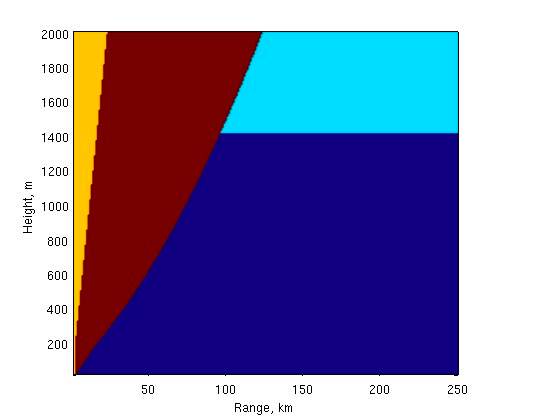}
\par\end{centering}

\caption{Zones of APM}
\label{Flo:apm zones}
\end{figure}

APM consists of four different propagation models - Flat Earth (FE),
Ray Optics (RO), Extended Optics (XO) and Parabolic Equations. APM
automatically calculates bounders of each model in the space and performs
them. However, there are only two models, which give PPF at low heights
- FE and PE. Therefore, it's obvious that the study of these two models
is sufficient within the framework of wind farm context. Close to
the surface FE works on the distance less than 2.5 km, while PE covers
further ranges. Typical boundaries of APM models are presented in
Figure \ref{Flo:apm zones}, where the yellow field - FE, brown -
RO, blue - PE and light blue - XO. Disposition of boundaries depends
on input parameters.

The principium of FE and RO is the same. It consists in the definition
of direct and reflected rays in each point of space. FE is simpler
than RO. In comparison with FE, RO uses more input parameters: the
refraction, terrain profile and some others. The conductivity and
permittivity of the terrain are taken into account in both models.
Two models have been validated by the evaluation of every step of
the APM code. Finally, the phase PPF was found.

The critical model, responsible for the propagation in the wind farm
area, is PE. The operation of PE differs completely from FE and RO
and is shown in details below.

The last model, XO, is the least precise. In fact, it uses the upper
layer of PE field and interpolates it to the top according to the
refraction profile. XO does not participate in propagation close to
the surface. This model is not considered here in details because
it is useless for the wind farm problem.

The module PE is examined in details because practically it covers
all ranges, where wind farms operates.

\section{Formulaic Definition}

In order to define the pattern propagation factor we cite the Radar
Handbook of Merrill Skolnik \cite{skolnik2008radar}. According to
the classical mathematical definition the pattern propagation factor,
for a point in space at a range \emph{R} and elevation angle $\theta$,
is the magnitude of the ratio of the electric field strength \emph{E(R,
$\theta$) }at that point (e.g., volts per meter) to the field strength
that would exist at the same range in free space and in the antenna
beam maximum (eqn. \ref{eq:PPF mathematical definition}).

\begin{equation}
F(R,\theta)=\frac{E(R,\theta)}{E_{0}(R)}\label{eq:PPF mathematical definition}\end{equation}

where 
\begin{quotation}
\emph{E$_{0}(R)$} - the beam-maximum free-space field strength at
range \emph{R}
\end{quotation}
To solve any reflection-interference problem, it is necessary to know
the value of the specular-reflection coefficient \emph{$\Gamma$ }of
the surface and the characteristics of the vertical-plane antenna
pattern, expressed as a pattern factor. The reflection coefficient
is a complex number of magnitude $\mathcal{\rho}$ and the phase angle
$\phi$. Similarly the pattern factor, a function of the vertical-plane
angle $\theta,$ has a magnitude |\emph{f($\theta$)}| and a phase
angle $\beta.$ In terms of these quantities, the general formula
of \emph{F} when multipath interference occurs is

\begin{equation}
F=|f_{d}+\rho f_{r}e^{-j\alpha}|\label{eq:PPF common definition}\end{equation}

where
\begin{quotation}
\emph{f$_{d}$} and \emph{f$_{r}$}- the magnitude of \emph{f($\theta$$_{d}$)
}and\emph{ f($\theta_{r}$) }respectively

$\alpha$ - the total phase difference of the direct and reflected
waves at their point of superposition, i.e., at the target of the
transmitted waves and at the receiving antenna for the returned echo
\end{quotation}
This total phase difference is the resultant of the phase difference
($\beta_{r}-\beta_{d}$) of the pattern factors, the phase shift $\phi$
that occurs in the reflection process, and the phase difference due
to the path-length difference. The absolute-value brackets indicate
that \emph{F} is a real number, although the reflection coefficient
$\Gamma$ and the pattern factor \emph{f($\theta$) }are in general
complex.

The propagation factor \emph{F} is often expressed in dB \cite{levy2000parabolic}.
If we approximate the distance \emph{d} between the terminals by range
\emph{x}, we get eqn. \ref{eq:PPF Levy}.

\begin{equation}
F(x,z)=20\log|u(x,z)|-10\log(x)-10\log(\lambda)\label{eq:PPF Levy}\end{equation}

where
\begin{quotation}
\emph{x} - range

\emph{z} - height

\emph{u} - complex field 
\end{quotation}
Anticipating things, one must say that the FE and RO use the total
phase difference as it is pointed out in eqn. \ref{eq:PPF common definition}.
The phase difference means the relative phase. In order to compute
RCS we need to have the absolute one.

On the other hand, the PE and XO calculate PPF using eqn. \ref{eq:PPF Levy}.
Moreover, \emph{u} contains the information about the phase.

The next section shows how to retrieve the absolute phase from important
models of APM.

\section{Extraction from Advanced Propagation Model}

\subsection{Flat Earth}

FE represents the simplified geometric model of a ray propagation
and its application has some limitations. FE operates in area, where
antenna elevation angles above 5 degrees or ranges less than approximately
2.5 km (see Figure \ref{Flo:apm zones}, the yellow field). As follows
from the term {}``Flat earth'', the model works only in the space
above the flat terrain. Otherwise it would compute the propagation
with marked errors. As a simple model FE does not take into account
many input parameters of APM, for example, the refractivity profile.
However this simplification has a reason. FE operates in a relatively
small space, where the effect of a refraction does not play a big
role. Also the model ignores surface effects that deviate a ray during
reflection.

\begin{figure}[H]
\begin{centering}
\includegraphics[scale=0.4]{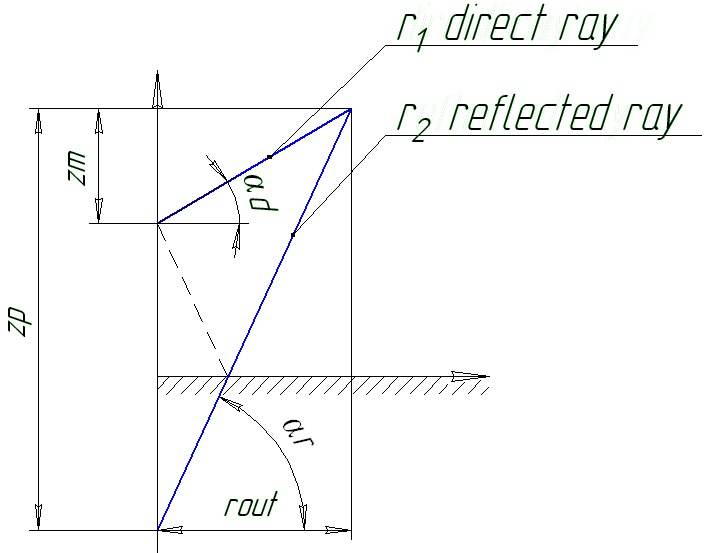}
\par\end{centering}

\caption{FE ray tracing}
\label{Flo:FE ray tracing}
\end{figure}

There are always two rays - direct and reflected. The model is shown
in Figure \ref{Flo:FE ray tracing}.

From the two path lengths \emph{r}$_{1}$ and \emph{r}$_{2}$, the
surface-reflection phase lag angle $\varphi$, and the free-space
wave number \emph{k}$_{0}$, the total phase angle is determined as
:

\begin{equation}
\Omega=(r_{1}-r_{2})k_{0}+\varphi\label{eq:total phase angle}\end{equation}

The planar travelling wave solution of the wave equations is:

\begin{equation}
E(r)=E_{0}e^{i\varphi}\label{eq:planar traveling wave}\end{equation}

The electromagnetic field of some point of the space (index \emph{p})
is a superposition of the direct and reflected waves (index \emph{d}
and \emph{r} respectively):

\begin{equation}
E_{p}e^{i\varphi_{p}}=E_{d}e^{i\varphi_{d}}+E_{r}e^{i\varphi_{r}}\label{eq:superposition}\end{equation}

\emph{E$_{d}$} and \emph{E$_{r}$} are determined according to the
antenna pattern factor of $\alpha_{d}$ and $\alpha_{r}$.

In order to compute the resultant amplitude it is enough to know the
phase difference $\Omega$:

\begin{equation}
|E_{p}e^{i\varphi_{p}}|^{2}=E_{d}^{2}+E_{r}^{2}+2E_{d}E_{r}\cos\Omega\label{eq:resultant amplitude  FE}\end{equation}

However, still we do not know the absolute resultant phase $\alpha_{p}.$
To retrieve the phase we can use:

\begin{equation}
\varphi_{p}=\arctan\left(E_{d}e^{ir_{1}k_{0}}+E_{r}e^{i(r_{2}k_{0}+\Omega)}\right)\label{eq:absolute resultant phase}\end{equation}

Eqn. \ref{eq:resultant amplitude  FE} is used by APM to calculate
the propagation factor. It does not operate with absolute phase that
needed for complex pattern propagation factor. With eqn. \ref{eq:absolute resultant phase}
we have both an amplitude and phase. Than it is possible to compute
the complex pattern propagation factor.

The APM subroutine that calculates FE is called \emph{fem}. In terms
of a code the solution is given in Algorithm \ref{Flo:Phase of FE model}.

\begin{algorithm}[H]
! Now get total phase lag and compute propagation factor and loss.

phdif = ( r2 - r1 ) {*} fko + rphase

frterm = facr {*} rmag 

ffac2 = facd{*}facd + frterm{*}frterm + 2. {*} facd {*} frterm {*}
dcos(phdif) 

~

!Added code begins 

ffac2 = atan(aimag(facd{*}exp(qi{*}r1{*}fko)+frterm{*}exp(qi{*}(r2{*}fko
+ rphase)) , real(facd{*}exp(qi{*}r1{*}fko)+frterm{*}exp(qi{*}(r2{*}fko
+ rphase))))

!Added code ends\caption{Phase of the Flat Earth model}

\label{Flo:Phase of FE model}
\end{algorithm}

where
\begin{quotation}
\emph{rmag} - magnitude of the reflection coefficient 

\emph{ffac2} - amplitude or phase of the complex field
\end{quotation}
The application of FE in our study is very limited. We mentioned that
APM uses FE only for short distances - 2.5 \emph{km} and less. In
reality there are no wind turbines situated very close to the radar.
But still if we deals with such a situation, PE can not give us the
satisfactory result. By the way, this is one of the reasons why APM
does not use only PE - it does not work for the short range.

\subsection{Ray Optics}

Basically RO is quite similar to FE but has a few qualitative differences.
Within the RO region (see Figure \ref{Flo:apm zones}, the brown field),
the propagation factor is calculated from the mutual interference
between direct-path and surface-reflected ray components using the
refractivity profile at a zero range. Full account is given to focusing
or de-focusing along both direct and reflected ray paths and to the
integrated optical path length difference between the two ray paths,
to give precise phase difference, and, hence, accurate coherent sums
for the computation of propagation loss \cite{barrios2002advanced}.
It means that in comparison with FE, RO considers the terrain, refractivity
profile (partly), parameters of the surface, etc. That makes RO computations
more complicated. RO still does not concern the effect of the diffraction.

\begin{figure}[H]
\centering{}\includegraphics[scale=0.3]{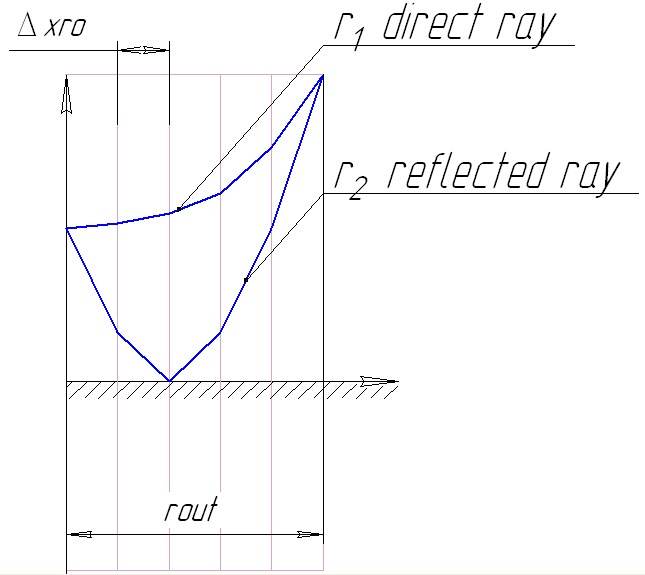}\caption{RO ray tracing}
\label{Flo:RO ray tracing}
\end{figure}

RO seems to be useless for our purposes because it does not compute
the field close to the surface. There are two main subroutines in
APM that compute a propagation factor. \emph{Rocalc} provides characteristics
of direct and reflected rays. And \emph{roloss} calculates the final
field using the result of \emph{rocalc} and stores propagation factor.

\subsection{Parabolic Equations\label{sub:Parabolic-Equations}}

\subsubsection{Theory of PE\label{sub:Theory-of-PE}}

Theory of parabolic equations is perfectly described in \cite{levy2000parabolic}.
We cite a few basic ideas from this book about the theory, which will
help us to understand how PE operates in APM.

The propagation of electromagnetic field is associated with the paraxial
direction \emph{x}:

\begin{equation}
u(x,z)=e^{-ikx}\psi(x,z)\label{eq:propagation of electromagnetic field}\end{equation}

The point of using this reduced function is that it is slowly varying
in range for energy propagation at angles close to the paraxial direction,
which gives it convenient numerical properties.

The scalar wave equation in terms of \emph{u} is:

\begin{equation}
\frac{\partial^{2}u}{\partial x^{2}}+2ik\frac{\partial u}{\partial x}+\frac{\partial^{2}u}{\partial z^{2}}+k^{2}(n^{2}-1)u=0\label{eq:scalar wave equation}\end{equation}

This can be formally factored as:

\begin{equation}
\left\{ \frac{\partial}{\partial x}+ik(1-Q)\right\} \left\{ \frac{\partial}{\partial x}+ik(1+Q)\right\} u=0\label{eq:factored scalar wave equation}\end{equation}

where the pseudo-differential operator \emph{Q} is defined by:

\begin{equation}
Q=\sqrt{\frac{1}{k}\frac{\partial^{2}}{\partial z^{2}}+n^{2}(x,z)}\label{eq:pseudo-differencial operator}\end{equation}

Pseudo-differential operators are constructed from partial derivatives
and ordinary functions of the variables. A formal mathematical framework
is required to give a precise meaning of the square-root symbol in
the expression of \emph{Q}. The square root corresponds to the composition
of operators, in the sense that:

\begin{equation}
Q(Q(u))=\frac{1}{k^{2}}\frac{\partial^{2}u}{\partial z^{2}}+n^{2}u\label{eq:composition of operators}\end{equation}

must be satisfied for all functions \emph{u} in a certain class. The
construction of the appropriate square-root symbol is linked to the
class of functions \emph{u} on which it operates, and this in turn
depends on the boundary conditions for the partial differential equations
given by eqn. \ref{eq:scalar wave equation}. More generally, we shall
assume that \emph{Q} can be defined unambiguously and that expansions
for the ordinary square root function can be applied to \emph{Q}.

There are errors inherent in the factorisation given in eqn. \ref{eq:factored scalar wave equation}:
if the refractive index \emph{n} varies with range \emph{x}, the operator
\emph{Q} does not commute with the range derivative and the factorisation
is incorrect. Hence some care must be taken in the applications to
make sure that the resulting error remains small.

The next step is to split the wave equation into the two terms defined
by eqn. \ref{eq:factored scalar wave equation} and to look at functions
satisfying one of the resulting pseudo-differential equations:

\begin{equation}
\frac{\partial u}{\partial x}=-ik(1-Q)u\label{eq: forward propagation wave}\end{equation}

\begin{equation}
\frac{\partial u}{\partial x}=-ik(1+Q)u\label{eq:back propagation wave}\end{equation}

Eqns. \ref{eq: forward propagation wave} and \ref{eq:back propagation wave}
correspond respectively to forward and back propagation waves. In
a ray picture, forward propagation corresponds to rays propagating
with increasing \emph{x}, and back forward propagation to rays propagating
with decreasing \emph{x}. Eqns. \ref{eq: forward propagation wave}
and \ref{eq:back propagation wave} are the outgoing and incoming
parabolic wave equations.

In a range-independent medium, where there are communicator problems
with the factorisation of eqn. \ref{eq:factored scalar wave equation},
a solution of either eqn. \ref{eq: forward propagation wave} or \ref{eq:back propagation wave}
will automatically satisfy the original reduced wave equation, eqn.
\ref{eq:scalar wave equation}. However such a solution does not in
general correspond to the actual electromagnetic field. For example
a solution of the outgoing eqn. \ref{eq: forward propagation wave}
neglects the backscattered field. In order to get exact solution of
eqn. \ref{eq:scalar wave equation}, eqns. \ref{eq: forward propagation wave}
and \ref{eq:back propagation wave} should be solved simultaneously
in the coupled system:

\begin{equation}
\begin{cases}
u=u_{+}+u_{-}\\
\frac{\partial u_{+}}{\partial x}=-ik(1-Q)u_{+}\\
\frac{\partial u_{-}}{\partial x}=-ik(1-Q)u_{-}\end{cases}\label{eq:coupled system}\end{equation}

The approximation we make by solving for each term separately is a
paraxial approximation: for example, for the outgoing parabolic wave
equation, we solve for energy propagating in a paraxial cone centred
on the positive \emph{x}-direction.

Eqns. \ref{eq: forward propagation wave} and \ref{eq:back propagation wave}
are pseudo-differential equations of first order in \emph{x} (hence
the {}``parabolic'' terminology). They can be solved by marching
techniques, given in the field of the initial vertical and the boundary
conditions at the top and the bottom of the domain. The outgoing parabolic
wave equation, eqn. \ref{eq: forward propagation wave}, has the formal
solution:

\begin{equation}
u(x+\triangle x,\,.)=e^{ik\triangle x(-1+Q)}u(x,\,.)\label{eq:PE formal solution}\end{equation}

The forward propagation field is obtained at a given range from the
field at a previous range, and appropriate boundary conditions at
the top and bottom of the domain, in other words the solution is marched
in range. The computation gain is substantial compared to the elliptic
wave equation, which is of second order in both \emph{x} and \emph{z}
and mist be solved simultaneously at all points of the integration
domain.

The splitting of the wave equation into two paraxial terms implies
that only energy propagating inside a paraxial cone can be modelled.
This limits the type of propagation which can be represented accurately.
However the paraxial representation is very accurate for a number
of problems and also has substantial computational advantages.

The PE derived in this section provides a paraxial approximation of
the two-dimensional scalar wave equation. Naturally the scalar framework
is no longer adequate for general three-dimensional problems, where
polarisation aspects require a vector description. In that case scalar
PE have to be written for the electromagnetic field components and
coupled through boundary conditions at interfaces and the divergence-free
conditions.

It is useful to look through a staircase terrain modelling. The theory
of this model is applied in APM. Let the terrain is represented as
a sequence of linear segments. The coordinate system measures height
from the terrain. In other words we define new range and height variables
by:

\begin{equation}
\begin{cases}
\xi=x\\
\zeta=z-h(x)\end{cases}\label{eq: range and height variables}\end{equation}

where
\begin{quotation}
\emph{h(x) }- terrain height
\end{quotation}
Assume the terrain has slope $\alpha$ on segment \emph{$x_{1}\leq x\leq x_{2}$.
}Then we have in the corresponding vertical slice:

\begin{equation}
\begin{cases}
\xi=x\\
\zeta=z-h(x_{1})-\alpha(x-x_{1})\end{cases}\label{eq:vertical slice}\end{equation}

We now look at the new function $\nu$ defined by

\begin{equation}
\nu(\xi,\zeta)=e^{ik\alpha\zeta}u(x,z)\label{eq:new function V}\end{equation}

This function compensates for the wavefront shift relative to the
terrain. We have:

\begin{flushleft}
\begin{equation}
\begin{array}{c}
\frac{\partial u}{\partial x}=e^{ik\alpha\zeta}\left\{ \frac{\partial\nu}{\partial\xi}-\alpha\frac{\partial\nu}{\partial\zeta}\right\} \\
\frac{\partial u}{\partial z}=e^{ik\alpha\zeta}\left\{ \frac{\partial\nu}{\partial\zeta}+ik\alpha\nu\right\} \\
\frac{\partial^{2}u}{\partial z^{2}}=e^{ik\alpha\zeta}\left\{ \frac{\partial^{2}\nu}{\partial^{2}\zeta}+2ik\alpha\nu\right\} \end{array}\label{eq:after wavefront shift}\end{equation}

\par\end{flushleft}

Hence the standard PE becomes in terms of $\nu$:

\begin{equation}
\frac{\partial^{2}\nu}{\partial\zeta^{2}}+2ik\frac{\partial\nu}{\partial\xi}+k^{2}(n^{2}-1)\nu=0\label{eq:PE in terms of V}\end{equation}

We are back to the usual PE, but we now solve for a function which
follows the terrain by applying appropriate angle shifts on successive
segments. Implementation with a sine transform is straightforward.
We denote by\emph{ (r$_{0}$, h$_{0}$), (r$_{1}$, h$_{1}$),.}..
the terrain break points and by $\alpha_{1},\,\alpha_{2},..$ the
successive terrain slopes. Suppose we have computed the field at range
\emph{r$_{m-1}.$}We then march the solution to range \emph{r$_{m}$}
with the sequence of operation below.
\begin{enumerate}
\item First we shift the wave front, putting \begin{equation}
\nu_{m}(r_{m-1},\zeta)=u(r_{m-1},h_{m-1}+\zeta)e^{-ik\alpha_{m}\zeta}\label{eq:PE terrain 1}\end{equation}

\item We now propagate $\nu_{m}$ to range \emph{r$_{m}$} using the sine
transform method. On interval $\left[r_{m-1},\, r_{m}\right]$, the
standard PE algorithm is given by:\begin{equation}
\nu_{m}(\xi+\triangle\xi,\zeta)=e^{-\frac{ik(n^{2}-1)}{2}\triangle\xi}\mathcal{S}^{-1}\left\{ e^{-\frac{2i\pi^{2}p^{2}}{k}\triangle x}\mathcal{S}\left\{ \nu_{m}(xi,\zeta')\right\} \right\} \label{eq:PE terrain 2}\end{equation}
This propagates the field along the \emph{m$^{th}$ }segment, automatically
enforcing the boundary condition that the field should be zero at
the ground.
\item At range \emph{r}$_{m}$ we go back to \emph{u} with the formula:\begin{equation}
u(r_{m},\, h_{m}+\zeta)=\nu_{m}(r_{m},\,\zeta)e^{ik\alpha_{m}\zeta}\label{eq:PE terrain 3}\end{equation}

\item To continue the process, we shift the wave front again. In terms of
$\nu_{m}$, the next function $\nu_{m+1}$ is given by:\begin{equation}
\nu_{m+1}(r_{m},\, h_{m}+\zeta)=\nu_{m}(r_{m},\,\zeta)e^{ik(\alpha_{m}-\alpha_{m+1})\zeta}\label{eq:PE terrain 4}\end{equation}

\end{enumerate}
Propagation angles are now measured relative to the sloping terrain
in each successive slice. From eqn. \ref{eq:new function V}, we see
that if we want the solution \emph{u} to represent propagation angles
up to $\theta_{max}$ accurately, then $\nu$ must represent propagation
angles up to $\theta_{max}+\alpha_{max}$, where $\alpha_{max}$ is
the maximum terrain slope modulus. As a result the vertical grid resolution
can increase substantially.

This algorithm is applied in APM. The idea is that if we can completely
understand how PE operates in APM, we can clearly define the phase
and retrieve the necessary data. In our case this is the phase of
electromagnetic wave.

\subsubsection{PE in APM}

In APM PE can be chosen for all the space. This mode is called {}``PE
only''. We use this to analyse PE model. According to the code, APM
divides the space into vertical layers. The one dimension matrix corresponds
to each vertical layer. This vector contains the field from zero to
the maximum height. At every step, APM propagates the previous layer
to the next range taking into account the refractivity and terrain
profile.

Firstly, APM starts a module computing the first layer for the propagation
at a zero distance for all heights (eqn. \ref{eq:zero-distance propagator}).
The subroutine of APM, responsible for these calculation, is called
\emph{xyinit}.

\begin{equation}
U_{j}=c_{a}s_{gain}\left[f\left(\alpha_{d}\right)e^{-ip_{j}antk_{0}}-f\left(-\alpha_{d}\right)e^{ip_{j}antk_{0}}\right]\,,\, H\, pol\label{eq:zero-distance propagator}\end{equation}

\[
U_{j}=c_{a}s_{gain}\left[f\left(\alpha_{d}\right)e^{-ip_{j}antk_{0}}+f\left(-\alpha_{d}\right)e^{ip_{j}antk_{0}}\right]\,,\, V\, pol\]

\[
\alpha_{d}=\arcsin\left(p_{j}\right)\]

\[
c_{a}=\left(1-p_{j}^{2}\right)^{\nicefrac{3}{4}}\]

where
\begin{quotation}
\emph{s$_{gain}$= $\nicefrac{\sqrt{\lambda}}{z_{max}}$ - }the normalisation
factor, where $\lambda$- wave length, \emph{z$_{max}$} - total height

$\alpha_{d}$ and $-\alpha_{d}$ - the antenna pattern factors for
the direct path and for the reflected path

\emph{p$_{j}$ = j\ensuremath{\Delta} $\theta$ }- where $\Delta\theta$
- the angle difference between mesh points in p-space, \emph{j} -
index from 0 to transform size

\emph{antk$_{0}$ = k}$_{0}\, ant_{ht}$ - a height-gain value at
the source, where \emph{ant$_{ht}$} is the transmitting antenna height
above the local ground, \emph{k$_{0}$ - }wave number

\emph{i} - imaginary unit
\end{quotation}
The free-space propagator phase array is defined as well. It is calculated
once and will be used to propagate every further layer (eqn. \ref{eq:free-space propagator}).

\begin{equation}
frsp_{j}=f_{norm}e^{i\Delta r_{PE}(\sqrt{k_{0}-(j\Delta p)^{2}}-k_{0})}\label{eq:free-space propagator}\end{equation}

where 
\begin{quotation}
\emph{f$_{norm}$} - the Fourier transform normalisation constant

$\Delta r_{PE}$- PE range step

$\Delta p$ = $\nicefrac{\pi}{z_{max}}$- the angle (or p-space) mesh
size
\end{quotation}
After the first layer has been initialised, the propagation begins
for the next one. A current layer is assigned as a previous one. The
purpose of the next step is to propagate a complex PE solution in
a free space by a one range step. Upon entry the PE field is transformed
to the p-space (Fourier space) and its array elements are multiplied
by corresponding elements in the free-space propagator array given
by eqn. \ref{eq:free-space propagator}. Finally, the PE field is
transformed back to a z-space \cite{barrios2002advanced}. 

After the free-space propagation, an information about terrain profile
is used. This step is quite simple. The PE field is shifted up or
down at the difference in heights between current and previous ranges.

Finally, the field is multiplied by a function, based on the refractivity
profile, and given by eqn. \ref{eq:U refractivity}.

\begin{equation}
U=Ue^{i\Delta r\, profint}\label{eq:U refractivity}\end{equation}

\emph{Profint} is a one dimension matrix that presents the refraction
from zero to maximum height. This is an interpolated original refraction
multiplied by a wave number.

Besides these critical steps there is a number of sub-functions, which
generally deal with input parameters: there are some differences in
the algorithm for V and H polarisation, for different antenna types,
for the presence of wind, air temperature, the Earth flattering transformation
takes place, etc. However, the main algorithm remains the same.

Considering our goal, there are critical steps in the PPF computation
that were analysed with both visualisation and explanation.

\begin{figure}[H]
\begin{centering}
\subfloat[Isotropic space]{\begin{centering}
\includegraphics[scale=0.25]{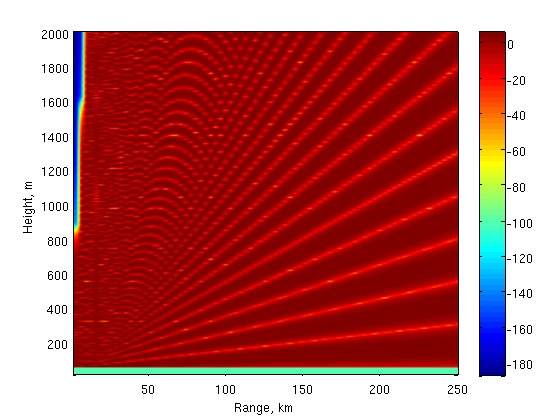}
\par\end{centering}

\label{Flo:fig 2 a}}\subfloat[Non-isotropic space]{\begin{centering}
\includegraphics[scale=0.25]{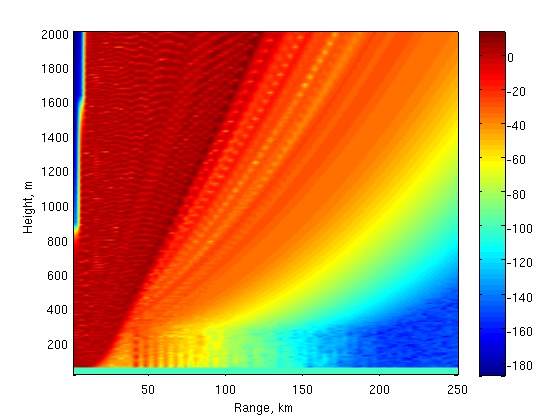}
\par\end{centering}

\label{Flo:fig 2 b}}\subfloat[Non-isotropic space, duct effect]{\begin{centering}
\includegraphics[scale=0.25]{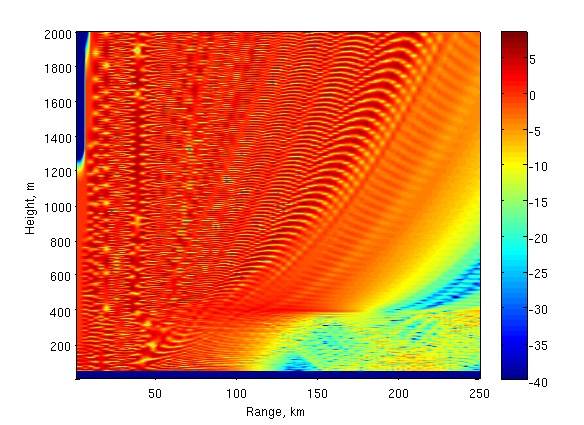}
\par\end{centering}

\label{Flo:fig 2 c}}
\par\end{centering}

\caption{PPF}
\label{Flo:refractivity}
\end{figure}

In Figure \ref{Flo:refractivity}, one can see the influence of the
refraction of non-isotropic space in comparison with the free space.
There are two ways to prove that the PE field does not take into account
the ray trace and differs from FE and RO:
\begin{enumerate}
\item Check the code in order to find calculations concerning the ray trace.
\item Change input parameters to see the influence in an output.
\end{enumerate}
To the first point, the PE source code with its subroutines does not
have functions performing the ray trace.

Concerning the second point, one can show the influence of the refraction
playing with input parameters. The terrain has the same height (50
\emph{m}) and same conductivity and permittivity for every example.
Results show that the difference is only due to the \textit{profint}
matrix - the refraction of space (eqn. \ref{eq:U refractivity}),
especially it is clear with the duct effect%
\footnote{Example from p. 90, Levy, M., Parabolic Equation Methods for Electromagnetic
Wave Propagation, IEE, London, UK (2000).%
} (Figure \ref{Flo:fig 2 c}). 

The refraction of these examples is given in Table \ref{Flo:Ref is non-is}.

\begin{table}[H]
\begin{centering}
\begin{tabular}{|c|c|c|c|}
\hline 
Height, \emph{m} & Isotropic, \emph{M-unit} & Non-isotropic, \emph{M-unit} & Non-isotropic, duct effect, \emph{M-unit}\tabularnewline
\hline 
0 & 330 & 330 & 330 \tabularnewline
\hline 
100 & 330 & 430 & 370 (300 \emph{m})\tabularnewline
\hline 
230 & 330 & 530 & 320 (400 \emph{m})\tabularnewline
\hline 
2000 & 330 & 630 & 500\tabularnewline
\hline
\end{tabular}
\par\end{centering}

\caption{Refraction of isotropic and non-isotropic spaces}
\label{Flo:Ref is non-is}
\end{table}

Moreover, the parameters of terrain such as conductivity and permittivity
do not impact on the propagation. These parameters do not participate
in calculations, and we will receive the same PPF with different values.

In the presence of terrain, initially, APM sets the height of the
current layer to zero even if there is the altitude. The current layer
is assumed as a previous one. Subsequently, APM applies a free-space
propagation. Then the result is shifted according to the terrain height
derivative. Finally, the refraction is considered. 

\begin{figure}[H]
\begin{centering}
\subfloat[Free space]{\begin{centering}
\includegraphics[scale=0.25]{Free-space}
\par\end{centering}

\label{Flo:fig 3 a}}\subfloat[With the terrain, {}``shift'' function commented]{\begin{centering}
\includegraphics[scale=0.25]{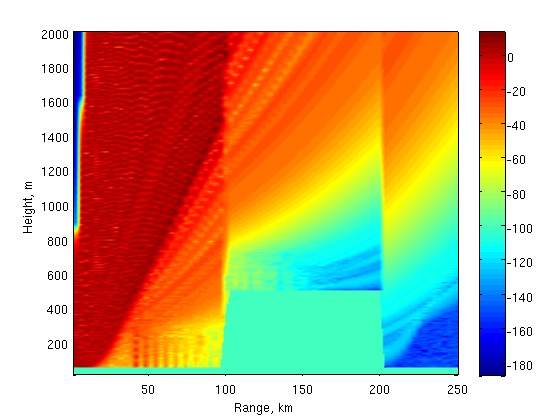}
\par\end{centering}

\label{Flo:fig 3 b}}\subfloat[Final result of APM]{\centering{}\includegraphics[scale=0.25]{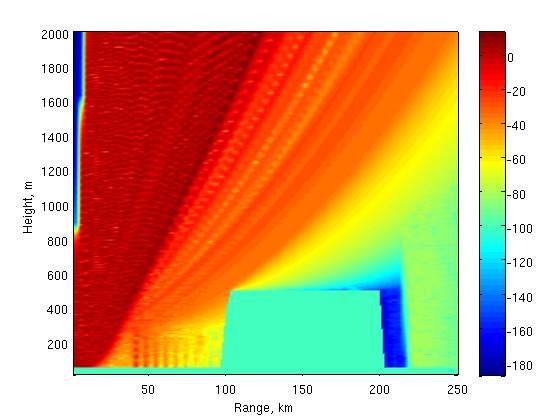}\label{Flo:fig 3 c}}
\par\end{centering}

\caption{Influence of the terrain}
\label{Flo: fig 3}
\end{figure}

Figure \ref{Flo:fig 3 b} shows PPF calculated without the shift (the
code was commented). One can see that the field on the terrain in
Figure \ref{Flo:fig 3 b} is the same as in Figure \ref{Flo:fig 3 a}
and almost the same as in Figure \ref{Flo:fig 3 c} but shifted down
by the height of this terrain. Nevertheless, there is an important
difference (Figure \ref{Flo:fig 3 b} - \ref{Flo:fig 3 c}). As a
matter of fact, the current layer, based on the field of previous
one (which contains the terrain), is multiplied by a free-space propagator
(see eqn. \ref{eq:free-space propagator}), which initially does not
have the terrain data. It leads to the final result that presents
an effect of diffraction as shown in Figure \ref{Flo:fig 3 c}.

Logically and mathematically this algorithm is presented in a literature
and shown in Subsubsection \ref{sub:Theory-of-PE}.

In contrast to other models of APM, PE exactly loses the phase when
calculates PPF, while other models do not calculate the phase at all.
It means that, during calculations, APM operates with PE field, which
is the array of complex numbers. In order to compute PPF, it takes
the absolute value (magnitude) of the complex field (eqn. \ref{eq:absolute U}),
interpolating it afterwards.

\begin{equation}
U=|U|\label{eq:absolute U}\end{equation}

Finally, APM calculates the pattern propagation factor using the usual
known equation, presented in different publications (eqn. \ref{eq:PE factor}).

\begin{equation}
F_{dB}=20\lg U+10\lg r\label{eq:PE factor}\end{equation}

where 
\begin{quotation}
\emph{r }- range where \emph{U} is calculated
\end{quotation}
Thus, one can easily retrieve the phase of PE field before the magnitude
is computed (eqn. \ref{eq:absolute U}). This phase is the angle between
real and imaginary part of \emph{U}. There should be eqn. \ref{eq:PE atan}
instead of eqn. \ref{eq:PE factor} in APM source code:

\begin{equation}
P=\arctan\left(\frac{\Im(U)}{\Re(U)}\right)\label{eq:PE atan}\end{equation}

where
\begin{quotation}
$\Im(U)$ and $\Re(U)$- imaginary and real parts of \emph{U=}$\Re+i\Im$
respectively
\end{quotation}
In that way P and F$_{dB}$ have the same resolution and dimension
that gives us the phase and amplitude of exactly the same point of
space.

At the level of the code the solution is given in Algorithm \ref{Flo:Phase of PE model}.

\begin{algorithm}[H]
u0=u(nb) 

u1=u(nbp1)

~

pmag0 = abs(u0) 

pmag1 = abs(u1) 

~

pmag = pmag0 + fr {*} (pmag1 - pmag0)

pmag = dmax1( pmag, pmagmin )

~

getpfac = 20.{*}dlog10( pmag ) + rlog

~

!Added code begins

getpfac = pmag !This is an amplitude of complex field 

getpfac = atan2( aimag(u0 + fr {*} (u1 - u0)), real(u0 + fr {*} (u1
- u0))) !This is a phase of complex field

!Added code ends\caption{Phase of the Parabolic Equations}

\label{Flo:Phase of PE model}
\end{algorithm}

where
\begin{quotation}
\emph{pmag} - interpolated magnitude of field

\emph{u1} and \emph{u2} - complex field at bin directly below and
above desired height

\emph{fr} - interpolation fraction

\emph{getpfac} - amplitude or phase of complex field
\end{quotation}
There are a few function in APM for PE. The main ones are \emph{getpfac}
and \emph{calclos}.

\subsection{Extended Optics}

XO performs a ray trace on all rays within one output range step and
returns the propagation loss up to the necessary height, storing all
angles and heights. It calculates loss values in the height region
above the maximum height of the PE model. APM uses only the XO and
PE models if the terrain profile is not flat for the first 2.5 \emph{km}
and if the antenna height is less than or equal to 100 \emph{m}. 

XO seems to be useless for our purposes because it does not compute
the field close to the surface as well as RO. \emph{Exto}, \emph{xeinit}
and \emph{xostep }are subroutines, which calculate XO.

\chapter{Application of the Assessment Means\label{cha:Application-of-the}}

\section{Introduction}

Previous Chapters give us a methodology. Here we apply the scenario,
where the radar operates in the presence of the wind farm. All the
data and information are real. Thus, there is an ATC radar STAR-2000,
Beinn Tharsuinn wind farm with 17 turbines, and environment in the
UK. In Section \ref{sec:Input-Parameters} we give the parameters
for the further simulation (Section \ref{sec:Implementation}). In
fact the Chapter is based on the solution given in Chapter \ref{cha:Assessment-means}.
Finally, we examine different scenarios of the wind farm disposition
and give the most suitable for the radar.

\section{Input Parameters\label{sec:Input-Parameters}}

\subsection{Wind Farm}

\begin{figure}[H]
\centering{}\includegraphics[scale=0.3]{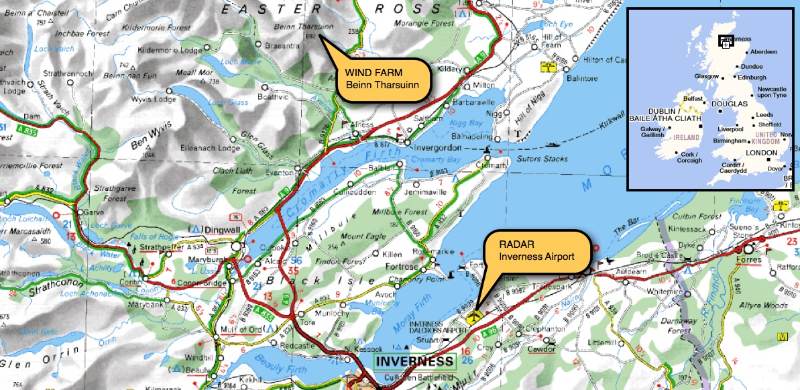}\caption{Wind farm and Star 2000}
\label{Flo:Wind farm and Star 2000}
\end{figure}

For out example we take a real wind farm that runs in United Kingdom.
The farm is situated at 32 \emph{km} from the ATC radar, which operates
in the Inverness airport (Figure \ref{Flo:Wind farm and Star 2000}).
The wind farm is located almost on the top of the Beinn Tharsuinn
hill. It means that the farm is good visible from the radar and negatively
acts on its signal processing. The common information about this wind
farm is given in Subsection \ref{sub:Others} in Table \ref{Flo:BWEA operational}.

\begin{figure}[H]
\centering{}\includegraphics[scale=0.3]{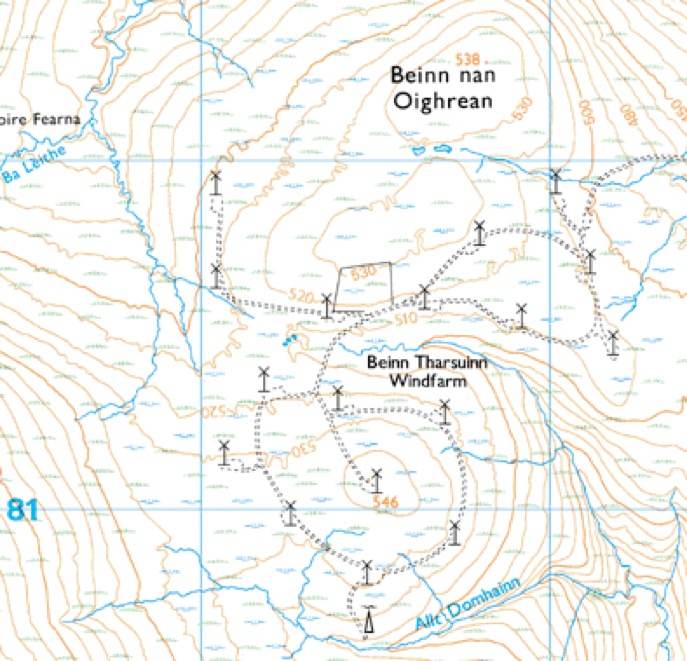}\caption{Beinn Tharsuinn farm with 17 wind turbines}
\label{Flo:17 wind turbines}
\end{figure}

The wind farm consists of 17 turbines Vestas V66 (Figure \ref{Flo:17 wind turbines})
\cite{BeinnTharsuinnEnvironmentalStatementNon-technicalSummary}.

\begin{figure}[H]
\centering{}\includegraphics[scale=0.4]{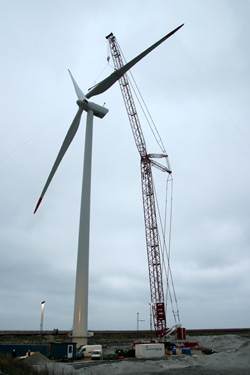}\caption{Vestas V66}
\label{Flo:Vestas V66}
\end{figure}

Vestas V66 is the model for inland locations (Figure \ref{Flo:Vestas V66}). 

In order to show the possible influence on the radar signal processing
we use some technical information about V66 \cite{V661.75MWV662.0MWOFFSHORE}
(Table \ref{Flo:Info Vestas V66}).

\begin{table}[H]
\begin{centering}
\begin{tabular}{|r||r|}
\hline 
Diameter & 66 \emph{m}\tabularnewline
\hline 
Area swept & 3421 \emph{m$^{2}$}\tabularnewline
\hline 
Revolution speed & 21.3 \emph{rpm}\tabularnewline
\hline 
Operational interval & 10.5-24.5 \emph{rpm}\tabularnewline
\hline 
Number of blades & 3\tabularnewline
\hline 
Hub height & 67 \emph{m}\tabularnewline
\hline 
Total weight & 197 \emph{t}\tabularnewline
\hline
\end{tabular}\caption{Technical information about Vestas V66}
\label{Flo:Info Vestas V66}
\par\end{centering}

\end{table}

One can find the velocity of the blade's top point at a normal revolution
speed (eqn. \ref{eq:Angular velocity}):

$V=\omega r=2\pi fr=2\cdot\pi\cdot(\frac{21.3}{60})\cdot33\thickapprox73.6\unitfrac{m}{s}=265\unitfrac{km}{h}$

The radial airplane speed could be close to the velocity of the blade.

Also dimensions of V66 are comparable with the airplane. Therefore,
the value of RCS can have the same order. The design of blades is
quite similar to plane's wings.

For our calculations we take 17 turbines of the Beinn Tharsuinn wind
farm. The only data we need are the distance of the wind turbine from
the source and its azimuth. It's enough having coordinates of the
radar. 

\begin{table}[H]
\centering{}\begin{tabular}{|c|c|c|}
\hline 
Wind turbine & Distance, \emph{m} & Azimuth, \textdegree{}\tabularnewline
\hline
\hline 
1 & 32262 & -30,769\tabularnewline
\hline 
2 & 32020 & -30,043\tabularnewline
\hline 
3 & 32338 & -29,836\tabularnewline
\hline 
4 & 32583 & -30,188\tabularnewline
\hline 
5 & 32543 & -30,835\tabularnewline
\hline 
6 & 32805 & -30,081 \tabularnewline
\hline 
7 & 32656 & -29,633\tabularnewline
\hline 
8 & 32510 & -29,181\tabularnewline
\hline 
9 & 32269 & -28,820\tabularnewline
\hline 
10 & 32520 & -28,812\tabularnewline
\hline 
11 & 32772 & -28,804\tabularnewline
\hline 
12 & 32762 & -29,170\tabularnewline
\hline 
13 & 33050 & -30,426 \tabularnewline
\hline 
14 & 33242 & -30,231\tabularnewline
\hline 
15 & 32045 & -30,508\tabularnewline
\hline 
16 & 32332 & -30,207\tabularnewline
\hline 
17 & 32704 & -30,546\tabularnewline
\hline
\end{tabular}\caption{Parameters of wind turbines}

\end{table}

\subsection{Radar}

\begin{figure}[H]
\centering{}\subfloat[STAR 2000 with high and low beam feeding of the antenna]{\centering{}\includegraphics[scale=0.42]{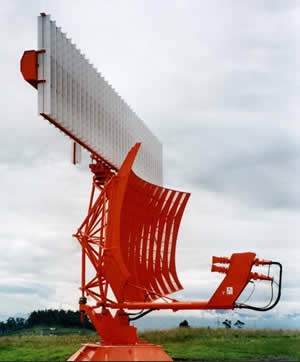}\label{Flo:STAR 2000 with high- and lowbeam}}\subfloat[STAR 2000 and Inverness airport]{\begin{centering}
\includegraphics[scale=0.35]{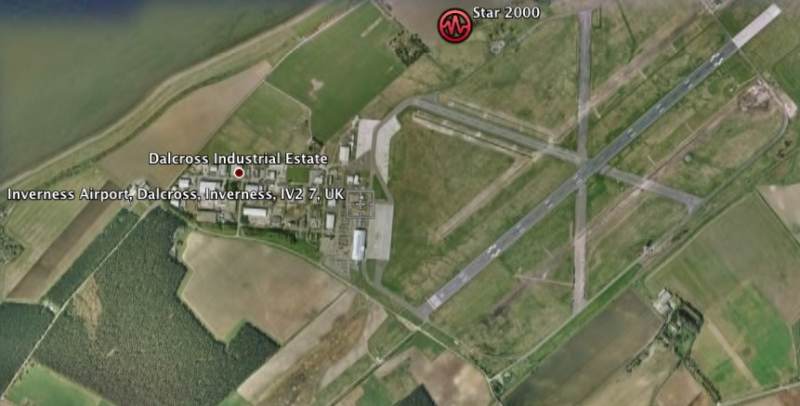}\label{Flo:STAR 2000 and Inverness airport}
\par\end{centering}

}\caption{STAR 2000 }
\label{Flo:STAR 2000}
\end{figure}

During this study we use the radar of Thales Air Systems STAR 2000
(Figure \ref{Flo:STAR 2000 with high- and lowbeam}). It is a dedicated,
solid-state, modular terminal approach radar, which is suitable for
both civilian and military air traffic control applications. The equipment
incorporates a dedicated weather channel and its overall range capability
can be extended from 60 \emph{Nm} (111 \emph{km}) to 90 \emph{Nm}
(167 \emph{km}) through the use of incremental power increases. STAR
2000 configurations exist for stand-alone, Monopulse Secondary Surveillance
Radar (MSSR)/Identification Friend-or-Foe (IFF) associated or Mode
S operation with the radar's data output format being configurable
to match all transmission formats \cite{STAR2000}.

In summary, the main features of STAR 2000 are the following \cite{www.radartutorial.eu}:
\begin{itemize}
\item fixed, shelter-mounted and transportable configurations 
\item full coherence and clutter driven adaptive processing for the improved
target detection in severe clutter conditions 
\item independent dual-polarisation weather channel 
\item modular, fail-safe, online maintainable, solid-state, frequency diverse/agile
transmitter 
\item digital frequency synthesiser and pulse compression with low time-sidelobes 
\item auto-adaptive moving target detection with clutter rejection techniques
\item false alarm free plot extraction and tracking of up to 1 000 targets 
\item MSSR/IFF beacon and Mode S reinforcement 
\item programmable output data formatting 
\item full built-in test and remote monitoring automatic reconfiguration
\end{itemize}
The radar operates in the Inverness airport (Figure \ref{Flo:STAR 2000 and Inverness airport})
- an international airport, the main gateway for travellers to the
north of Scotland with a wide range of scheduled services throughout
the UK and Ireland, and limited charter and freight flights into Europe.

For our example we have certain parameters of the radar (Table \ref{Flo:Parameters STAR2000}).

\begin{table}[H]
\centering{}\begin{tabular}{|c|c|}
\hline 
Parameter & Value\tabularnewline
\hline
\hline 
Frequency, \emph{MHz} & 2800\tabularnewline
\hline 
Polarization & Vertical\tabularnewline
\hline 
Antenna height, \emph{m} & 15\tabularnewline
\hline 
Latitude & 52\textdegree{} 32' 48.1'' N\tabularnewline
\hline 
Longitude & 4\textdegree{} 03' 25.7'' W\tabularnewline
\hline
\end{tabular}\caption{Parameters of the radar}
\label{Flo:Parameters STAR2000}
\end{table}

\begin{figure}[H]
\centering{}\includegraphics[scale=0.5]{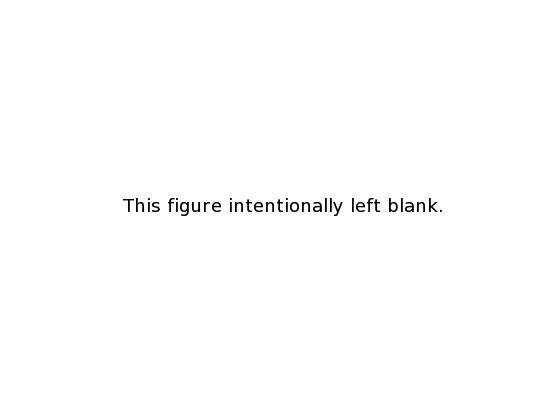}\caption{Vertical antenna pattern (normalised)}
\label{Flo:vertical antenna pattern}
\end{figure}

Figure \ref{Flo:vertical antenna pattern} shows normalised antenna
pattern. This parameter is used when APM calculates the complex electromagnetic
field.

\subsection{Environment}

\begin{figure}[H]
\centering{}\subfloat[Profile]{\begin{centering}
\includegraphics[scale=0.5]{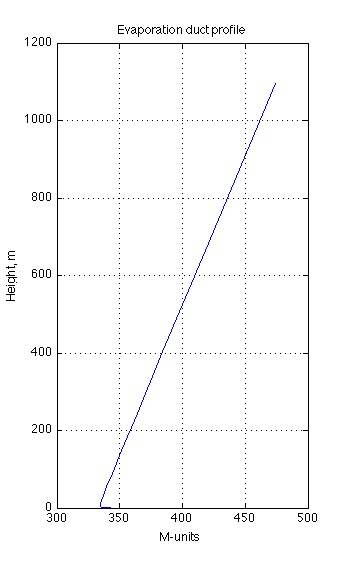}
\par\end{centering}

}\subfloat[Zoomed area]{\begin{centering}
\includegraphics[scale=0.3]{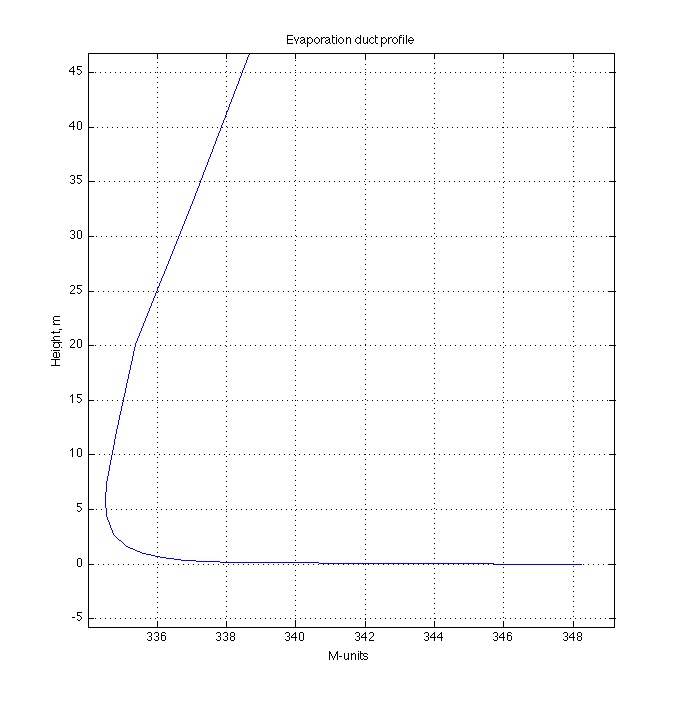}
\par\end{centering}

}\caption{Evaporation duct profile}
\label{Flo:Evaporation duct profile}
\end{figure}

All the atmospheric data is based on DCS and Sfc data received from
AREPS. The statistical information is presented for an average day
of July. This data is shown in Table \ref{Flo:Parameters of the space}
and Figure \ref{Flo:Evaporation duct profile}.

\begin{table}[H]
\centering{}\begin{tabular}{|c|c|}
\hline 
Parameter & Value\tabularnewline
\hline
\hline 
Surface absolute humidity, \emph{g/m$^{3}$} & 10.55\tabularnewline
\hline 
Surface air temperature, \emph{\textdegree{}C} & 14.6\tabularnewline
\hline 
Gaseous absorbtion attenuation rate, \emph{dB/km} & 0\tabularnewline
\hline 
Windspeed, \emph{m/s} & 6.1\tabularnewline
\hline
\end{tabular}\caption{Parameters of the space}
\label{Flo:Parameters of the space}
\end{table}

Actually, it is possible to receive such an information in real time.
There are services that collect, consolidate, perform and offer the
data from weather station. Thus, one can make and apply the module
that implements this information for needs of this project.

\section{Implementation\label{sec:Implementation}}

\subsection{Complex Pattern Propagation Factor}

Amplitude is less sensitive to resolution than the phase does. In
fact, losses of an amplitude depend on the refractivity profile, presence
of the terrain, wind and some others. It changes APPF on relatively
big dimensions, which let choose the low resolution. 

There are another demands on PPPF. In case of PPPF, the resolution
of the space should depend on the frequency of the source. For instance,
we deal with relatively small object like a wind turbine. If we have
the resolution of APPF less than the dimension of a turbine, we can
increase the resolution artificially, interpolating the data. Unfortunately,
we can not do that with PPPF, if the initial resolution was less than
the dimension of a turbine. According to the Kotelnikov sampling theorem
\ref{thm:If-a-function}, we have the discrete frequency of PPPF.
\begin{thm}
If a function x(t) contains no frequencies higher than B hertz, it
is completely determined by giving its ordinates at a series of points
spaced 1/(2B) seconds apart.\label{thm:If-a-function}

\begin{equation}
B<\frac{f_{s}}{2}\end{equation}

\begin{equation}
T\stackrel{\mathrm{def}}{=}\frac{1}{f_{s}}\end{equation}

\end{thm}
Fortunately, APM lets receive any resolution and it is defined solely
by input parameters. The calculation time increases accordingly. 

As the example concerns ATM radars, frequencies of the source lie
in band \emph{S}. Here we use \emph{f = 2.8} \emph{GHz} that corresponds
to $\mathcal{\lambda}$\emph{ $\thickapprox$0,1 m}. Thus, the resolution
of our PPPF should be more than twice \emph{0,1 m}. 

We choose the first turbine to show the complex PPF.

\begin{figure}[H]
\begin{centering}
\subfloat[APPF in the direction of  turbine 1 (in terms of amplitude), \emph{dB}]{\centering{}\includegraphics[scale=0.45]{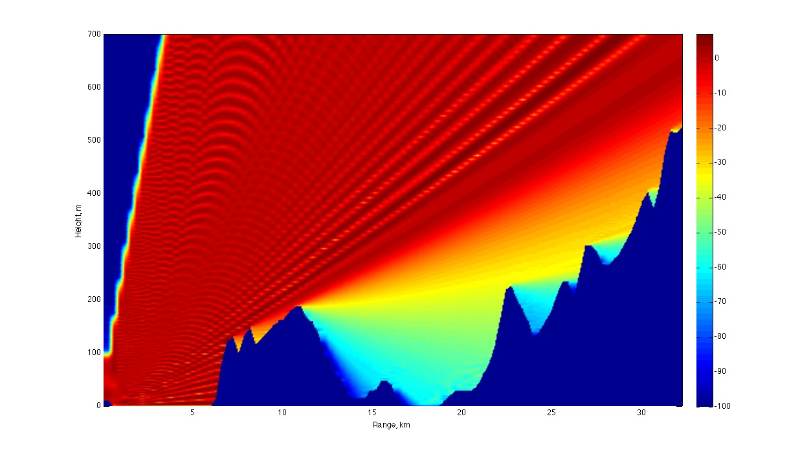}\label{Flo:amplitude 1}}
\par\end{centering}

\centering{}\subfloat[Height-APPF of  wind turbine 1]{\begin{centering}
\includegraphics[scale=0.22]{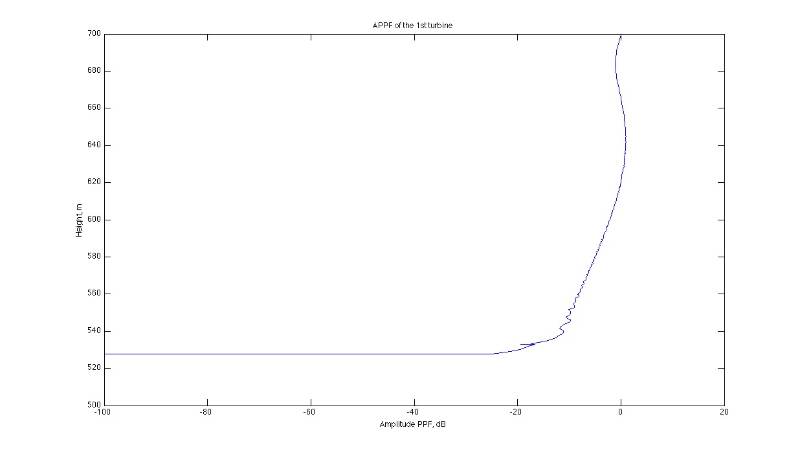}
\par\end{centering}

\centering{}\label{Flo:amplitude height 1}}\subfloat[Zooomed area]{\centering{}\includegraphics[scale=0.22]{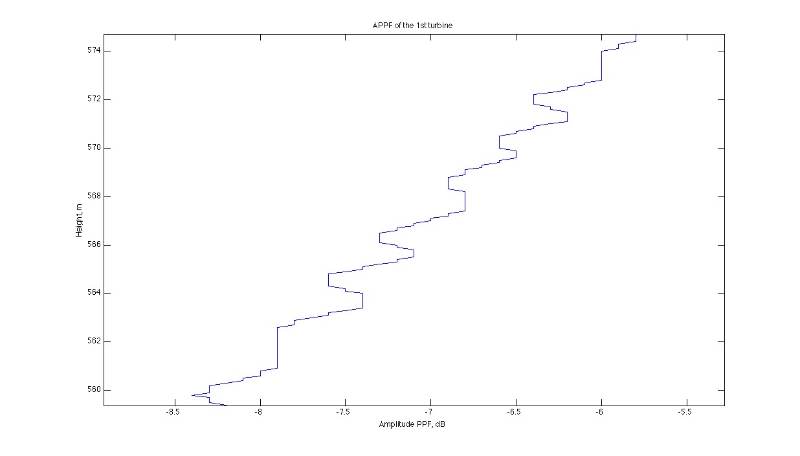}\label{Flo:amplitude height 1 zoomed}}\caption{Amplitudes of PPF of  wind turbine 1}
\label{Flo:amplitudes of PPF of wind turbine 1}
\end{figure}

Firstly, we examine APPF. Figure \ref{Flo:amplitude 1} shows the
propagation of electromagnetic field in terms of an amplitude of the
pattern propagation factor. The radar emits from the left and the
turbine is situated on the hill on the right. Figure \ref{Flo:amplitude height 1}
presents the APPF, where the turbine operates depending on the height.
Its zoomed area is shown in Figure \ref{Flo:amplitude height 1 zoomed}.
Here, one can see that the resolution chosen for calculation is satisfactory.

\begin{figure}[H]
\begin{centering}
\subfloat[PPPF in the direction of  turbine 1 (in terms of phase), \emph{rad}]{\centering{}\includegraphics[scale=0.45]{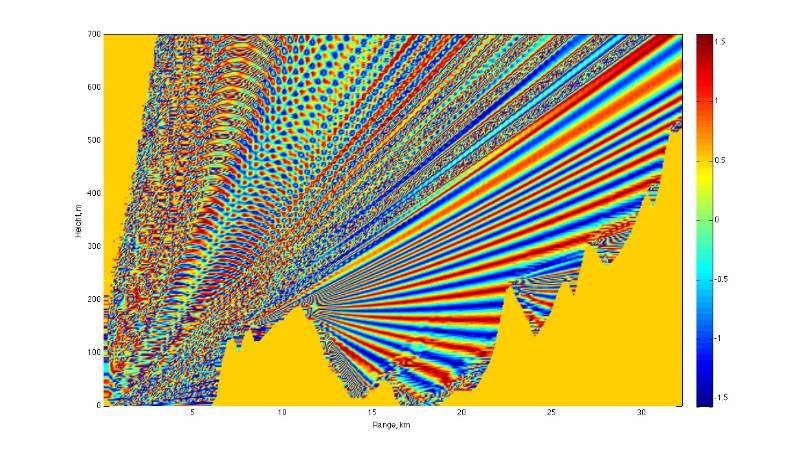}\label{Flo:phase 1}}
\par\end{centering}

\centering{}\subfloat[Height-PPPF of  wind turbine 1]{\begin{centering}
\includegraphics[scale=0.22]{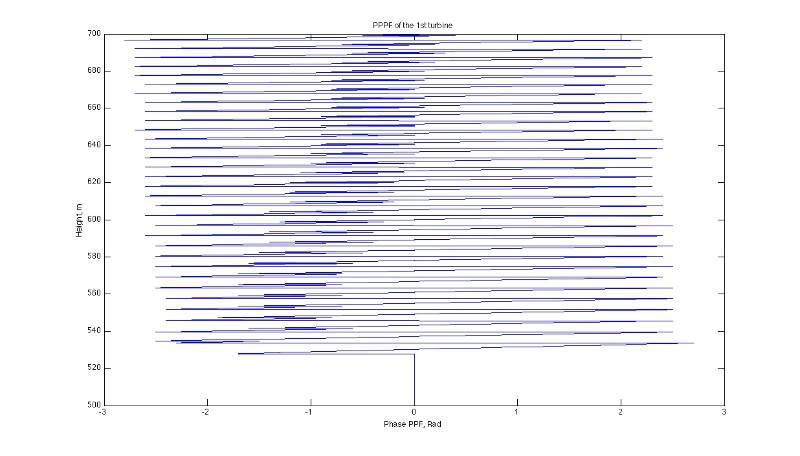}
\par\end{centering}

\centering{}\label{Flo:phase unzoomed}}\subfloat[Zooomed area]{\centering{}\includegraphics[scale=0.22]{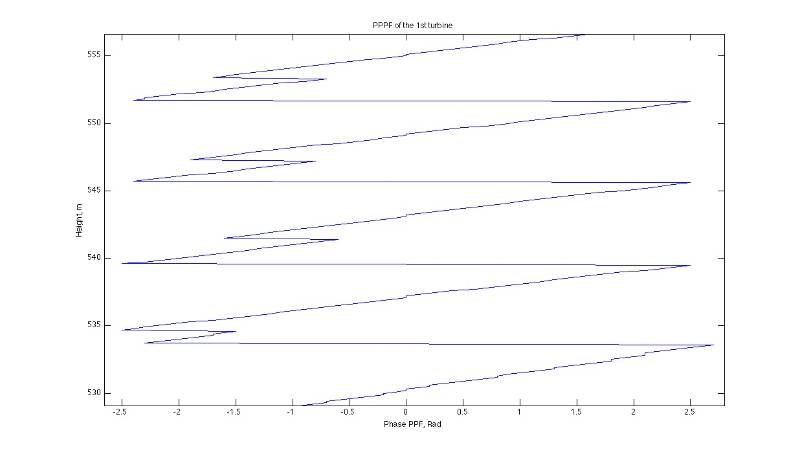}\label{Flo:phase zoomed}}\caption{Phases of PPF of  wind turbine 1}
\label{Flo:phases of PPF of wind turbine 1}
\end{figure}

Now we examine PPPF. Figure \ref{Flo:phases of PPF of wind turbine 1}
is similar to Figure \ref{Flo:amplitudes of PPF of wind turbine 1}
but shows the phase of an electromagnetic field. The phase is presented
in radians.

\begin{figure}[H]
\centering{}\includegraphics[scale=0.45]{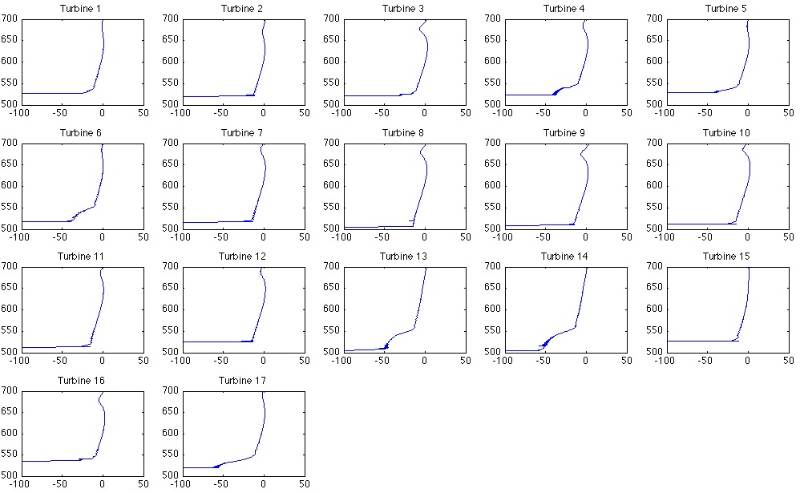}\caption{Height-APPF of the wind turbines 1-17}
\label{Flo:appf 17}
\end{figure}

\begin{figure}[H]
\centering{}\includegraphics[scale=0.45]{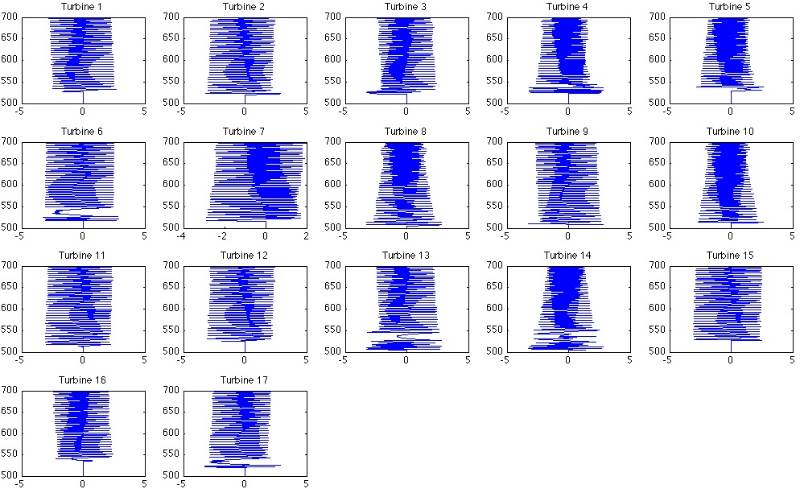}\caption{Height-PPPF of the wind turbines 1-17}
\label{Flo:pppf 17}
\end{figure}

The Beinn Tharsuinn wind farm has 17 turbines. For every disposition
of the turbine we calculate APPF (Figure \ref{Flo:appf 17}) and PPPF
(\ref{Flo:pppf 17}). Finally we are ready to compute the RCS of wind
turbine based on the complex pattern propagation factor.

\subsection{CAD Model of Wind Turbine}

As the wind farm consists of turbines Vestas V66, we have to get its
CAD model. The most precise model of the wind turbine can be given
only by its producer. However, it is possible to make our own turbine,
which will be very close to original one. We showed in Subsection
\ref{sub:CAD-Model-of}, the parts of the turbine are standardised
and, using the open information, one can make its model.

\begin{figure}[H]
\subfloat[FFA-W3-211 Airfoil]{\centering{}\includegraphics[scale=0.2]{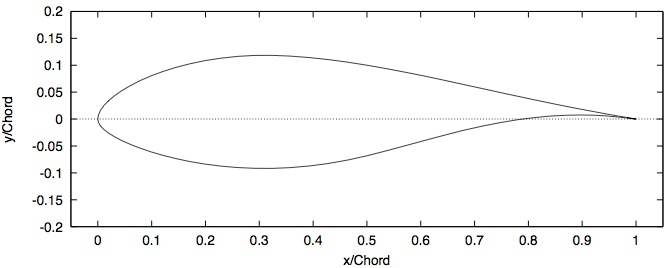}\label{Flo:FFA-W3-211}}\subfloat[FFA-W3-241 Airfoil]{\centering{}\includegraphics[scale=0.2]{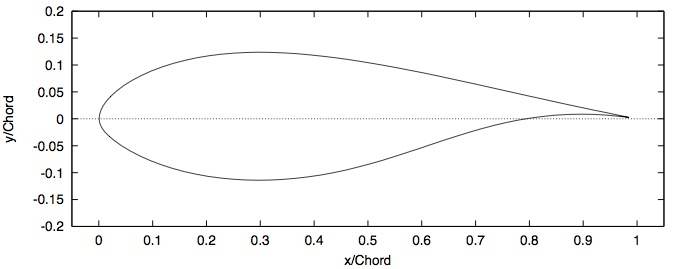}\label{Flo:FFA-W3-241}}\subfloat[FFA-W3-301 Airfoil]{\centering{}\includegraphics[scale=0.2]{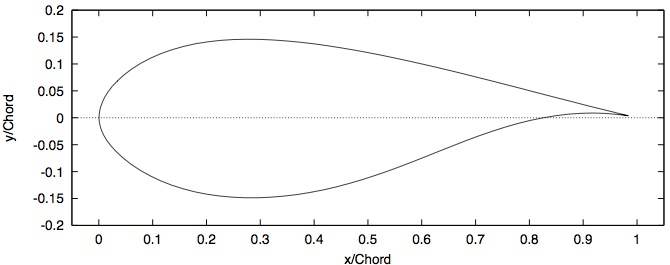}\label{Flo:FFA-W3-301}}\caption{Models of airfoils}
\label{Flo:models of airfoil}
\end{figure}

The data about V66 have been found from Internet. Basing on this information
we create Rhinoceros NURBS CAD model. FFA-W3-211, FFA-W3-241 and FFA-W3-301
airfoil (Figure \ref{Flo:models of airfoil}) have been successively
chosen along the blade axis. 

\begin{figure}[H]
\centering{}\subfloat[View from front]{

\centering{}\includegraphics[scale=0.2]{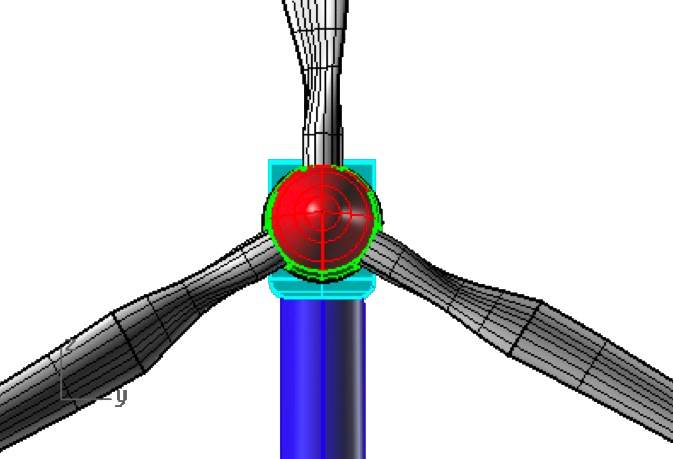}\hspace{2.6cm}\label{Flo:View from front}}\subfloat[View from left]{\begin{centering}
\includegraphics[scale=0.45]{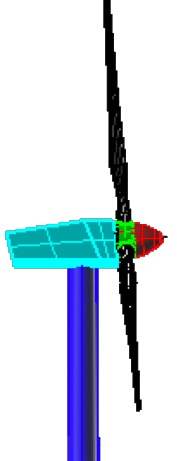}\label{Flo:View from left}
\par\end{centering}

}\\
\subfloat[View from above]{\begin{centering}
\includegraphics[scale=0.4]{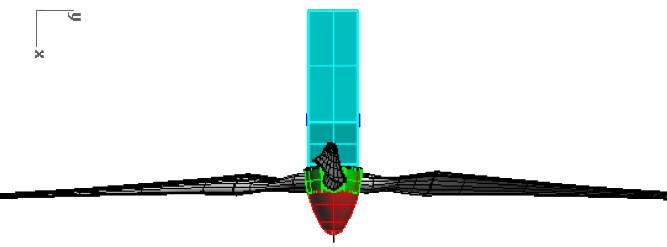}\label{Flo:View from above}
\par\end{centering}

}\subfloat[ALBEDO 3D view]{

\begin{centering}
\includegraphics[scale=0.3]{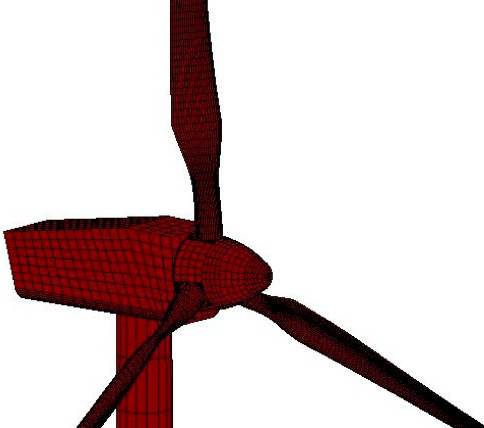}\label{Flo:View 3D}
\par\end{centering}

}\caption{Rhinoceros NURBS CAD model of Vestas V66}
\label{Flo:Rhinoceros NURBS CAD model of Vestas V66}
\end{figure}

The tower is considered as a truncated cone. The whole model is composed
by 51 non-uniform rational B-spline (NURBS) surfaces (Figures \ref{Flo:View from front},
\ref{Flo:View from left}, \ref{Flo:View from above}).

The file with CAD model has been transformed to match with a physical
theory of diffraction of ALBEDO\textquoteright{}s geometrical inputs
(a software computing RCS). It is then made of 539 patches. 

The turbine is considered as a perfect conductor (Figure \ref{Flo:View 3D}).

\subsection{Radar Cross Section\label{sub:Radar-Cross-Section}}

Computations of RCS have been made by the tool shown in Subsection
\ref{sub:Radar-cross-section}. 

\begin{figure}[H]
\begin{centering}
\includegraphics[scale=0.3]{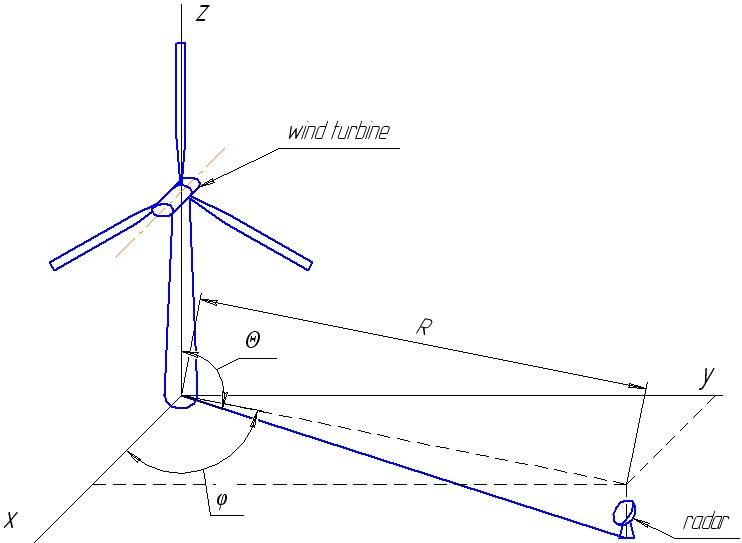}
\par\end{centering}

\caption{Relative position of the turbine and radar}

\label{Flo:relative position}
\end{figure}

Having a 3D CAD model of the wind turbine and knowing an electromagnetic
field from its top to the bottom, we can simulate different scenarios
of the wind turbine operation. It means we can calculate RCS for any
angle of blades and for different position of our radar relative to
the turbine. In fact, we deal with three angles and one distance shown
in Figure \ref{Flo:relative position}.

Vestas V66 has three blades. The angle between them is $\frac{\pi}{3}$.
It means that turning the turbine has three states equal for the radar.
If we consider that initial state is 0 (the blade is aimed vertically
upwards) and its full turn is 1, thus one third of the turn is $\frac{1}{3}.$ 

For this step we need the complex electromagnetic field depending
on the height for certain wind turbine. Therefore, we prepare the
data that consists of 1011 points (Algorithm \ref{Flo:complex data 1}).
Each point is the value of complex field on the different height -
HEIGHT (REAL PART, IMAGINARY PART).

\begin{algorithm}[H]
1 1011 

-67 ( 0.000000000000 , 0.000000000000 ) 

-66.9 ( 0.000000000000 , 0.000000000000 ) 

-66.8 ( 0.000000000000 , 0.000000000000 ) 

-66.7 ( 0.000000000000 , 0.000000000000 ) 

-66.6 ( 0.000000000000 , 0.000000000000 ) 

-66.5 ( 0.000000000000 , 0.000000000000 ) 

-66.4 ( 0.000000000000 , 0.000000000000 ) 

-66.3 ( 0.000000000000 , 0.000000000000 ) 

-66.2 ( 0.000000000000 , 0.000000000000 ) 

-66.1 ( 0.000000000000 , 0.000000000000 ) 

-66 ( 0.000000000000 , 0.000000000000 ) 

-65.9 ( 0.000009153603 , -0.000326871858 ) 

-65.8 ( 0.000025242162 , -0.000335050494 ) 

-65.7 ( 0.000043548126 , -0.000343248541 ) 

-65.6 ( 0.000064158115 , -0.000351187608 ) 

-65.5 ( 0.000085202205 , -0.000356972806 ) 

... 

0 ( 0.002571270631 , -0.003091441143 )

... 

33.5 ( 0.001333394317 , 0.005907384836 ) 

33.6 ( 0.000574750876 , 0.006025651038 ) 

33.7 ( -0.000149071504 , 0.006033158599 ) 

33.8 ( -0.000873455115 , 0.005952254376 ) 

33.9 ( -0.001583575708 , 0.005812130761 ) 

34 ( -0.002281865277 , 0.005602097095 )\caption{The data of complex field to compute RCS of  wind turbine 1}
\label{Flo:complex data 1}

\end{algorithm}

\begin{figure}[H]
\begin{centering}
\includegraphics[scale=0.3]{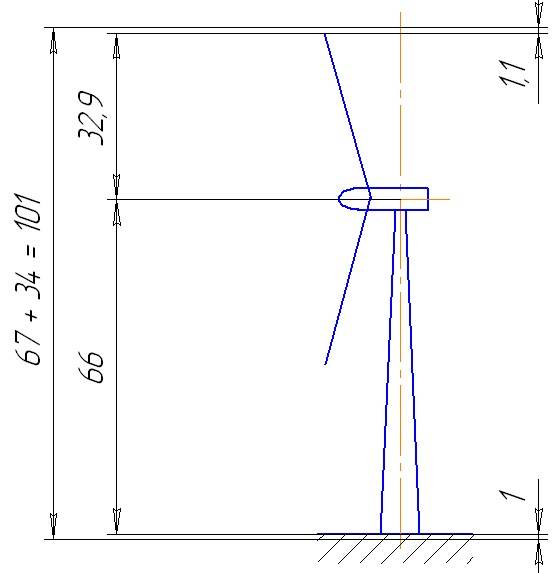}
\par\end{centering}

\caption{Dimensions for calculations, \emph{m}}
\label{Flo:dimensions for calculations}

\end{figure}

We assume that the zero point is the nacelle of a turbine. The data
is given for the entire turbine and additionally 1 \emph{m} below
and 1.1 \emph{m} above (Figure \ref{Flo:dimensions for calculations}).

\begin{figure}[H]
\begin{centering}
\subfloat[Disposition of the turbine]{\begin{centering}
\includegraphics[scale=0.3]{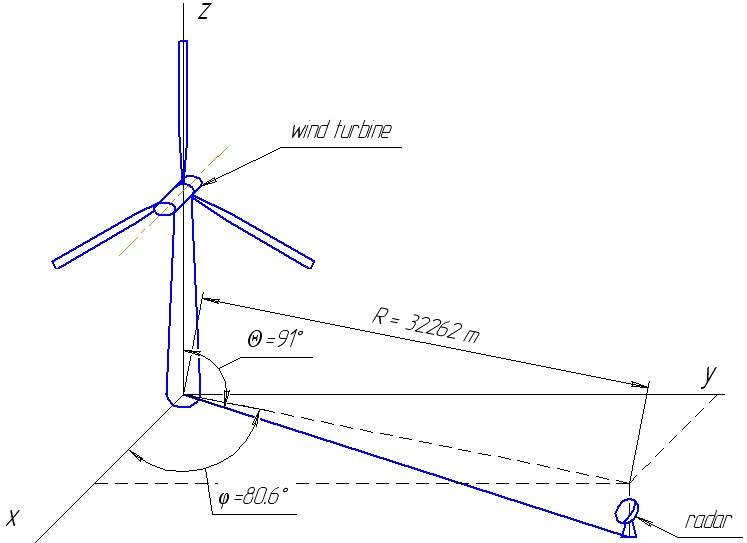}\label{Flo:RCS turbine disposition 32262}
\par\end{centering}

}\subfloat[Terrain profile and PPF]{\centering{}\includegraphics[scale=0.22]{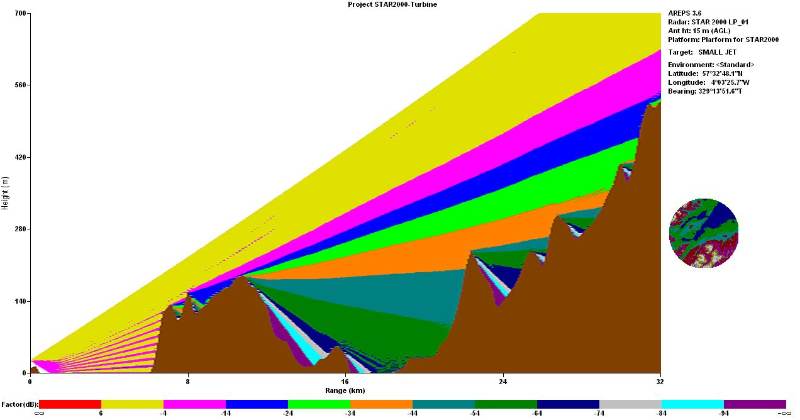}\label{Flo:RCS terrain profile 32262}}
\par\end{centering}

\begin{centering}
\subfloat[Amplitude]{\begin{centering}
\includegraphics[scale=0.22]{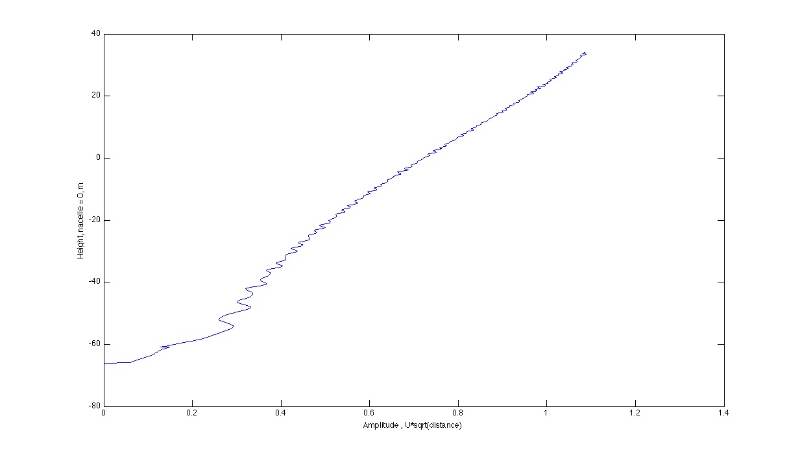}\label{Flo:RCS turbine amplitude 32262}
\par\end{centering}

}\subfloat[Phase]{\centering{}\includegraphics[scale=0.22]{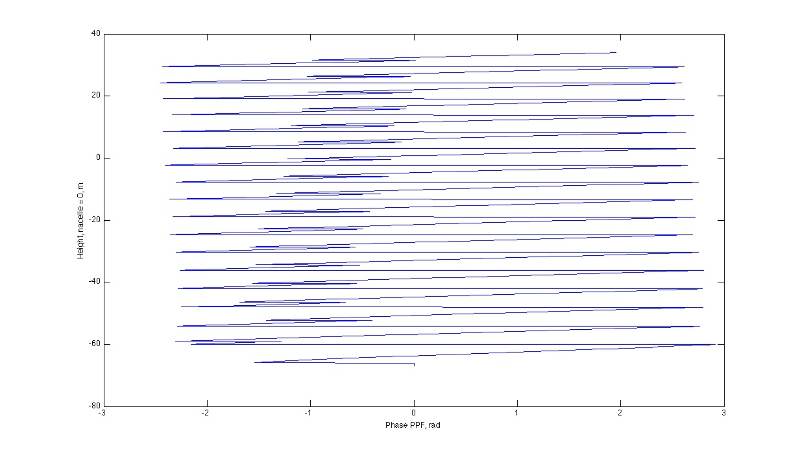}\label{Flo:RCS turbine phase 32262}}
\par\end{centering}

\centering{}\subfloat[RCS without the phase (only an amplitude)]{\centering{}\includegraphics[scale=0.23]{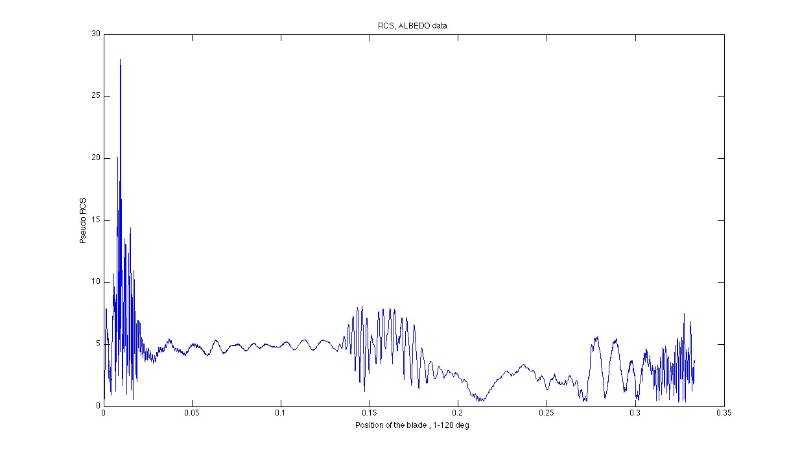}\label{Flo:RCS without the phase 32262}}\subfloat[RCS using the complex electromagnetic fiels (an amplitude and a phase)]{\begin{centering}
\includegraphics[scale=0.23]{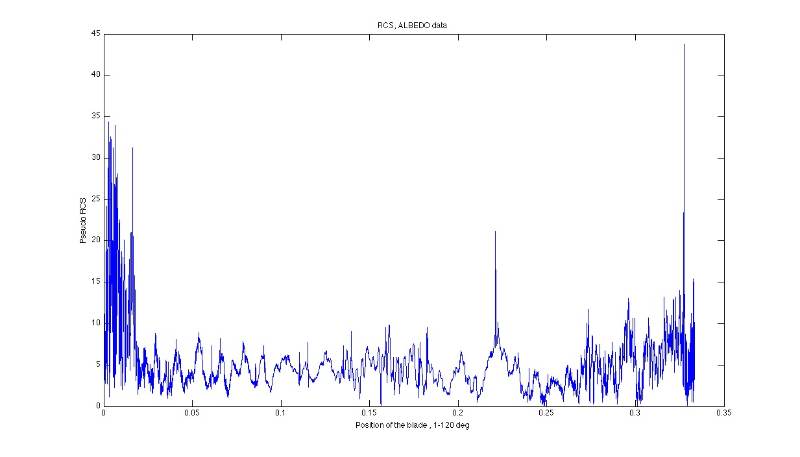}\label{Flo:RCS  complex 32262}
\par\end{centering}

}\caption{Turbine 1, range 32262 \emph{m}}
\label{Flo:RCS turbine 1 80.6 32262m}
\end{figure}

Figure \ref{Flo:RCS turbine disposition 32262} presents the disposition
of the radar relative to the axis of the wind turbine. The wind turbine
is situated on the surface on the right of the terrain profile (Figure
\ref{Flo:RCS terrain profile 32262}), while the radar is on the left.
As we can see the wind turbine operates on the top of the hill. It
interferes with the radar because this is the most visible position.
The pattern propagation factor in this area proves that.

Figures \ref{Flo:RCS without the phase 32262} and \ref{Flo:RCS  complex 32262}
show the wind turbine RCS evolution that turns form the state 0 to
$\frac{1}{3}$ ($\alpha$ from 0 to $\frac{\pi}{3}$). Actually, this
is the period of a signal evolution. Then the signal repeats itself.
This is the first turbine of the Beinn Tharsuinn wind farm. With respect
to the radar it has $\theta=91\lyxmathsym{\textdegree}$ , $\varphi=80.6\lyxmathsym{\textdegree}$.
The peak of RCS takes place when the turbine has a maximal visible
surface, i.e. $\alpha=0$, or, in other words, when one blade stands
vertically. Also this result corresponds with previous publications
and calculations \cite{poupart2003wind,Stealthbladesaprogressreport}.
Figures \ref{Flo:RCS without the phase 32262} and \ref{Flo:RCS  complex 32262}
are based on different electromagnetic field. Thus, in order to compute
RCS, presented in Figure \ref{Flo:RCS without the phase 32262}, we
only use the amplitude of the electromagnetic field (Figure \ref{Flo:RCS turbine amplitude 32262}).
RCS shown in Figure \ref{Flo:RCS  complex 32262} takes into account
the complex field, in other words - an amplitude and a phase (Figures
\ref{Flo:RCS turbine amplitude 32262} and \ref{Flo:RCS turbine phase 32262}
respectively). One can see that the ordinates of the Figures \ref{Flo:RCS turbine amplitude 32262}
and \ref{Flo:RCS turbine phase 32262} mean the altitude and corresponds
with dimensions presented in Figure \ref{Flo:dimensions for calculations}
(-67 \emph{m} .. 34 \emph{m}). The revolution speed in every example
is 13 \emph{rpm}.

The reason why we show the Figure \ref{Flo:RCS without the phase 32262}
is that, before, we took into account only an amplitude because APM,
inicially, does not compute the phase of the field. One of the objects
of this study is to retrieve the phase in order to compute RCS using
a complex electromagnetic field (that is shown in Chapter \ref{cha:Complex-pattern-propagation}
in details). Now we can compare the impact the previous version (Figure
\ref{Flo:RCS without the phase 32262}) and the new one (\ref{Flo:RCS  complex 32262}).

\begin{figure}[H]
\begin{centering}
\subfloat[Disposition of the turbine]{\centering{}\includegraphics[scale=0.3]{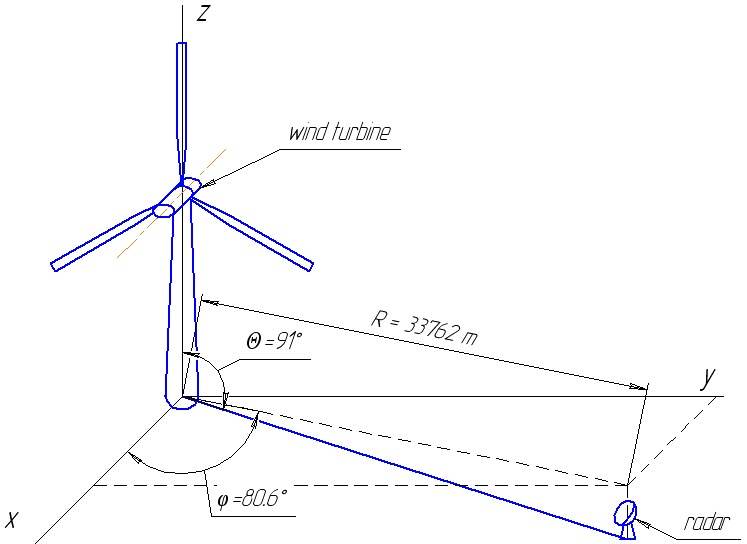}\label{Flo:RCS turbine disposition 33762}}\subfloat[Terrain profile and PPF]{\centering{}\includegraphics[scale=0.22]{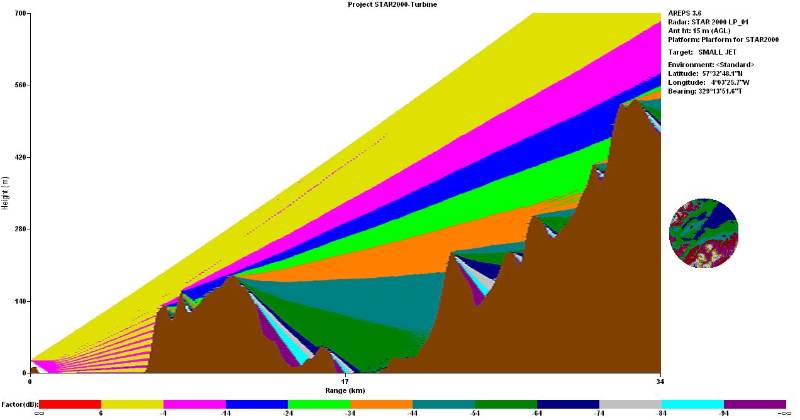}\label{Flo:RCS terrain profile 33762}}
\par\end{centering}

\begin{centering}
\subfloat[Amplitude]{\centering{}\includegraphics[scale=0.22]{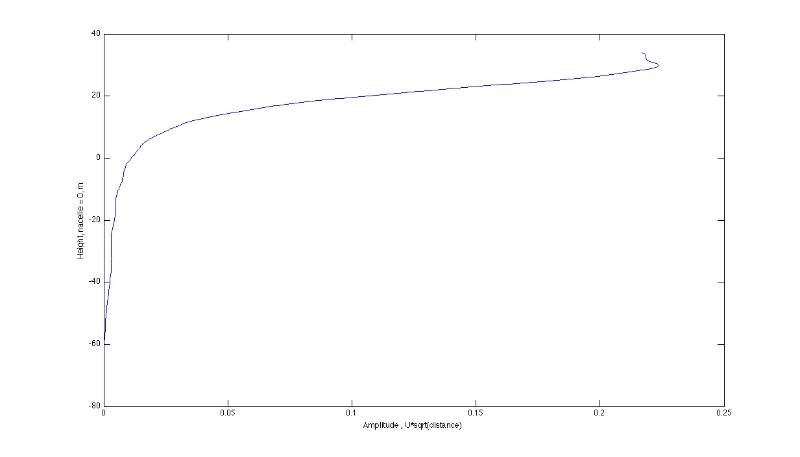}\label{Flo:RCS turbine amplitude 33762}}\subfloat[Phase]{\centering{}\includegraphics[scale=0.22]{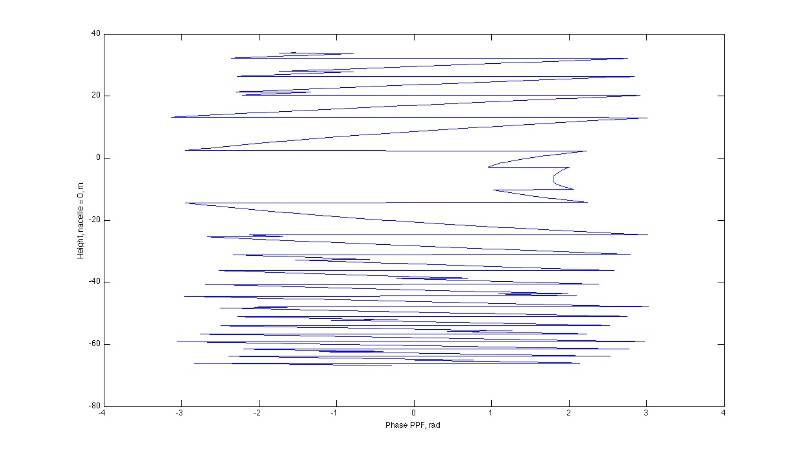}\label{Flo:RCS turbine phase 33762}}
\par\end{centering}

\begin{centering}
\subfloat[RCS without the phase (only an amplitude)]{\centering{}\includegraphics[scale=0.23]{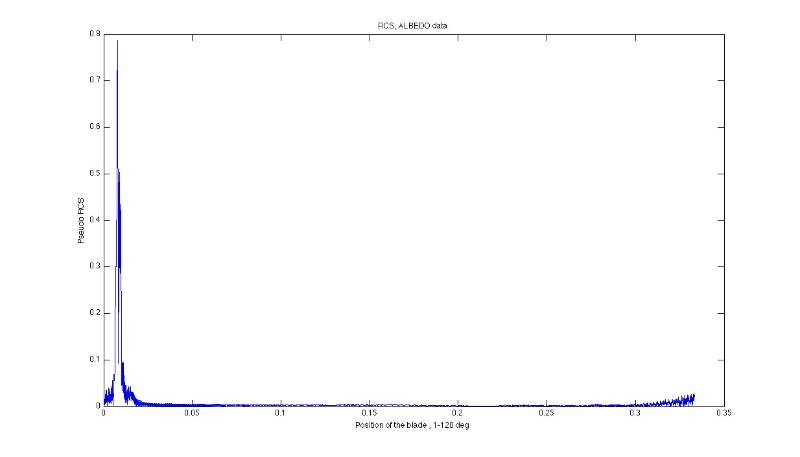}\label{Flo:RCS without the phase 33762}}\subfloat[RCS using the complex electromagnetic fiels (an amplitude and a phase)]{\centering{}\includegraphics[scale=0.23]{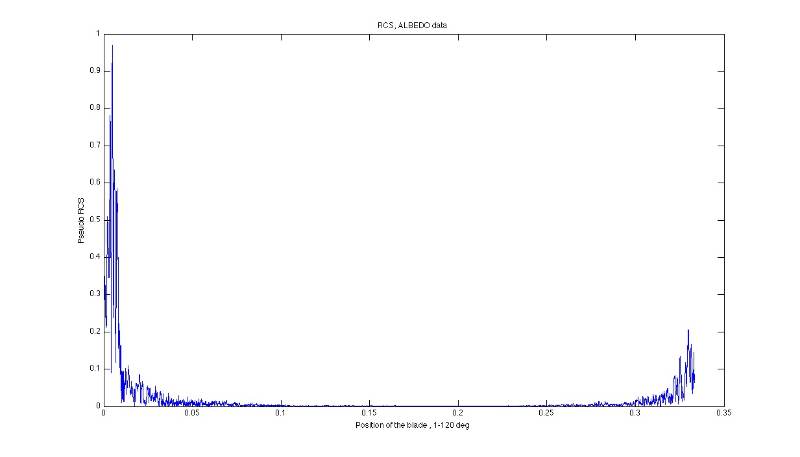}\label{Flo:RCS  complex 33762}}\caption{Turbine 1\textbf{, }range 33762 \emph{m}}

\par\end{centering}

\label{Flo:RCS turbine 1 80.6 33762m}
\end{figure}

An electromagnetic field depends on the place in the space. If we
take the first turbine of the Beinn Tharsuinn wind farm and relocate
1500 \emph{m} further , we can calculate the complex field and, more
importantly, simulate RCS. So, changing the disposition of the first
turbine, angles remain the same $\theta=91\lyxmathsym{\textdegree}$
and $\varphi=80.6\lyxmathsym{\textdegree}$, but the distance radar/turbine
increases from 32262 \emph{m }to 33762 \emph{m} (+1500 \emph{m}).
As we can see (Figure \ref{Flo:RCS terrain profile 33762}) the turbine
is now situated behind the hill in a little hole. Let us examine the
RCS there (Figure \ref{Flo:RCS turbine 1 80.6 33762m}) and compare
with the original disposition (Figure \ref{Flo:RCS turbine 1 80.6 32262m}).

The maximum RCS value of the current wind turbine disposition (Figure
\ref{Flo:RCS  complex 32262}) is about 30 times greater than for
the new disposition (Figure \ref{Flo:RCS  complex 33762}). The maximum
value takes place when one blade stands vertically. If we take other
values, we will see that the difference reaches hundreds times. Therefore,
the new disposition makes a wind turbine almost invisible for the
radar.

The possible place of the Beinn Tharsuinn wind farm can be close to
the original disposition but more hidden from the radar behind the
hill.

\section{Recommendations as a Result}

The Beinn Tharsuinn wind farm is situated on the hill at a height
510-550 \emph{m} (Figure \ref{Flo:17 wind turbines}). If the wind
farm is only being considered to install, we can recommend the disposition,
where the farm has less influence. Despite that we simulated only
one turbine, it is possible to spread its parameters to the whole
wind farm. RCS of the turbine on the current disposition (Figure \ref{Flo:RCS  complex 32262})
is higher than the RCS if the turbine would operate 1500 \emph{m}
further (dozens-hundreds times). At the same time the altitude decreases
from 540 \emph{m} to 450 \emph{m}. Therefore, it is necessary to propose
new disposition because the wind farm impact will be less. If the
position of the wind turbine producer and its customer is flexible
and they are ready to lose some power efficiency, than the wind farm
can be built 1500 \emph{m} further.

Meanwhile, the new location does not always lead to desirable result.
It is necessary to have in mind a duct effect, when due to the special
refractivity profile, electromagnetic waves can propagate behind the
hill (Figure \ref{Flo:example of APM}). That is why we need to use
the propagation model like APM.

\chapter{Recommendations for Further Work\label{cha:Recommendations-for-Further}}

\section{Pseudo 3D propagation}

APM is a powerful tool used by different software such as AREPS, NEMESIS
and TEMPER. All of them operate in 2D mode. 

One of the most successful previous projects regarding pseudo 3D study
is the research of QinetiQ, the UK \cite{poupart2003wind}. However,
they calculate RCS of wind turbine independently from the presence
of the terrain. There are some other limitations in their work as
well, which our study avoids. Using directly APM source code we can
make more powerful tool met our requirements. Before some pseudo 3D
studies based on PE have been provided and, recently, there were published
a few articles \cite{ra2004modeling,hawkes-three,dockery2007overview}.

Initially, APM requires a list of input parameters of the source,
space and terrain, and it gives 2D image of propagation.

This module written in Matlab (see DVD) reads the DTED data and runs
APM for every layer according to the lateral resolution. At the same
time it uses all the set of APM parameters. The module detects the
DTED level (1, 2, 3) in order to set the required resolution.

To show the results of this module, we take the Inverness airport
in the UK, where Star 2000 operates in the presence of the wind farm.

\begin{figure}[H]
\subfloat[Terrain DTED, E28 N45, \emph{100 km X 80 km}]{\includegraphics[scale=0.3]{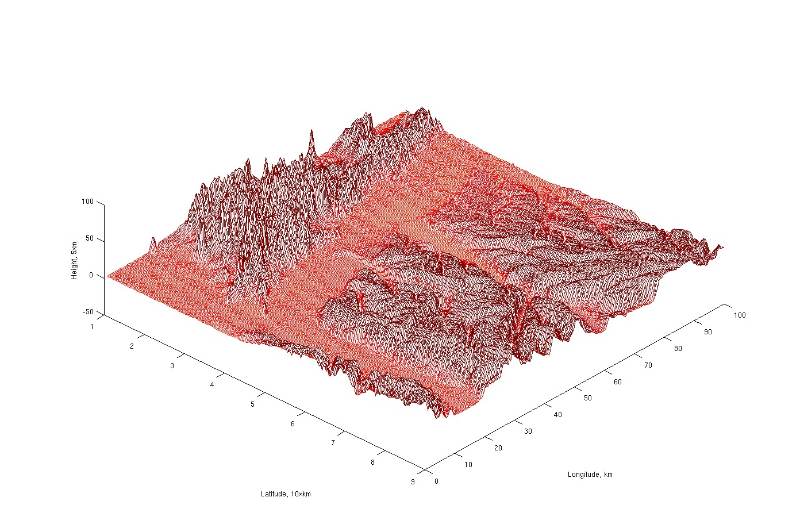}\label{Flo:Terrain DTED}}\subfloat[Layers of the propagation based on DTED]{\includegraphics[scale=0.3]{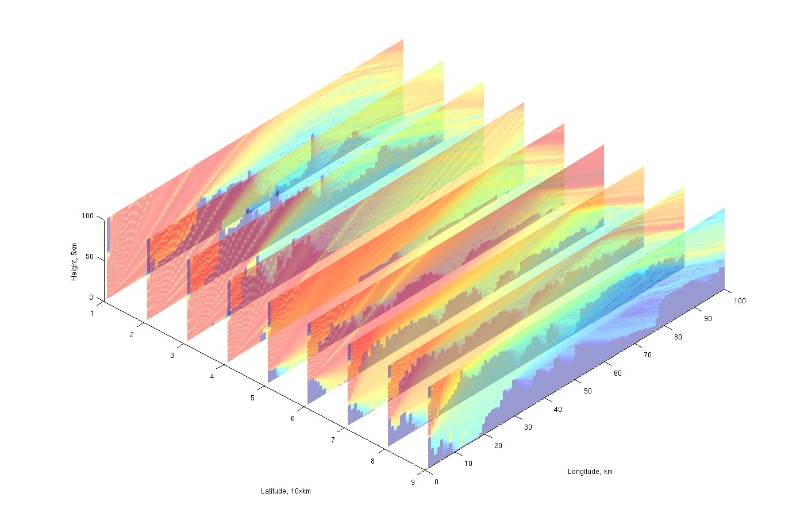}\label{Flo:Layers of propagation }}

\subfloat[Combined model]{\includegraphics[scale=0.5]{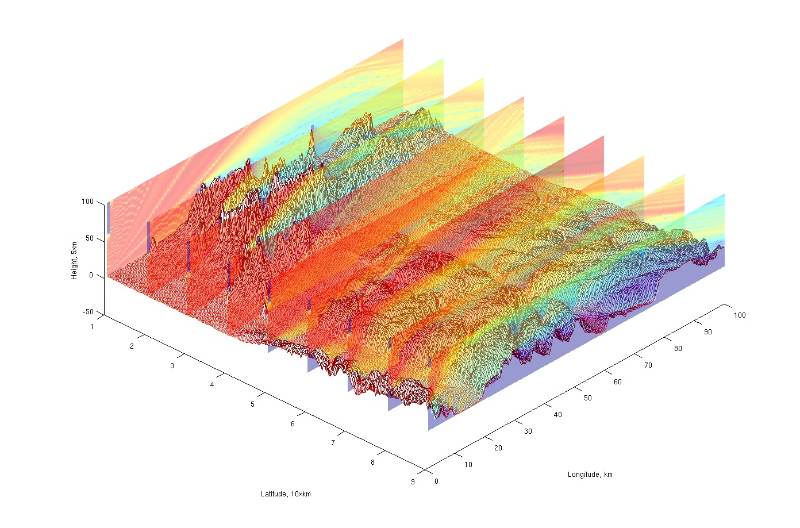}\label{Flo:Combined model}}\caption{Pseudo 3D propagation model }
\label{Flo:Pseudo 3D propagation model }
\end{figure}

In our case, we use DTED of level 1 from data E28 N45 (area N42-N60,
W014-E082) 100\emph{ km X }80\emph{ km}, which is in Figure \ref{Flo:Terrain DTED}.

Basing on this terrain, the written Matlab script runs APM for every
layer with the defined step. The result of this is shown in Figure
\ref{Flo:Layers of propagation }.

Finally, one can combine previous images and receive pseudo 3D propagation
model (Figure \ref{Flo:Combined model}).

Obviously, the pseudo 3D propagation model has similar problems as
APM does:
\begin{itemize}
\item Backscatter waves are not considered;
\item Terrain is approximated into rectangles;
\item Initial theoretical assumption of used APM models.
\end{itemize}
The main shortcoming of pseudo 3D propagation model is the absence
of the consideration of a horizontal propagation. In fact, this is
the reason why the model is called {}``pseudo''.

In spite of these drawbacks, pseudo 3D propagation model could became
the useful tool in wind farm problematic. To predict the influence
of wind farms on the ATM radar we offer the method based on the following
steps:
\begin{enumerate}
\item Wind farm RCS based on the complex PPF considering the real terrain.
Here we need APM, its application on the 3D environment, CAD of the
wind turbine and RCS software.
\item Simulation of the wind farm impact on the radar. Here we need to know
the performance of our radar with its probability thresholds and the
tool to simulate an output of the radar.
\item Solution to reduce the wind farm impact. Here we need the tool to
automate the previous steps (written in Fortran 90 and in Matlab at
least) in order to receive different scenarios and to choose the more
sufficient one. ASTRAD is an excellent tool to realise this step.
\end{enumerate}
Another advantage of this study is that it gives spectacular results
(Figure \ref{Flo:Pseudo 3D propagation model }). It could be useful
for marketing aspects.

State-of-the-art of the pseudo 3D model is given in Subsection \ref{sub:Pseudo-3D}.

Otherwise, if the pseudo 3D model does not satisfy our needs, it is
necessary to use a vector PE model. There is no successeful software
that can replace 2D PE. Therefore, one can wait when the vector PE
model will appear on the market or develop by ST\&I. However, this
is a very complicated task that needs time and resources.

\section{Publications \& Patents}

Publications and patents should be made as soon as possible. According
to the patent analysis we can declare that still there is no patents
that give the solution for the radar regarding wind turbines. We have
been searching patents in databases of European Patent Office%
\footnote{www.espacenet.com/%
}, World Intellectual Property Organization%
\footnote{www.wipo.int/pctdb%
} and in Google Patents service%
\footnote{www.google.com/patents%
}. 

In our case, the object to be patented is not a process, machine,
article of the manufacture or composition of matter. This is the method
to evaluate the wind farm influence. It includes the set of software
and mathematical instrument with defined algorithm. Therefore, it
could be difficult to patent the method. However, in practice it is
possible to receive this patent in the U.S., which is called a \emph{software
patent}. In other countries, especially with the developed wind power
industry, one can obtain a \emph{utility model }that is very similar
to the patent, but usually has a shorter term (often 6 or 10 years)
and less stringent patentability requirements. PCT application is
very expensive (about 30000 €).

Another way to protect our intellectual property is to publish the
articles. There are a lot of presentations and reports about radar
and wind turbines. At the same time it is hard to find the article
in famous scholarly journal. Nowadays, the publications are closely
related with the patents and play the role of state-of-the-art when
the patent application is considered. Thus, this is the easiest way
to prevent the potential use of our ideas. In general, it is important
to deal with both societies - the electronics and wind power industry.

On the part of the electrical and electronic engineering, IEEE journals
and magazines are appropriate. There are a few relevant journals:
Antennas \& Propagation Magazine, Antennas and Propagation, Control
Systems and Technology, Signal Processing, etc. Also IEEE conferences
offer to publish: Radar Conference, Waveform Diversity \& Digital
Radar Conference, etc.

On the part of the wind power industry, the possible journals are
Renewable Energy World Magazine, Wind Engineering, Wind Energy Weekly,
etc.

\section{Phase of APM}

In Chapter \ref{cha:Complex-pattern-propagation} we show how to retrieve
the phase from APM. However, there is one bug that gives a reason
for concern. As one can see from Figures \ref{Flo:phase unzoomed}
and \ref{Flo:phase zoomed} there are horizontal {}``teeth'', however
theoretically the phase should last smoothly and without big leaps.
There are two possible solutions of that. We can shift the result
of APM. It means when we have the phase, we can find all the leaps
and make a smooth line shifting one after another. This solution concern
only the effect.

Another solution is to find a bug in APM source code. That seems more
preferable because it solves exactly the cause but not the effect.

\chapter{Conclusion\label{cha:Conclusion}}
\begin{quotation}
Look, your worship, what we see there are not giants but windmills
\cite{DonQuixote}.
\end{quotation}
\begin{flushright}
- Sancho Panza
\par\end{flushright}

Considering the statement of a question given in Introduction and
understanding the problem, we gave an outline of possible solutions
to mitigate the wind farm influence such as a signal processing, air
traffic management, construction and materials of wind turbines, and
how to choose the disposition of the wind farm, so that its impact
will be minimal. Among proposed solutions we have stopped on the last
one. So the visibility of the wind farm for the radar is determined
by RCS. In order to receive RCS it is necessary to have a 3D model
of a turbine and to know an electromagnetic field around the turbine
that comes from the radar through the atmosphere. We showed how to
draw a satisfactorily 3D model basing on specifications and on standard
airfoils. The electromagnetic field can be found by propagation models.
We considered different modern models and chose PE, which is implemented
in APM. Afterwards we improved APM source code in order to have the
complex field - with an amplitude and with a phase. Finally, we proposed
to combine RCS computation, propagation model and CAD model of the
turbine together using ASTRAD with connecting elements written in
Matlab.

The Chapter \ref{cha:Application-of-the} brought together ideas considered
above. In order to show the practical application of proposed solution,
we took an existed wind farm in Beinn Tharsuinn, the UK and the radar
STAR-2000 developed in Thales Air Systems. We evaluated two possible
disposition of the wind farm (1500 \emph{m} between them). Finally,
we received RCS of the wind turbine from these dispositions. There
is a huge difference between two locations. From the first position
the wind turbine reflects 30 times more energy than from the second
one. It means that the farm could be almost invisible for the radar
and, therefore, cause no problems for the air traffic management.
We did not concern ASTRAD to automate this solution and did the algorithm
{}``manually''. The reason is that it can be supplemented with another
tools like pseudo 3D model that has been considered as a further work.

The reasonable question could be why do we apply the complicated propagation
model? Why do not we use the terrain profile only? Indeed, using the
terrain profile we can hide the turbine from the radar. However, the
diffraction, reflection and refractivity make the propagation more
complicated and not obvious. It depends on different parameters of
the surface, atmosphere and weather conditions. This data is known
and statistically constant. Therefore, we proposed the means for an
assessment of the wind farm impact on the radar.

Also we showed the possible collaboration with different actors of
the wind power industry.

\begin{flushright}
\bibliographystyle{plain}
\clearpage\addcontentsline{toc}{chapter}{\bibname}\nocite{*}
\bibliography{Assessment_Of_The_Wind_Farm_Impact_On_The_Radar}

\par\end{flushright}
\end{document}